\documentclass[abbrvbib]{article}

\usepackage{xspace}
\usepackage{subcaption}
\usepackage[export]{adjustbox}

\graphicspath{{figures/},{graphs/}}
\usepackage{amsmath, latexsym}
\usepackage{amsfonts}
\usepackage{mathtools}
\usepackage{ntheorem}
\usepackage{dsfont}
\usepackage[dvipsnames]{xcolor}
\usepackage[colorinlistoftodos]{todonotes}
\usepackage{booktabs}
\usepackage{xfrac}
\usepackage{bbm}

\usepackage{algorithm}
\usepackage{algorithmicx}

\usepackage[most]{tcolorbox}
\usepackage{xparse}
\usepackage{lipsum}
\usepackage{changepage}
\usepackage{enumitem}

\newcommand{\qedwhite}{\hfill \ensuremath{\Box}}

\newcommand\cut[1]{}

\usepackage[percent]{overpic}

\newcommand{\squishlist}{
   \begin{list}{$\bullet$}
    { \setlength{\itemsep}{0pt}      \setlength{\parsep}{3pt}
      \setlength{\topsep}{3pt}       \setlength{\partopsep}{0pt}
      \setlength{\leftmargin}{1.5em} \setlength{\labelwidth}{1em}
      \setlength{\labelsep}{0.5em} } }

\newcommand{\squishlisttwo}{
   \begin{list}{$\bullet$}
    { \setlength{\itemsep}{0pt}    \setlength{\parsep}{0pt}
      \setlength{\topsep}{0pt}     \setlength{\partopsep}{0pt}
      \setlength{\leftmargin}{2em} \setlength{\labelwidth}{1.5em}
      \setlength{\labelsep}{0.5em} } }

\newcommand{\squishend}{
    \end{list}  }

\newenvironment*{gradexample}[1]{\textbf{\textit{Example (#1).}}}{$\qedwhite$}

\newcommand{\KLpq}[2]{\mathrm{KL}\left[{#1}\|{#2}\right]}

\newcommand{\myvec}[1]{\mathbf{#1}}
\newcommand{\myvecsym}[1]{\boldsymbol{#1}}

\newcommand{\vzero}{\myvecsym{0}}

\newcommand{\vbeta}{\myvecsym{\beta}}

\newcommand{\vepsilon}{\myvecsym{\epsilon}}

\newcommand{\veta}{\myvecsym{\eta}}
\newcommand{\vgamma}{\myvecsym{\gamma}}

\newcommand{\vmu}{\myvecsym{\mu}}

\newcommand{\vomega}{\myvecsym{\omega}}

\newcommand{\vphi}{\myvecsym{\phi}}

\newcommand{\vpsi}{\myvecsym{\psi}}

\newcommand{\vtheta}{\myvecsym{\theta}}

\newcommand{\vSigma}{\myvecsym{\Sigma}}
\newcommand{\vtau}{\myvecsym{\tau}}
\newcommand{\vupsilon}{\myvecsym{\upsilon}}

\newcommand{\vxhn}{\vxh^{(n)}}

\newcommand{\va}{\myvec{a}}

\newcommand{\vh}{\myvec{h}}

\newcommand{\vs}{\myvec{s}}
\newcommand{\vt}{\myvec{t}}

\newcommand{\vw}{\myvec{w}}

\newcommand{\vx}{\myvec{x}}
\newcommand{\vxh}{\hat{\vx}}

\newcommand{\vy}{\myvec{y}}

\newcommand{\vz}{\myvec{z}}

\newcommand{\vH}{\myvec{H}}
\newcommand{\vI}{\myvec{I}}

\newcommand{\vL}{\myvec{L}}

\newcommand{\vR}{\myvec{R}}

\newcommand{\vX}{\myvec{X}}
\newcommand{\expect}[2]{\mathbb{E}_{{#1}} \left[ {#2} \right]}
\newcommand{\expectGiven}[3]{\mathbb{E}_{{#1}} \left[ {#2} \middle| {#3} \right]}
\newcommand{\VarGiven}[3]{\Var_{{#1}}\left[ {#2} \middle| {#3}\right]}
\newcommand{\Cov}{\mbox{Cov}}
\newcommand{\Var}{\mathbb{V}}

\newcommand{\be}{\begin{equation}}
\newcommand{\ee}{\end{equation}}
\newcommand{\bea}{\begin{eqnarray}}
\newcommand{\eea}{\end{eqnarray}}
\newcommand{\beaa}{\begin{eqnarray*}}
\newcommand{\eeaa}{\end{eqnarray*}}

\DeclareMathAlphabet{\mathpzc}{OT1}{pzc}{m}{n}

\newcommand{\condSet}{\mathcal{S}}
\newcommand{\condSetC}{\mathcal{S}^{c}}
\newcommand{\gradCost}{cost\xspace}
\newcommand{\gradMeasure}{measure\xspace}
\newcommand{\dist}{p}
\newcommand{\distParams}{{\vtheta}}
\newcommand{\cost}{f}
\newcommand{\costParams}{{\vphi}}
\newcommand{\intd}{{\textrm{d}}}

\newcommand{\detchangevar}{\left|\nabla_{\vepsilon}g(\vepsilon; \distParams)\right|}

\newcommand{\gradientproblem}{\nabla_\distParams 
\expect{\dist(\vx; \distParams)}{\cost(\vx)}}

\newcommand{\ppos}{p^{+}}
\newcommand{\pneg}{p^{-}}

\usepackage{jmlr2e}
\jmlrheading
	{21}
	{2020}
	{1-63}
	{03/19; Revised 04/20}
	{04/20}
	{19-346}
	{S. Mohamed, M. Rosca, M. Figurnov, A. Mnih}

\ShortHeadings{Monte Carlo Gradient Estimation in Machine Learning}
{S. Mohamed, M. Rosca, M. Figurnov, A. Mnih}
\firstpageno{1}

\begin{document}
\title{Monte Carlo Gradient Estimation in Machine Learning}

\author{\name Shakir Mohamed${^*} {^1}$  \email shakir@google.com \\
\name Mihaela Rosca${^*} {^1} \ {^2}$  \email mihaelacr@google.com\\
\name Michael Figurnov$^* {^1}$  \email mfigurnov@google.com\\
\name Andriy Mnih$^* {^1}$  \email amnih@google.com\\
\addr $^*\!$Equal contributions; \\
\addr $1 \!$ DeepMind, London\\
\addr $2 \!$ University College London\\
}

\editor{Jon McAuliffe}
\maketitle

\begin{abstract}%
This paper is a broad and accessible survey of the methods we have at our
disposal for \textit{Monte Carlo gradient estimation} in machine learning and
across the statistical sciences: the problem of computing the gradient of an
expectation of a function with respect to parameters defining the distribution
that is integrated; the problem of sensitivity analysis. In machine learning
research, this gradient problem lies at the core of many learning problems, in
supervised, unsupervised and reinforcement learning. We will generally seek to
rewrite such gradients in a form that allows for Monte Carlo estimation,
allowing them to be easily and efficiently used and analysed. We explore three
strategies---the pathwise, score function, and measure-valued gradient
estimators---exploring their historical development, derivation, and underlying
assumptions. We describe their use in other fields, show how they are related
and can be combined, and expand on their possible generalisations. Wherever
Monte Carlo gradient estimators have been derived and deployed in the past,
important advances have followed. A deeper and more widely-held understanding of
this problem will lead to further advances, and it is these advances that we
wish to support.
\end{abstract}

\begin{keywords}
	gradient estimation, Monte Carlo, sensitivity analysis, score-function estimator, pathwise
	estimator, measure-valued estimator, variance reduction
\end{keywords}

\section{Introduction} %
Over the past five decades the problem of computing the gradient of an
expectation of a function---a stochastic gradient---has repeatedly
emerged as a fundamental tool in the advancement of the state of the art in
the computational sciences. An ostensibly anodyne gradient lies invisibly within
many of our everyday activities: within the management of modern supply-chains
\citep{siekman2000, kapuscinski1999optimal}, in the pricing and hedging of
financial derivatives \citep{glasserman2013monte}, in the control of traffic
lights \citep{rubinstein1993discrete}, and in the major milestones in the
ongoing research in artificial intelligence \citep{silver2016mastering}. Yet, computing
the stochastic gradient is not without complexity, and its fundamental importance
requires that we deepen our understanding of them to
sustain future progress. This is our aim in this paper: to provide a broad,
accessible, and detailed understanding of the tools we have to compute gradients
of stochastic functions. We also aim to describe their instantiations in other
research fields, to consider the trade-offs we face in choosing amongst the
available solutions, and to consider questions for future research.

Our central question is the computation of a general probabilistic objective $\mathcal{F}$ of the form
\begin{equation}
\mathcal{F}(\distParams) := \int \dist(\vx; \distParams) \cost(\vx; \costParams) \intd\vx =
\expect{\dist(\vx; \distParams)}{\cost(\vx; \costParams)} .
\label{eq:expectation_function}
\end{equation}
Equation \eqref{eq:expectation_function} is a \textit{mean-value analysis}, in
which a function $\cost$ of an input variable $\vx$ with \textit{structural
	parameters} $\costParams$ is evaluated on average with respect to an input
distribution $\dist(\vx; \distParams)$ with \textit{distributional parameters}
$\vtheta$. We will refer to $\cost$ as the \textit{\gradCost} and to $\dist(\vx;
\distParams)$ as the \textit{\gradMeasure}, following the naming used in
existing literature. We will make one restriction in this review, and that is
to problems where the measure $p$ is a probability distribution that is
continuous in its domain and differentiable with respect to its distributional
parameters. This is a general framework that allows us to cover
problems from queueing theory and variational inference to portfolio design and
reinforcement learning.

The need to learn the distributional parameters $\vtheta$ makes the gradient of the
function \eqref{eq:expectation_function} of greater interest to us:
\begin{equation}
\veta := \nabla_\distParams \mathcal{F}(\distParams) = \nabla_\distParams \expect{\dist(\vx;
	\distParams)}{\cost(\vx; \costParams)}.
\label{eq:gradient}
\end{equation}
Equation \eqref{eq:gradient} is the \textit{sensitivity analysis} of
$\mathcal{F}$, i.e.~the gradient of the expectation with respect to the
distributional parameters. It is this gradient that lies at the core of tools
for model explanation and optimisation. But this gradient problem is difficult in
general: we will often not be able to evaluate the expectation in closed form;
the integrals over $\vx$ are typically high-dimensional making quadrature
ineffective;  it is common to request gradients with respect to high-dimensional
parameters $\vtheta$, easily in the order of tens-of-thousands of dimensions;
and in the most general cases, the \gradCost function may itself not be
differentiable, or be a black-box function whose output is all we are able to
observe. In addition, for applications in machine learning, we will need
efficient, accurate and parallelisable solutions that minimise the number of
evaluations of the \gradCost. These are all challenges we can overcome by
developing Monte Carlo estimators of these integrals and gradients.

\textit{Overview.}
We develop a detailed understanding of Monte Carlo gradient estimation by first
introducing the general principles and considerations for Monte Carlo methods in
Section \ref{sect:stoch_opt}, and then showing how stochastic gradient estimation
problems of the form of \eqref{eq:gradient} emerge in five distinct research
areas. We then develop three classes of gradient estimators: the
\textit{score-function estimator}, the \textit{pathwise estimator}, and the \textit{measure-valued
	gradient estimator} in Sections
\ref{sect:score_func_estimators}--\ref{sect:weak_deriv_estimators}. We discuss
methods to control the variance of these estimators in Section
\ref{sect:var_reduction}. Using a set of case studies in Section
\ref{sect:case_studies}, we explore the behaviour of gradient
estimators in different settings
to make their application and design more concrete. We
discuss the generalisation of the estimators developed and other
methods for gradient estimation in Section \ref{sect:discussion}, and conclude
in Section \ref{sect:conclusion} with guidance on choosing estimators and
some of the opportunities for future research. This paper follows in the footsteps of several related
reviews that have had an important influence on our thinking, specifically,
 \citet{fu1994optimization}, \citet{pflug1996optimization}, \citet{vazquez2000course}, and
\citet{glasserman2013monte}, and which we generally recommend reading. 

\textit{Notation.}
Throughout the paper, bold symbols indicate vectors, otherwise variables are
scalars. $f(\vx)$  is  function of variables $\vx$ which may depend on
structural parameters $\vphi$, although we will not explicitly write out this
dependency since we will not consider these parameters in our exposition. We
indicate a probability distribution using the symbol $p$, and use the notation
$p(\vx; \vtheta)$ to represent the distribution over random vectors $\vx$ with
distributional parameters $\vtheta$. $\nabla_{\vtheta}$ represents the gradient
operator that collects all the partial derivatives of a function with respect to
parameters in $\vtheta$, i.e.~$\nabla_{\vtheta}f = [\tfrac{\partial f}{\partial
	\theta_1}, \ldots, \tfrac{\partial f}{\partial \theta_D}]$, for $D$-dimensional
parameters; $\nabla_\theta$ will also be used for scalar $\theta$. By convention, we consider vectors
to be columns and gradients as rows. We represent the sampling or
simulation of variates $\hat{x}$ from a
distribution $p(x)$ using the notation $\hat{x} \sim p(x)$. We use
$\mathbb{E}_p[f]$ and $\mathbb{V}_p[f]$ to denote the expectation and variance
of the function $f$ under the distribution $p$, respectively. Appendix \ref{app:distribution} lists the shorthand
notation used for distributions, such as $\mathcal{N}$ or $\mathcal{M}$ for the
Gaussian and double-sided Maxwell distributions, respectively.

\textit{Reproducibility.} 
Code to reproduce Figures \ref{fig:grad_var_quadratic} and
\ref{fig:grad_var_exp_cos}, sets of unit tests for gradient estimation, and for the
experimental case study using Bayesian logistic regression in Section
\ref{sect:blr_experiments} are available at
\texttt{\url{https://www.github.com/deepmind/mc_gradients}}\,.

\section{Monte Carlo Methods and Stochastic Optimisation}
\label{sect:stoch_opt}
This section briefly reviews the Monte Carlo method and the stochastic
optimisation setting we rely on throughout the paper. Importantly, this section
provides the motivation for why we consider the gradient estimation problem \eqref{eq:gradient} to be so
fundamental, by exploring an impressive breadth
of research areas in which it appears.
\subsection{Monte Carlo Estimators}
 \label{sect:MCE} 
The Monte Carlo method is one of the most general tools we have for the
computation of probabilities, integrals and summations. Consider the mean-value
analysis problem \eqref{eq:expectation_function}, which evaluates the expected
value of a general function $\cost$ under a distribution $\dist$. In most
problems, the integral \eqref{eq:expectation_function} will not be known in
closed-form, and not amenable to evaluation using numerical quadrature. However,
by using the Monte Carlo method \citep{metropolis1949monte} we can easily
approximate the value of the integral. The Monte Carlo method says that we can
numerically evaluate the integral by first drawing independent samples
$\hat\vx^{(1)}, \ldots, \hat\vx^{(N)}$ from the distribution $\dist(\vx;
\distParams)$, and then computing the average of the function evaluated at these
samples:
\begin{equation} 
\bar{\mathcal{F}}_N = \frac{1}{N} \sum_{n=1}^{N} \cost\left(\hat\vx^{(n)}\right), 
\text{ where } \hat\vx^{(n)} \sim \dist(\vx; \distParams) \text{ for }n=1,...,N.
\label{eq:mce_definition}
 \end{equation} 
The quantity $\bar{\mathcal{F}}_N$ is a random variable, since it depends on the
specific set of random variates $\{\hat\vx^{(1)}, \dots, \hat\vx^{(n)}\}$ used, and we
can repeat this process many times by constructing multiple sets of such random
variates. Equation \eqref{eq:mce_definition} is a Monte Carlo
\textit{estimator} of the expectation \eqref{eq:expectation_function}. 

As long
as we can write an integral in the form of Equation
\eqref{eq:expectation_function}---as a product of a function and a distribution
that we can easily sample from---we will be able to apply the Monte Carlo method
and develop Monte Carlo estimators. This is the strategy
we use throughout this paper.

There are four properties we will always ask of a Monte Carlo estimator:
\vspace{-4mm}
\begin{description}
\item[Consistency.] As we increase the number of samples $N$ in
\eqref{eq:mce_definition}, the estimate $\bar{\mathcal{F}}_N$ should converge to the true value of the
integral $\expect{\dist(\vx; \distParams)}{\cost(\vx)}$. This usually follows
from the strong law of large numbers.
\item[Unbiasedness.]  If we repeat the estimation process many times, we should
find that the estimate is centred on the actual value of the integral on average.  The
Monte Carlo estimators for \eqref{eq:expectation_function} satisfy this property
easily:
\begin{equation}
\expect{\dist(\vx; \distParams)}{\bar{\mathcal{F}}_N} = \expect{p(\vx; \vtheta)}{\frac{1}{N} 
\sum_{n=1}^{N}
 \cost\left(\vxhn\right)} =
\frac{1}{N} \sum_{n=1}^{N}  \expect{\dist(\vx; \distParams)}{\cost\left(\vxhn\right)} =
 \expect{\dist(\vx; \distParams)}{\cost(\vx)}.
\end{equation}
Unbiasedness is always preferred because it allows us to guarantee the
convergence of a stochastic optimisation procedure. Biased estimators can
sometimes be useful but require more care in their use \citep{mark2001bias}. 
\item[Minimum variance.] Because an estimator \eqref{eq:mce_definition} is a
random variable, if we compare two unbiased estimators using the same number of samples $N$, we will prefer
the estimator that has lower variance. We will repeatedly emphasise the
importance of low variance estimators for two reasons: the resulting gradient
estimates are themselves more accurate, which is essential for problems
in sensitivity analysis where the actual value of the gradient is the object of
interest; and where the gradient is used for stochastic optimisation,
low-variance gradients makes learning more efficient, allowing larger
step-sizes (learning rates) to be used, potentially reducing the overall number
of steps needed to reach convergence and hence resulting in faster training.

\item[Computational efficiency.] We will always prefer an estimator that is
computationally efficient, such as those that allow the expectation to be computed
using the fewest number of samples, those that have a computational
cost linear in the number of parameters, and those whose computations can be
easily parallelised.
\end{description}
We can typically assume that our estimators are consistent because of the
generality of the law of large numbers. Therefore most of our effort will be
directed towards characterising their unbiasedness, variance and computational
cost, since it is these properties that affect the choice we make between
competing estimators. Monte Carlo methods are widely studied, and the books by
\citet{robert2013monte} and \citet{owen2013} provide a deep coverage of their
wider theoretical properties and practical considerations. 
\subsection{Stochastic Optimisation} 
\label{sect:stoch_optim_subsect} 
The gradient \eqref{eq:gradient} supports at least two key computational tasks, those of
explanation and optimisation. Because the gradient provides a
computable value that characterises the behaviour of the cost---the cost's sensitivity
to changing settings---a gradient is directly useful as a tool with which to
\textit{explain} the behaviour of a probabilistic system. More importantly, the gradient
\eqref{eq:gradient} is the key quantity needed for \textit{optimisation} of the
distributional parameters $\vtheta$.
\begin{figure}[bt]
	\centering
	\includegraphics[height=3cm]{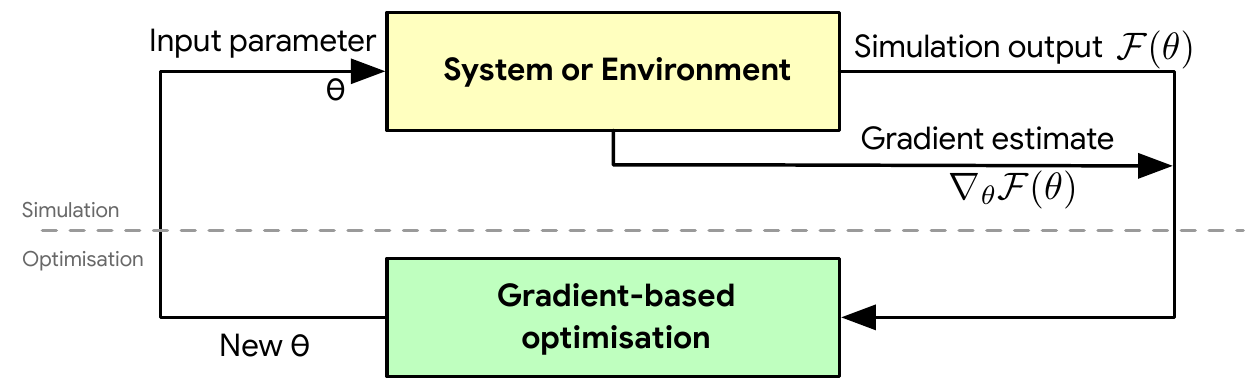}
	\caption{Stochastic optimisation loop comprising a simulation phase and an optimisation phase. 
	The simulation phase produces a simulation of the stochastic system or interaction with the 
	environment, as well as  unbiased estimators of the gradient.}
	\label{fig:stoch_optim_loop}
\end{figure}

Figure \ref{fig:stoch_optim_loop} (adapted from \citet{pflug1996optimization} sect. 1.2.5) depicts the general stochastic optimisation loop,
which consists of two phases: a simulation phase and an optimisation phase. This
is a stochastic system because the system or the environment has elements of
randomness, i.e.~the input parameters $\vtheta$ influence the system in a 
probabilistic manner. We will consider several case studies in Section
\ref{sect:case_studies} that all operate within this optimisation framework. Whereas in a
typical optimisation we would make a call to the system for a function value,
which is deterministic, in stochastic optimisation we make a call for an
\textit{estimate} of the function value; instead of calling for the gradient, we
ask for an estimate of the gradient; instead of calling for the Hessian, we will
ask for an estimate of the Hessian; all these estimates are random variables. Making a clear separation between the
simulation and optimisation phases  allows us to focus our attention on
developing the best \textit{estimators of gradients} we can, knowing that when
used with gradient-based optimisation methods available to us, we can guarantee
convergence so long as the estimate is \emph{unbiased}, i.e.~the estimate of the
gradient is correct on average \citep{kushner2003stochastic}.

If the optimisation phase is also used with stochastic approximation
\citep{robbins1951stochastic, kushner2003stochastic}, such as stochastic
gradient descent that is widely-used in machine learning, then this loop can be
described as a \textit{doubly-stochastic optimisation}
\citep{titsias2014doubly}. In this setting, there are now two sources of
stochasticity. One source arises in the simulation phase from the use of Monte
Carlo estimators of the gradient, which are random variables because of the
repeated sampling from the measure, like those in \eqref{eq:mce_definition}. A
second source of stochasticity arises in the optimisation phase from the use of
the stochastic approximation method, which introduces a source of randomness
through the subsampling of data points (mini-batching) when computing the
gradient. In Section \ref{sect:blr_experiments} we explore some of the
performance effects from the interaction between these sources of stochasticity.

\subsection{The Central Role of Gradient Estimation} 

Across a breadth of research areas, whether in approximate inference,
reinforcement learning, experimental design, or active learning, the need for
accurate and efficient computation of stochastic gradients and their
corresponding Monte Carlo estimators appears, making the gradient question one
of the fundamental problems of statistical and machine learning research. We
make the problem~\eqref{eq:gradient} more concrete by briefly describing
its instantiation in five areas, each with independent and thriving research
communities of their own. In them, we can see the central role of gradient
estimation by matching the pattern of the problem that arises, and in so doing, see
their shared foundations.
\vspace{-4mm}
\begin{description}
\item[Variational Inference.]
Variational inference is a general method for approximating complex and unknown
distributions by the closest distribution within a tractable family.
Wherever the need to approximate distributions appears in machine learning---in
supervised, unsupervised or reinforcement learning---a form of variational inference can
be used. Consider a generic probabilistic model $p(\vx | \vz)p(\vz)$ that
defines a generative process in which observed data $\vx$ is generated from a
set of unobserved variables $\vz$ using a data distribution $p(\vx | \vz)$ and a
prior distribution $p(\vz)$. In supervised learning, the unobserved variables
might correspond to the weights of a regression problem, and in unsupervised
learning to latent variables. The posterior distribution of this generative
process $p(\vz | \vx)$ is unknown, and is approximated by a variational
distribution $q(\vz | \vx; \vtheta)$, which is a parameterised family of
distributions with variational parameters $\vtheta$, e.g., the mean and variance
of a Gaussian distribution. Finding the distribution $q(\vz | \vx; \vtheta)$
that is closest to $p(\vz | \vx)$ (e.g., in the KL sense) leads to an objective, the variational
free-energy, that optimises an expected log-likelihood $\log p(\vx | \vz)$
subject to a regularisation constraint that encourages closeness between the
variational distribution $q$ and the prior distribution $p(\vz)$
\citep{jordan1999introduction, blei2017variational}. Optimising the distribution $q$  requires 
the gradient of the free energy with respect to the variational parameters $\vtheta$:
\begin{equation} 
\veta = \nabla_{\vtheta}\expect{q{(\vz | \vx; \vtheta)}}{\log p(\vx | \vz) - \log \frac{q(\vz |\vx; \vtheta)}{
p(\vz)}}. 
\end{equation} 
This is an objective that is in the form of Equation
\eqref{eq:expectation_function}: the cost is the difference between a
log-likelihood and a log-ratio (of the variational distribution and the prior
distribution), and the measure is the variational distribution $q(\vz | \vx;
\vtheta)$. This problem also appears in other research areas, especially in
statistical physics, information theory and utility theory
\citep{honkela2004variational}. Many of the solutions that have been developed
for scene understanding, representation learning, photo-realistic image generation, or
the simulation of molecular structures also rely on variational inference
\citep{eslami2018neural, kusner2017grammar, higgins2017beta}. In variational
inference we find a thriving research area where the problem \eqref{eq:gradient}
of computing gradients of expectations lies at its core.

\item[Reinforcement Learning.] Model-free policy search is an area of
reinforcement learning where we learn a policy---a distribution over
actions---that on average maximises the accumulation of long-term rewards. 
Through interaction in an environment, we can generate trajectories $\vtau =
(\vs_1, \va_1, \vs_2, \va_2, \dots, \vs_T, \va_T)$ that consist of pairs of states
$\vs_t$ and actions $\va_t$ for time period $t=1, \dots, T$. A policy is
learnt by following the \textit{policy gradient} \citep{sutton1998sutton}
\begin{equation} 
\veta = \nabla_{\vtheta} \expect{p(\vtau; \vtheta)}{\sum_{t=0}^{T} \gamma^t r(\vs_t, \va_t)},
\end{equation} 
which again has the form of Equation \eqref{eq:expectation_function}. The 
\gradCost is the return over the trajectory, which is a weighted sum of rewards
obtained at each time step $r(\vs_t, \va_t)$, with the discount factor $\gamma \in [0,1]$.
The measure is the joint distribution over states and actions
$p(\vtau; \vtheta) = \left[ \prod_{t=0}^{T-1} p(\vs_{t+1} | \vs_{t}, \va_t)p(\va_t | \vs_t;
\vtheta) \right] p(\va_T | \vs_T;\vtheta)$, which is the product of a state transition probability $p(\vs_{t+1} |
\vs_{t}, \va_t)$ and the policy distribution $p(\va_t | \vs_t; \vtheta)$ with
policy parameters $\vtheta$. The Monte Carlo gradient estimator used to
compute this gradient with respect to the policy parameters (see score function estimator later) leads to the
policy gradient theorem, one of the key theorems in reinforcement learning, which
lies at the heart of many successful applications, such as the AlphaGo system
for complex board games \citep{silver2016mastering}, and robotic control
\citep{deisenroth2013survey}. In reinforcement learning, we find yet another thriving area of
research where gradient estimation plays a fundamental role.

\item[Sensitivity Analysis.]
The field of sensitivity analysis is dedicated to the study of problems of the
form of \eqref{eq:gradient}, and asks what the sensitivity (another term for gradient) of an
expectation is to its input parameters. Computational finance is one area where
sensitivity analysis is widely used to compare
investments under different pricing and return assumptions, in order to choose a
strategy that provides the best potential future payout. Knowing the value of
the gradient gives information needed to understand the sensitivity of future payouts
to different pricing assumptions, and provides a direct measure of the financial
risk that an investment strategy will need to manage. In finance, these
sensitivities, or gradients, are referred to as the \textit{greeks}. A classic
example of this problem is the Black-Scholes option pricing model, which can be
written in the form of Equation \eqref{eq:expectation_function}: the \gradCost
function is the discounted value of an option with price $s_T$ at the time
of maturity $T$, using discount factor $\gamma$; and the \gradMeasure is a
log-normal distribution over the price at maturity $p(s_T; s_0)$, and is a
function of the initial price $s_0$. The gradient with respect to the initial
price is known as the Black-Scholes Delta \citep{chriss1996black}
\begin{equation} 
\Delta = \nabla_{s_0} \expect{p(s_T;s_0)}{e^{-\gamma T}\max(s_T - K, 0)},
\label{eq:delta_bs}
\end{equation} 
where $K$ is a minimum expected return on the investment;
delta is the risk measure that is actively minimised in
delta-neutral hedging strategies. The Black-Scholes delta \eqref{eq:delta_bs} above can be computed
in closed form, and the important greeks have closed or semi-closed form formulae
in the Black-Scholes formulation. Gradient estimation methods are used when the cost function
(the payoff) is more complicated, or in more realistic models where the measure
is not log-normal. In these more general settings, the gradient estimators we review
in this paper are the techniques that are still widely-used in financial
computations today \citep{glasserman2013monte}.

\item[Discrete Event Systems and Queuing Theory.] 
An enduring problem in operations research is the study of queues, the waiting
systems and lines that we all find ourselves in as we await our turn for a
service. Such systems are often described by a stochastic recursion \mbox{$L_t =
	x_{t} + \max(L_{t-1} - a_{t}, 0)$}, where $L_t$ is the total system time (of people
in the queue plus in service), $x_t$ is the service time for the $t$-th customer,
$a_t$ is the inter-arrival time between the $t$-th and the $(t+1)$-th customer \citep{vazquez2000course,
	rubinstein1996score}.  The number of customers 
served in this system is denoted by $\tau$. For convenience
we can write the service time and the inter-arrival time using the variable $y_t
= \{x_t, a_t\}$, and characterise it by a distribution $p(y_t; \vtheta) = p(x_t;
\theta_x)p(a_t;\theta_a)$ with distributional parameters $\vtheta = \{\theta_x,
\theta_a\}$. The expected steady-state behaviour of this system gives a problem
of the form of \eqref{eq:gradient}, where the \gradCost is the ratio of the
average total system time to the average number of customers served, and the
\gradMeasure is the joint distribution over service times $p(\vy_{1:T}; \vtheta)$
\citep{rubinstein1986monte}
\begin{equation}
\veta = \nabla_{\vtheta} \expect{p(\vy_{1:T};\vtheta)}{\frac{\sum_{t=1}^{T}L_t(y_{1:t})}{\tau(y_{1:T})}}.
\end{equation}
 In Kendall's notation, this is a general 
description for G/G/1 queues
\citep{newell2013applications}: queues with general distributions for the
arrival rate, general distributions for the service time, and a single-server
first-in-first-out queue. This problem also appears in many other areas and
under many other names, particularly in regenerative and semi-Markov processes,
and discrete event systems \citep{cassandras2009introduction}. In all these
cases, the gradient of the expectation of a cost, described as a ratio, is
needed to optimise a sequential process. Queues permeate all parts of our lives,
hidden from us in global supply chains and in internet routing, and visible to
us at traffic lights and in our online and offline shopping, all made efficient
through the use of Monte Carlo gradient estimators.

\item[Experimental Design.]
In experimental design we are interested in finding the best designs---the
inputs or settings to a possibly unknown system---that result in outputs that
are optimal with respect to some utility or score.  Designs are problem
configurations, such as a hypothesis in an experiment, the locations of sensors
on a device, or the hyperparameters of a statistical model
\citep{chaloner1995bayesian}. We evaluate the system using a given design $\vtheta$
and measure its output $y$, usually in settings where the measurements are
expensive to obtain and hence cannot be taken frequently. One such problem is
the probability-of-improvement, which allows us to find designs $\vtheta$ that on
average improve over a currently best-known outcome. This is an objective that
can be written in the form of Equation \eqref{eq:expectation_function}, where
the \gradCost is the score of a new design being better that the current best
design, and the \gradMeasure is a predictive function $p(y ; \vtheta)$, which
allows us to simulate the output $y$ for an input design $\vtheta$. To
find the best design, we will need to compute the gradient of the
probability-of-improvement with respect to the design $\vtheta$:
\begin{equation} 
\veta = \nabla_{\vtheta} \expect{p(y|\vtheta)}{\mathds{1}_{\{y < y_\text{best}\}}},
\end{equation}
where the indicator $\mathds{1}_{y<y_\text{best}}$ is one if the condition is
met, and zero otherwise. There are many such objectives, in areas
such as Bayesian optimisation, active learning and bandits
\citep{shahriari2016taking, wilson2017reparameterization}, all of which involve
computing the gradient of an expectation of a loss function, with wide use in
computer graphics, model architecture search, automatic machine learning, and
treatment design; again highlighting the central role that general-purpose
gradient estimators play in modern applications.

\end{description}
While these five areas are entire fields of their own, they are also important
problems for which there is ongoing effort throughout machine learning. There
are also many other problems where the need for computing stochastic gradients
appears, including systems modelling using stochastic differential equations,
parameter learning of generative models in algorithms such as variational
autoencoders, generative adversarial networks and generative stochastic networks
(\citet{rezende2014stochastic}; \citet{kingma2014auto};
\citet{goodfellow2014generative}; \citet{bengio2014deep}), in bandits and online
learning \citep{hu2016bandit},  in econometrics and simulation-based estimation
\citep{gourieroux1996simulation}, and in instrumental-variables estimation and
counter-factual reasoning \citep{hartford2016counterfactual}. An ability to
compute complicated gradients gives us the confidence to tackle increasingly
more complicated and interesting problems.

\section{Intuitive Analysis of Gradient Estimators}
\label{sect:intuitive_analysis}
The structure of the sensitivity analysis problem $\gradientproblem$
\eqref{eq:gradient} directly suggests that gradients can be computed in two
ways:
\vspace{-4mm}
\begin{description}
\item[Derivatives of Measure.] The gradient can be computed by differentiation
of the measure $\dist(\vx; \distParams)$. Gradient
estimators in this class include the score function
estimator (Section \ref{sect:score_func_estimators}) and the measure-valued
gradient (Section \ref{sect:weak_deriv_estimators}).
\item[Derivatives of Paths.] The gradient can be computed by differentiation of
the cost $\cost(\vx)$, which encodes the pathway from parameters $\distParams$,
through the random variable $\vx$, to the cost value. In this class of
estimators, we will find the pathwise gradient (Section
\ref{sect:pathwise_estimators}), harmonic gradient estimators and finite
differences (Section \ref{sect:harmonic_estimators}),  and Malliavin-weighted
estimators (Section \ref{sect:malliavinweighted}).
\end{description}
\vspace{-4mm}
We focus our attention on three classes of gradient estimators: the
\textit{score function}, \textit{pathwise} and \textit{measure-valued} gradient
estimators. All three estimators satisfy two desirable properties that we
identified previously, they are \textit{consistent} and \textit{unbiased}; but
they differ in their \textit{variance} behaviour and in their
\textit{computational cost}. Before expanding on the mathematical descriptions of these three gradient estimators, we compare their performance in simplified problems to develop an intuitive view of the differences
between these methods with regards to performance, computational cost,
differentiability, and variability of the cost function.

\begin{figure}[t]
	\centering
	\includegraphics[width=\textwidth]{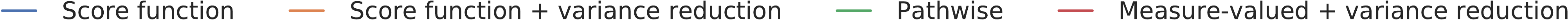} \\
	\includegraphics[width=0.42\textwidth]{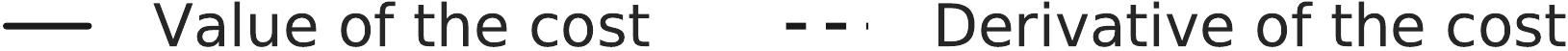} \\
	\vspace{0.2cm}
	\begin{subfigure}[t]{0.45\textwidth}
		\centering
		\includegraphics[width=\linewidth]{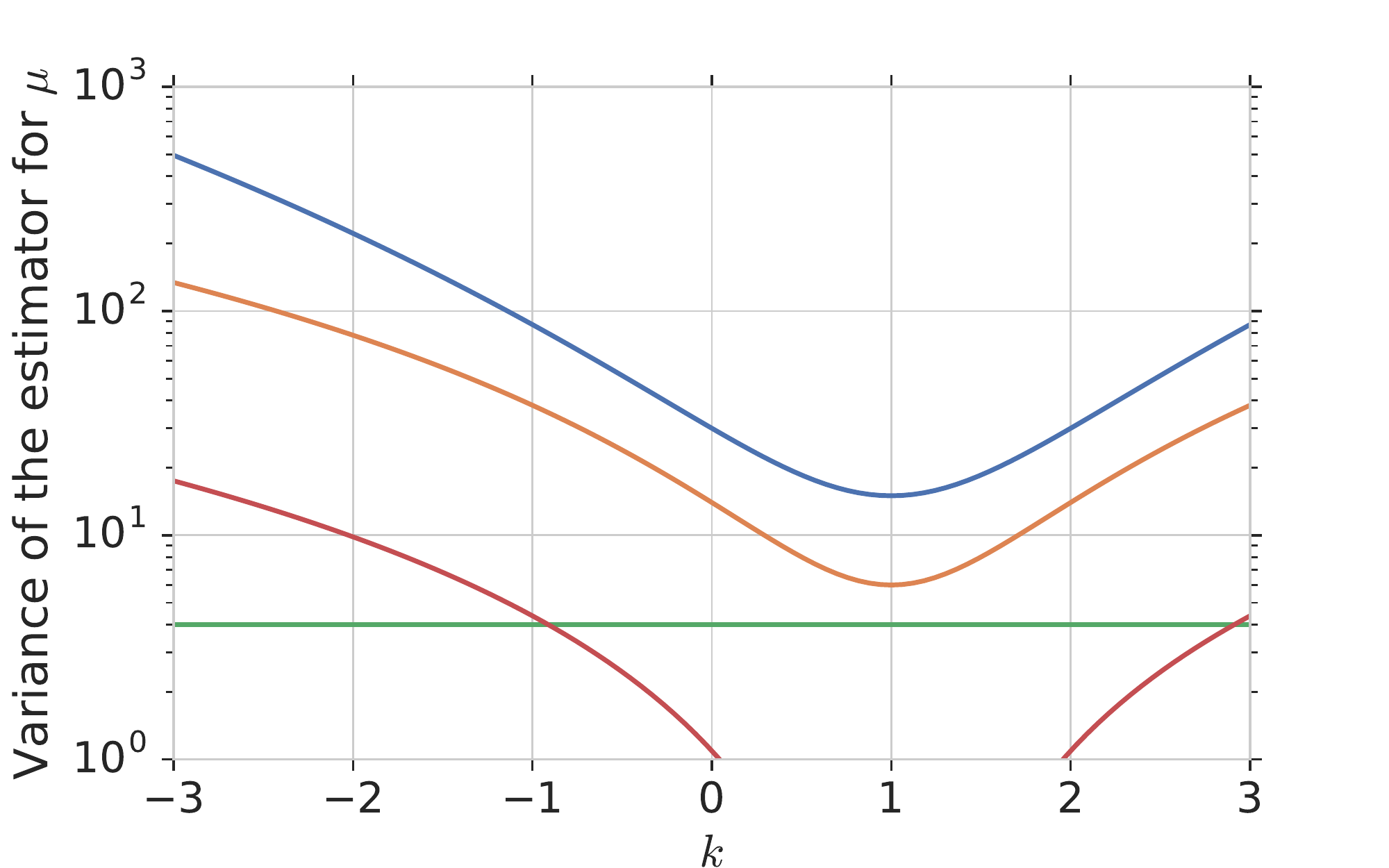} \\
		\includegraphics[width=0.9\linewidth,right]{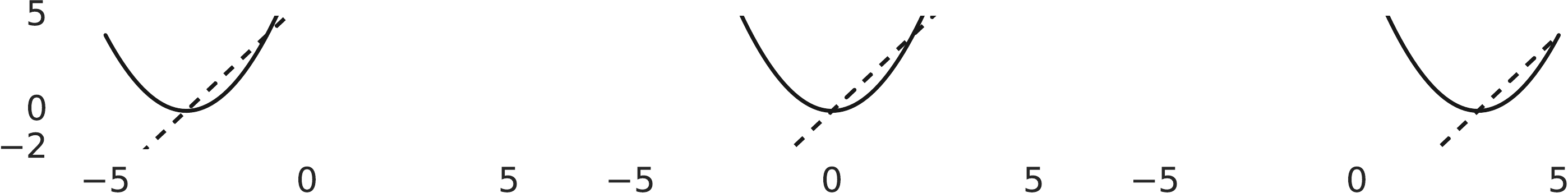}
	\end{subfigure}%
	\hspace{1cm}
	\begin{subfigure}[t]{0.45\textwidth}
		\centering
		\includegraphics[width=\linewidth]{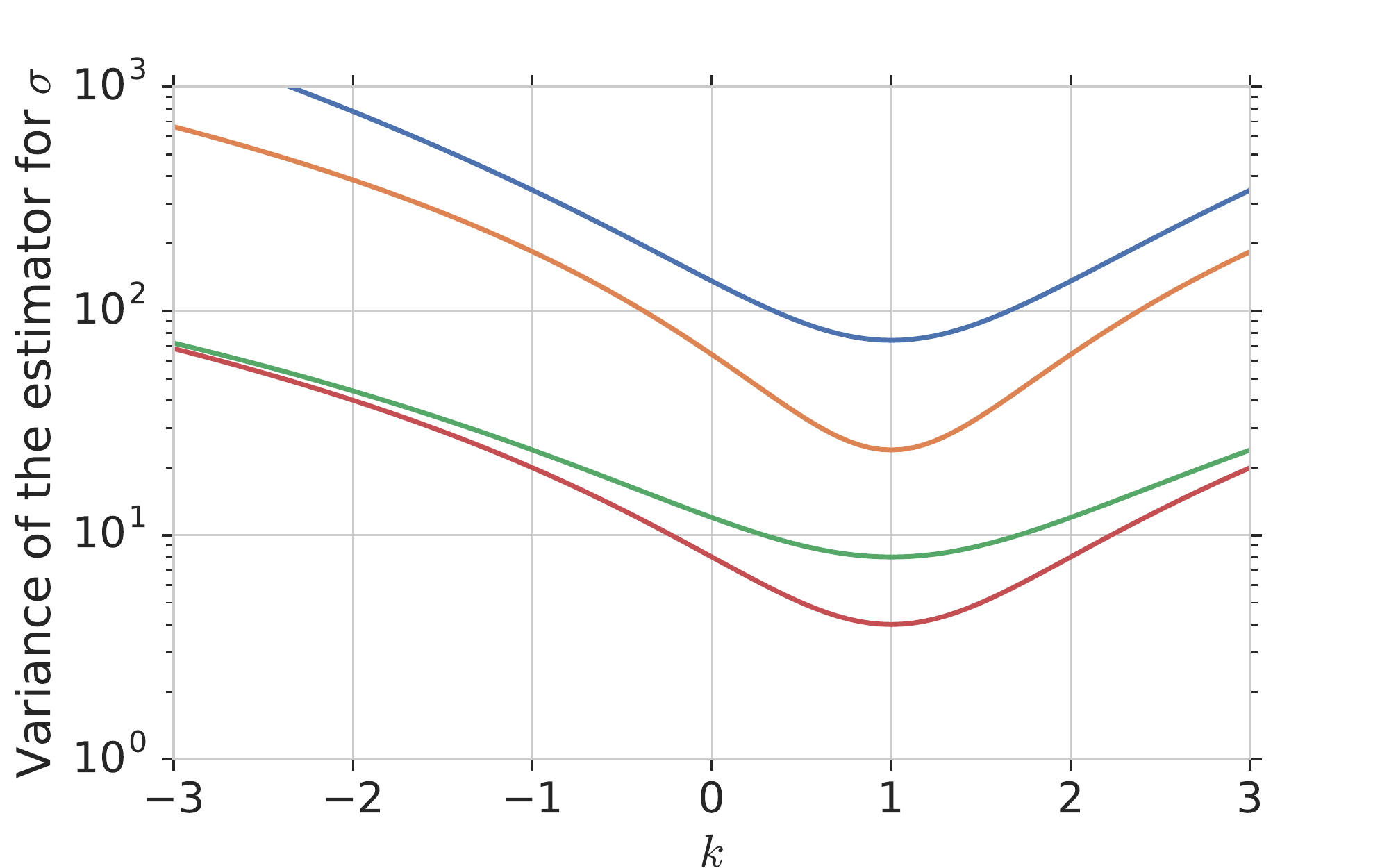} \\
		\includegraphics[width=0.9\linewidth,right]{costs_square}
	\end{subfigure}
	\caption{Variance of the stochastic estimates of $\nabla_{\theta} \expect{\mathcal{N}(x | \mu,
			\sigma^2)}{(x - k)^2}$ for $\mu = \sigma = 1$ as a function of $k$ for three different classes
		of gradient estimators. Left:  $\theta = \mu$; right: $\theta = \sigma$. The graphs in the bottom
		row
		show the function (solid) and its gradient (dashed) for $k \in \{-3, 0, 3\}$.}
	\label{fig:grad_var_quadratic}
\end{figure}

Consider the stochastic gradient problem
\eqref{eq:gradient} that uses Gaussian measures for three simple families of
cost functions, quadratics, exponentials and cosines:
\begin{equation}
\eta = \nabla_{\theta} \int \mathcal{N}(x|\mu, \sigma^2) f(x; k) dx;
\quad \theta \in \{\mu, \sigma\};
\quad f \in \{(x-k)^2, \exp(-kx^2), \cos(kx)\}.
\label{eq:grad_basic_example}
\end{equation}
We are interested in estimates of the gradient \eqref{eq:grad_basic_example}
with respect to the mean $\mu$ and the standard deviation $\sigma$ of the
Gaussian distribution. The cost functions vary with a parameter $k$, which
allows us to explore how changes in the cost affect the gradient. In the graphs
that follow, we use numerical integration to compute the variance of these
gradients. To reproduce these graphs, see the note on code in the introduction.
\paragraph{Quadratic costs.}
Figure \ref{fig:grad_var_quadratic} compares the variance of several gradient
estimators, for the quadratic function $f(x,k)=(x-k)^2$. For this function we
see that we could create a rule-of-thumb ranking: the highest variance
estimator is the score function, lower variance is obtained by the pathwise
derivative, and the lowest variance estimator is the measure-valued derivative.
But for the gradient with respect to the mean $\mu$, we also see that this is
not uniformly true, since there are quadratic functions, those with small or
large offsets $k$, for which the variance of the pathwise estimator is lower
than the measure-valued derivative. This lack of universal ranking is a general
property of gradient estimators.

The \textit{computational cost} of the estimators in Figure
\ref{fig:grad_var_quadratic} is the same for the score-function and pathwise
estimators: they can both be computed using a single sample in the Monte Carlo
estimator ($N=1$ in \eqref{eq:mce_definition}), even for multivariate
distributions, making them computationally cheap. The measure-valued derivative
estimator will require $2D$ evaluations of the cost function for $D$ dimensional
parameters, and for this the reason will typically not be preferred in
high-dimensional settings. We will later find that if the cost function is
\textit{not differentiable}, then the pathwise gradient will not be applicable.
These are considerations which will have a significant impact on the choice of
gradient estimator for any particular problem.

\begin{figure}[t]
	\centering
	\includegraphics[width=\textwidth]{estimators_legend} \\
	\includegraphics[width=0.42\textwidth]{costs_legend} \\
	\vspace{0.2cm}
	\begin{subfigure}[t]{0.45\textwidth}
		\centering
		\begin{overpic}[width=\linewidth,tics=10]{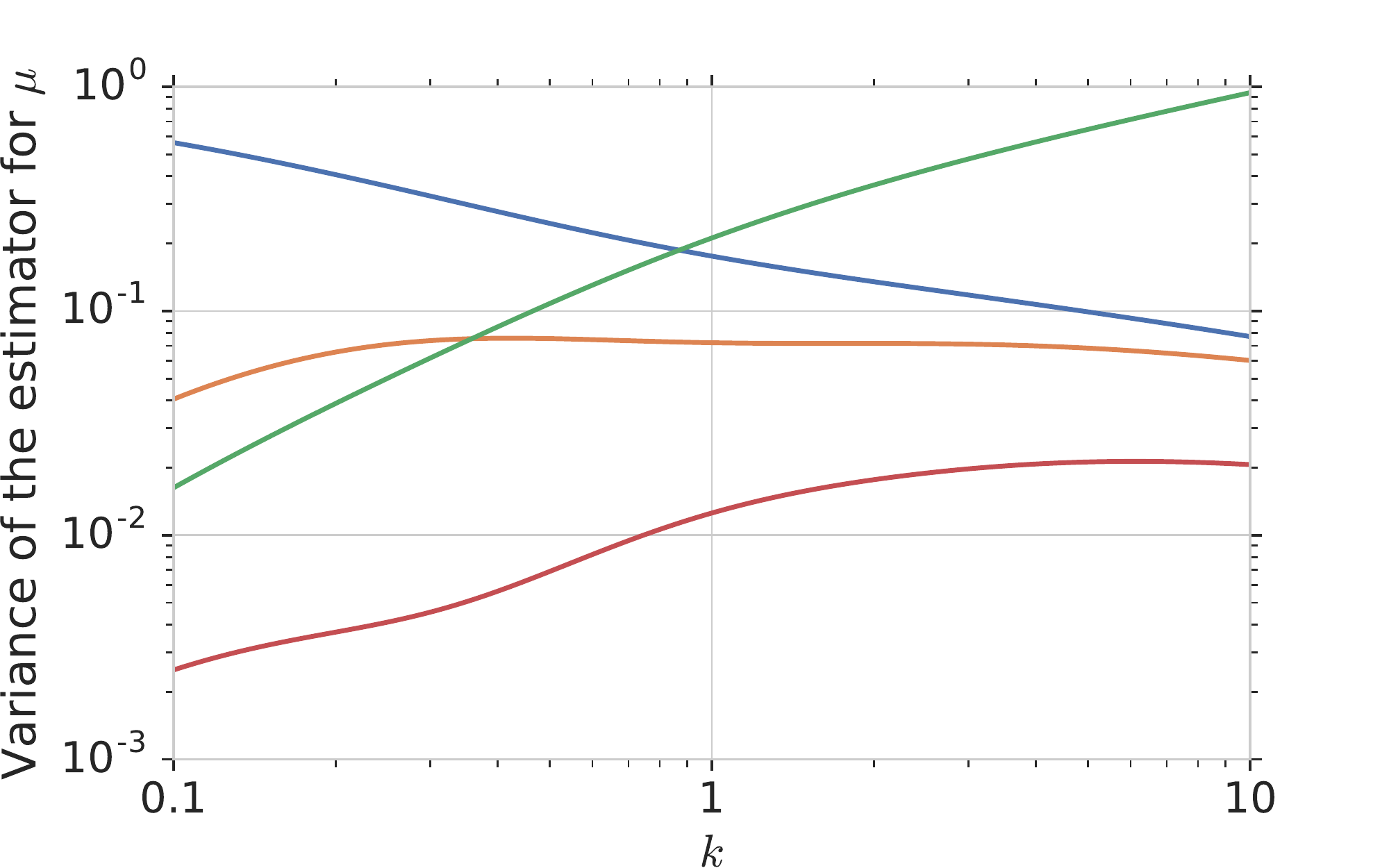}
			\put (14,59) {\tiny exp}
		\end{overpic}\\
		\includegraphics[width=0.9\linewidth,right]{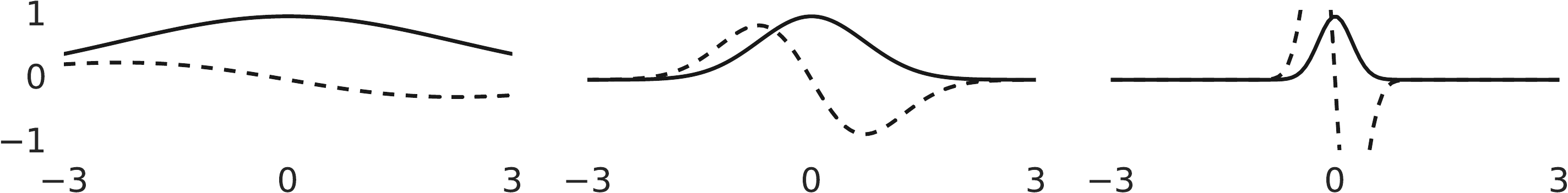}
	\end{subfigure}%
	\hspace{1cm}
	\begin{subfigure}[t]{0.45\textwidth}
		\centering
		\includegraphics[width=\linewidth]{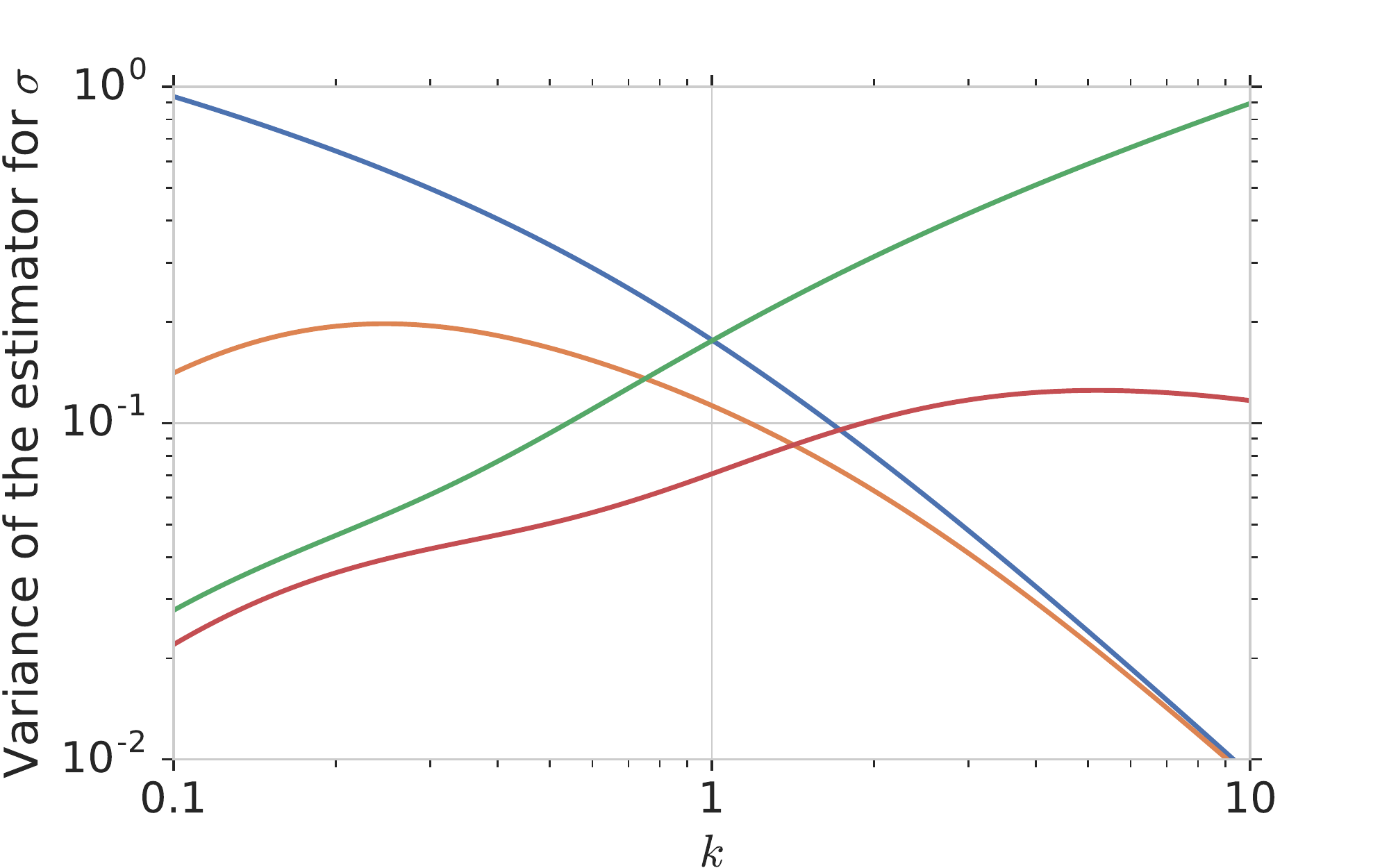} \\
		\includegraphics[width=0.9\linewidth,right]{costs_exp}
	\end{subfigure}
	\vspace{0.3cm}
	\begin{subfigure}[t]{0.45\textwidth}
		\centering
		\begin{overpic}[width=\linewidth,tics=10]{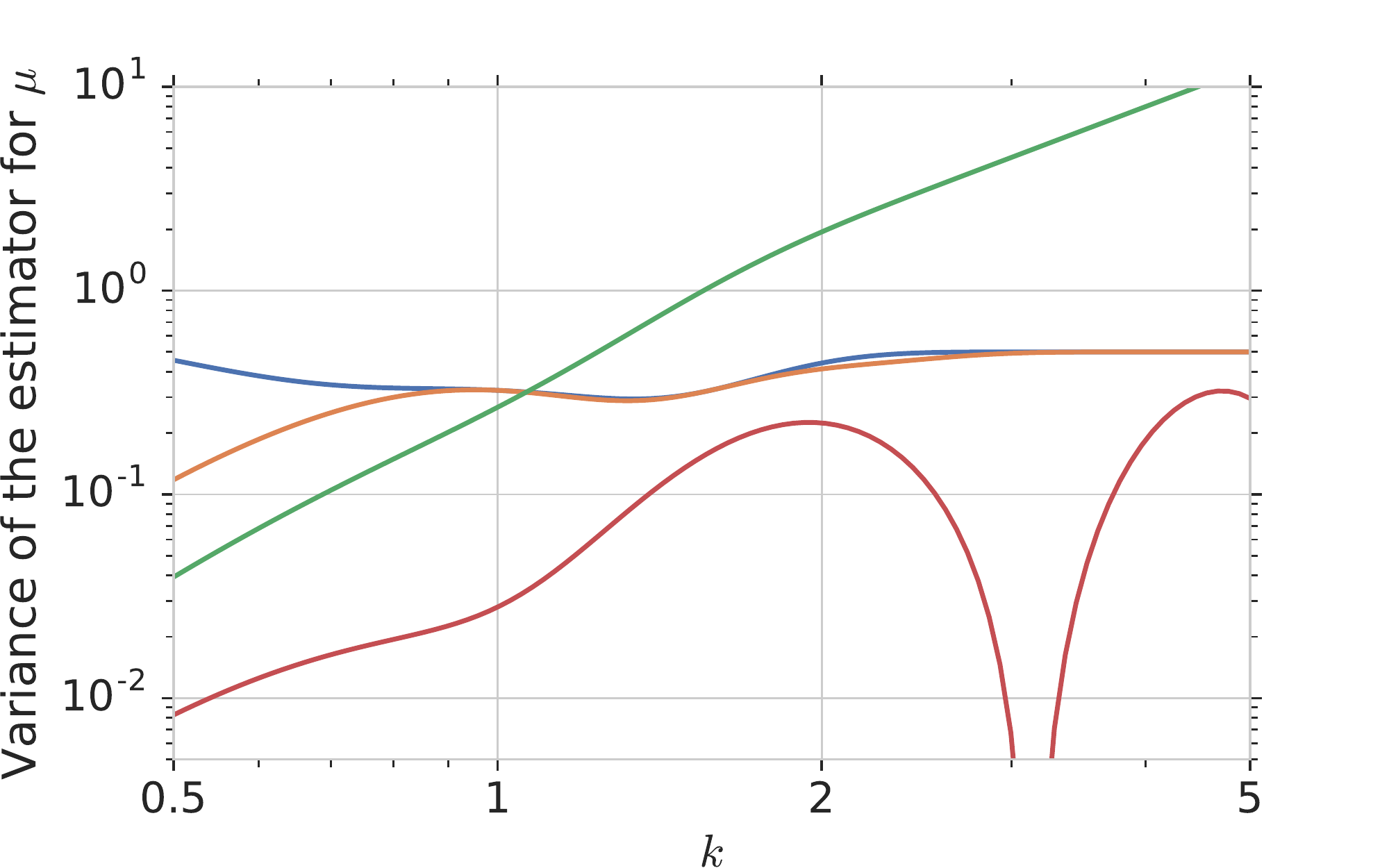}
			\put (14,59) {\tiny cos}
		\end{overpic}\\
		\includegraphics[width=0.9\linewidth,right]{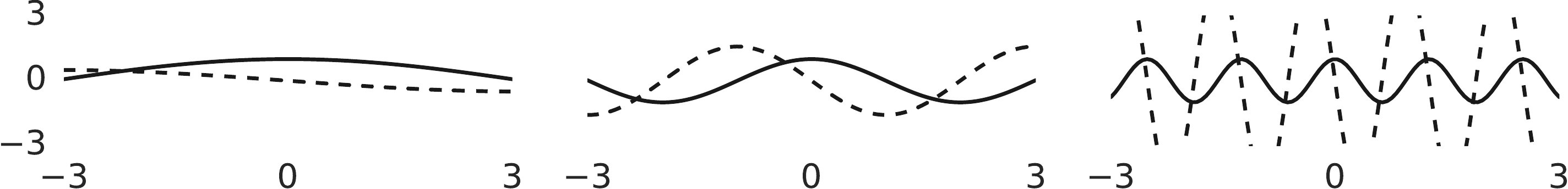}
	\end{subfigure}%
	\hspace{1cm}
	\begin{subfigure}[t]{0.45\textwidth}
		\centering
		\includegraphics[width=\linewidth]{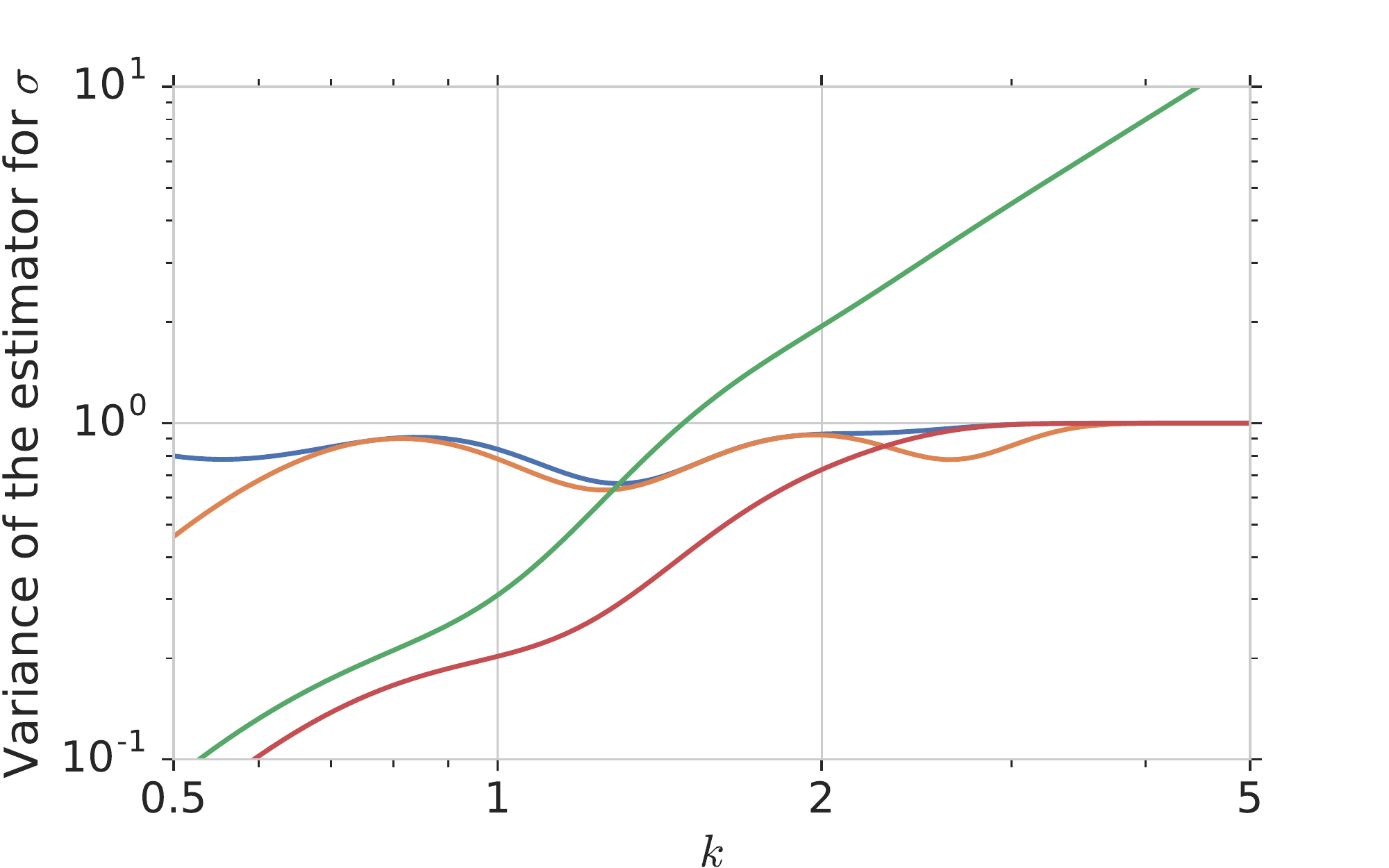} \\
		\includegraphics[width=0.9\linewidth,right]{costs_cos}
	\end{subfigure} \\
	\caption{Variance of the stochastic estimates of $\nabla_{\theta} \expect{\mathcal{N}(x | \mu,
			\sigma^2)}{\cost(x; k)}$ for $\mu = \sigma = 1$ as a function of $k$. Top:
		$\cost(x; k) = \exp(-kx^2)$, bottom: $\cost(x; k) = \cos kx$. Left: $\theta = \mu$; right: $\theta
		= \sigma$. The graphs in the bottom row
		show the function (solid) and its gradient (dashed): for $k \in \{0.1, 1, 10\}$ for the exponential
		function, and $k \in \{0.5, 1.58, 5.\}$ for the cosine function.}
	\label{fig:grad_var_exp_cos}
\end{figure}

\paragraph{Exponential and Cosine costs.}
Figure \ref{fig:grad_var_exp_cos} shows the variance behaviour for two other
functions, $\exp(-kx^2)$ and $\cos(kx)$. As the parameter $k$ of these functions
varies, the functions can become very sharp and peaked (see the black graphs at
the bottom), and this change in the cost function can change the effectiveness
of a gradient estimator: from being the lowest-variance estimator in one regime,
to being the highest-variance one in another. The green curve in Figure
\ref{fig:grad_var_exp_cos} for the pathwise estimator shows its variance
increasing as $k$ becomes larger. The measure-valued derivative (red curve) has
similar behaviour. The blue curve for the score function for the exponential
cost shows that its variance can behave in the opposite way, and can be lower
for higher values of $k$. We highlight this because, in machine learning
applications, we usually learn the cost function (by optimising its structural
parameters), and face a setting akin to varying $k$ in these graphs.
Therefore the process of learning influences the variance behaviour of an
estimator differently at different times over the course of learning, which we will need to take steps to control.

Figures \ref{fig:grad_var_quadratic} and \ref{fig:grad_var_exp_cos} also
demonstrate the importance of \textit{variance reduction}. The score function
estimator is commonly used with a \textit{control variate}, a way to reduce the variance
 of the gradient that we explore further in Section \ref{sect:var_reduction}. We see a
large decrease in variance by employing this technique. The variance of the
measure-valued derivative estimator in these plots is also shown with a form of
variance reduction (known as coupling), and for these simple cost functions,
there are regimes of the function that allow corrections that drive the variance
to zero; we can see this where the kink in the plot for the variance of the
mean-gradient for the cosine cost function.

From this initial exploration, we find that there are several criteria to be
judged when choosing an unbiased gradient estimator: computational cost,
implications on the use of differentiable and non-differentiable cost functions,
the change in behaviour as the cost itself changes (e.g., during learning), and the
availability of effective variance reduction techniques to achieve low variance.
We will revisit these figures again in subsequent sections as we develop the
precise description of these methods. We will assess each estimator based on
these criteria, working towards building a deeper understanding of them and
their implications for theory and practice.

\section{Score Function Gradient Estimators} \label{sect:score_func_estimators}
The score function estimator is in the class of derivatives of measure, and is
one of the most general types of gradient estimators available to us. Because of its
widespread and general-purpose nature, it appears under various names
throughout the literature, including the score function estimator
\citep{kleijnen1996optimization, rubinstein1996score}, the likelihood ratio method
\citep{glynn1990likelihood}, and the REINFORCE estimator
\citep{williams1992simple}. In this section, we will derive the estimator, expose some
of its underlying assumptions and possible  generalisations, explore its
behaviour in various settings, and briefly review its historical
development.

\subsection{Score Functions}
The \textit{score function} is the derivative of the log of a probability
distribution $\nabla_{\distParams} \log p(\vx; \distParams)$ with respect
to its distributional parameters, which can be expanded, using the
rule for the derivative of the logarithm, as
\begin{equation}
\nabla_{\distParams} \log p(\vx;
\distParams) = \frac{\nabla_{\distParams} p(\vx; \distParams)}{p(\vx;\distParams)}.
 \label{eq:log_deriv_identity}
\end{equation}
This identity is useful since it relates the derivative of a probability to the
derivative of a log-probability; for Monte Carlo estimators it will allow us to
rewrite integrals of gradients in terms of expectations under the measure $p$. It
is the appearance of the score function in the gradient estimator we develop in
this section that will explain its name.

The score function is important since, amongst other uses, it is the key
quantity in maximum likelihood estimation \citep{stigler2007epic}. One property
of the score that will later be useful is that its expectation is zero:
\begin{align}
 & \expect{p(\vx; \distParams)}{\nabla_{\distParams} \log
	p(\vx; \distParams)} = \int p(\vx; \distParams) \frac{\nabla_{\distParams}
	p(\vx; \distParams)}{p(\vx; \distParams)} d\vx = \nabla_{\distParams} \int p(\vx;
\distParams) d\vx = \nabla_{\distParams} 1 = \vzero.
\label{eq:expected_score}
\end{align}
We show this in \eqref{eq:expected_score} by first replacing the score using the
ratio given by the identity \eqref{eq:log_deriv_identity}, then cancelling
terms, interchanging the order of differentiation and integration, and finally
recognising that probability distributions must integrate to one, resulting in a
derivative of zero. A second important property of the score is that its
variance, known as the Fisher information, is an important
quantity in establishing the Cramer-Rao lower bound \citep{lehmann2006theory}.
\subsection{Deriving the Estimator}
Using knowledge of the score function, we can now derive a general-purpose
estimator for the sensitivity analysis problem \eqref{eq:gradient}; its
derivation is uncomplicated and insightful.
\begin{subequations}
\begin{align}
\veta & = \nabla_\distParams \expect{\dist(\vx; \distParams)}{\cost(\vx)}  = \nabla_\distParams
\int{\dist(\vx;
\distParams)\cost(\vx) d\vx} = \int{\cost(\vx) \nabla_\distParams \dist(\vx;
	\distParams) d\vx}
\label{eq:score_estimation_nabla_int_swap} \\
& = \int{\dist(\vx; \distParams) \cost(\vx) {\nabla_\distParams
		\log \dist(\vx; \distParams)} d\vx}
	 \label{eq:score_estimation_score_intro} \\
& = \expect{\dist(\vx; \distParams)}{\cost(\vx) \nabla_\distParams
	\log \dist(\vx; \distParams)}
\label{eq:score_estimation_final} \\
\bar{\veta}_N & = \frac{1}{N}\sum_{n=1}^N \cost(\hat\vx^{(n)}) \nabla_\distParams \log
\dist(\hat\vx^{(n)};
\distParams); \quad \hat\vx^{(n)} \sim \dist(\vx; \distParams). \label{eq:score_estimation_mce}
\end{align}
\label{eq:score_estimator_derivation}
\end{subequations}
\noindent
In the first line \eqref{eq:score_estimation_nabla_int_swap}, we expanded the
definition of the expectation as an integral and then exchanged the order of the
integral and the derivative; we discuss the validity of this operation in Section
\ref{sect:pathwise_unbiased}. In
\eqref{eq:score_estimation_score_intro}, we use the score identity
\eqref{eq:log_deriv_identity} to replace the gradient of the probability by the
product of the probability and the gradient of the log-probability. Finally, we
obtain an expectation \eqref{eq:score_estimation_final},  which is in the form we need---a product of a
distribution we can easily sample from and a function we can evaluate---to provide a Monte Carlo
estimator of the gradient in \eqref{eq:score_estimation_mce}.

Equation \eqref{eq:score_estimation_final} is the most basic form in which we
can write this gradient. One simple modification replaces the cost function with a
shifted version of it
\begin{equation}
 \veta = \expect{\dist(\vx; \distParams)}{(\cost(\vx) - \beta) \nabla_\distParams
	\log \dist(\vx; \distParams)},
\label{eq:score_estimation_final_baseline}
\end{equation}
where $\beta$ is a constant that we will call a baseline. For any value of $\beta$, we
still obtain an unbiased estimator because the additional term it introduces
has zero expectation due to the property \eqref{eq:expected_score} of the score.
This baseline-corrected form should be preferred to
\eqref{eq:score_estimation_final}, because, as we will see in Section \ref{sect:var_reduction}, it
allows for a simple but effective form of variance reduction.
\vspace{-2mm}
\subsection{Estimator Properties and  Applicability}
\label{sect:score_func_estimators_properties}
By inspecting the form \eqref{eq:score_estimation_final}, we can intuitively see that
the score-function estimator relates the overall gradient to the
gradient of the measure reweighted by the value of the cost function. This
intuitiveness is why the score function estimator was one of the first and most
widely-used estimators for sensitivity analysis. But there are several
properties of the score-function gradient to consider that have a deep impact on
its use in practice.
\vspace{-2mm}
\subsubsection{Unbiasedness}
\label{sect:pathwise_unbiased}
When the interchange between differentiation and integration in
\eqref{eq:score_estimation_nabla_int_swap} is valid, we will obtain an unbiased
estimator of the gradient \citep{lecuyer1995note}. Intuitively, since
differentiation is a process of limits, the validity of the interchange will
relate to the conditions for which it is possible to exchange limits and
integrals, in such cases most often relying on the use of the dominated
convergence theorem or the Leibniz integral rule
\citep{flanders1973differentiation, grimmett2001probability}. The interchange
will be valid if the following conditions are satisfied:
\begin{itemize}[noitemsep]
	\vspace{-4mm}
\item The measure $\dist(\vx; \distParams)$ is continuously differentiable in its
parameters $\distParams$.
\item The product  $\cost(\vx)\dist(\vx; \distParams)$ is both integrable and
	differentiable for all parameters $\distParams$.
\item There exists an integrable function $g(\vx)$ such that
	$\sup_{\distParams} \| \cost(\vx)\nabla_\distParams \dist(\vx; \distParams) \|_1 \le g(\vx)$ $\forall \vx$.
 \end{itemize}
	\vspace{-4mm}
These assumptions usually hold in machine learning applications, since the 
probability distributions that appear most often in machine
learning applications are smooth functions of their parameters.
\citet{lecuyer1995note} provides an in-depth discussion on the validity of
interchanging integration and differentiation, and also develops additional
tools to check if they are satisfied.
\subsubsection{Absolute Continuity}
\label{sect:abs_cont}
An important behaviour of the score function estimator is exposed by rewriting
it in one other way. Here, we consider a scalar distributional parameter
$\theta$ and look at its derivatives using first principles.
\begingroup
\allowdisplaybreaks
\begin{subequations}
\begin{align}
\nabla_{\theta} \expect{\dist(\vx; \theta)}{\cost(\vx)} & =
\int{\nabla_\theta \dist(\vx; \theta)\cost(\vx) d\vx} \label{eq:scorefn_abscont_def} \\
& = \int \lim_{h\to0}{\frac{\dist(\vx; \theta + h) - \dist(\vx;
		\theta)} {h}\cost(\vx)d\vx}
	\label{eq:scorefn_abscont_limit}\\
& = \lim_{h\to0} \int {\frac{\dist(\vx; \theta + h) - \dist(\vx;
		\theta)} {h}\cost(\vx)d\vx}
\label{eq:scorefn_abscont_limit_exchange}\\ %
& =  \lim_{h\to0} \frac{1}{h} \int {\dist(\vx;\theta) \frac{\dist(\vx;
		\theta + h) - \dist(\vx; \theta)}{\dist(\vx; \theta)}\cost(\vx)d\vx} \label{eq:scorefn_abscont_identity}\\
& = \lim_{h\to0} \frac{1}{h} \int {\dist(\vx;\theta) \left(
	\frac{\dist(\vx; \theta + h)}{\dist(\vx; \theta)} - 1 \right)
	\cost(\vx)d\vx} \label{eq:scorefn_abscont_simplify}\\
&=  \lim_{h\to0} \tfrac{1}{h}\left( \expect{\dist(\vx; \theta)} { \omega(\theta, h)
	\cost(\vx)} - \expect{\dist(\vx; \theta)}{\cost(\vx)}\right).
	\label{eq:scorefn_abscont_is}
\end{align}
\end{subequations}
\endgroup
In the first line \eqref{eq:scorefn_abscont_def}, we again exchange the order of
integration and differentiation, which we established is safe in most use-cases.
We then expand the derivative in terms of its limit definition in
\eqref{eq:scorefn_abscont_limit}, and swap the order of the limit and the
integral in \eqref{eq:scorefn_abscont_limit_exchange}. We introduce an identity
term in \eqref{eq:scorefn_abscont_identity} to allow us to later rewrite the
expression as an expectation with respect to the distribution $\dist(\vx;
\theta)$. Finally, we simplify the expression, separating it in two terms
\eqref{eq:scorefn_abscont_simplify}, denoting the importance weight $\omega(\theta, h) :=
\tfrac{\dist(\vx; \theta + h)}{\dist(\vx; \theta)}$, and rewrite the gradient
using the expectation notation \eqref{eq:scorefn_abscont_is}. It is this final
expression that exposes a hidden requirement of the score function estimator.

The ratio $\omega(\theta, h)$ in \eqref{eq:scorefn_abscont_is}  is similar to
the one that appears in importance sampling \citep{robert2013monte}. Like
importance sampling, the estimator makes an implicit assumption of absolute
continuity, where we require $\dist(\vx; \theta + h)>0$ for all points where
$\dist(\vx; \theta) > 0$. Not all distributions of interest satisfy this
property, and failures of absolute continuity can result in a biased gradient.
For example, absolute continuity is violated when the parameter $\theta$
defines the support of the distribution, such as the uniform distribution
$\mathcal{U}[0, \theta]$.

\begin{gradexample}{Bounded support}
Consider the score-function
estimator for a \gradCost $f(x) = x$ and distribution $\dist(x; \theta) =
\frac{1}{\theta}\mathds{1}_{\{0<x<\theta\}}$, which is differentiable in $\theta$
when $x \in (0, \theta)$; the score $s(x) = -\sfrac{1}{\theta}$. This is a popular example also used by
\citet{glasserman2013monte} and \citet{pflug1996optimization}, amongst others.
Comparing the two gradients:
\begin{subequations}
\begin{eqnarray}
\textrm{True gradient: }& \nabla_\theta \expect{\dist(x; \theta) }{x} =
\nabla_\theta \left( \left.
\frac{x^2}{2\theta}
\right|_0^\theta\right) = \frac{1}{2}. \label{eq:uniform_dist_example} \\
\textrm{Score-function gradient: } & \expect{\dist(x; \theta) }{x
\frac{-1}{\theta}}=-\frac{\theta/2}{\theta}
=-\frac{1}{2}.
\end{eqnarray}
\end{subequations}
In this example, the estimator fails to provide the correct gradient because
$\dist(x; \theta)$ is not absolutely continuous with respect to $\theta$ at the
boundary of the support.
\end{gradexample}
\subsubsection{Estimator Variance}
From the intuitive analysis in Section \ref{sect:intuitive_analysis}, where we compared the
score function estimator to other gradient estimation techniques (which we
explore in the subsequent sections), we see that even in those simple
univariate settings the variance of the score-function estimator can vary
widely as the cost function varies (blue curves in Figures
\ref{fig:grad_var_quadratic} and \ref{fig:grad_var_exp_cos}). Starting from the
estimator \eqref{eq:score_estimation_mce} and denoting the estimator mean as
$\mu(\theta) :=\expect{p(\vx; \theta)}{\bar{\eta}_N}$, we can write the
variance of the score function estimator for $N=1$ as
\begin{equation}
\mathbb{V}_{p(\vx;\theta)}[\bar{\eta}_{N=1}] = \expect{p(\vx; \theta)}{\big(f(\vx)\nabla_{\theta}\log
p(\vx; \theta)\big)^2} - \mu(\theta)^2,
\label{eq:score_fn_var_def}
\end{equation}
which writes out the definition of the variance. Although this is a scalar expression, it allows us to intuitively see that the parameter dimensionality can play an important role in the estimator's variance because the score function in \eqref{eq:score_fn_var_def} has the same dimensionality as the distributional parameters. 
The alternative
form of the gradient we explored in Equation \eqref{eq:scorefn_abscont_is} provides another way to
characterise the variance
\begin{equation}
\mathbb{V}_{p(\vx;\theta)}[\bar{\eta}_{N=1}] = \lim_{h\to0} \tfrac{1}{h}\expect{\dist(\vx;
\theta)}{(\omega(\theta, h)-1)^2
	\cost(\vx)^2} -  \mu(\theta)^2,
\label{eq:score_fn_var_abscont}
\end{equation}
which, for a fixed $h$, exposes the dependency of the variance on the importance weight $\omega$.
Although we will not explore it further, we find it instructive to connect these variance expressions to
the
variance bound for the estimator given by the Hammersley-Chapman-Robbins bound \citep[ch
2.5]{lehmann2006theory}
\begin{equation}
\mathbb{V}_{\dist(\vx; \theta)}[\bar{\eta}_{N=1}] \geq \sup_h
\frac{\left(\mu(\theta + h) - \mu(\theta)\right)^2}{\expect{p(\vx;
\theta)}{\frac{p(\vx; \theta + h)}{p(\vx; \theta)} - 1}^2} =  \sup_h \frac{\left(\mu(\theta + h) -
\mu(\theta)\right)^2}{\expect{p(\vx;
\theta)}{\omega(\theta,h) - 1}^2},
\label{eq:hcr_bound}
\end{equation}
which is a generalisation of the more widely-known Cramer-Rao bound and describes the
minimal variance achievable by the estimator.

Our understanding of the gradient variance can then be built by
exploring three sources of variance: contributions from the implicit importance
ratio $\omega(\theta,h)$ that showed the need for absolute continuity,
contributions from the dimensionality of the parameters, and contributions from
the variance of the cost function. These are hard to characterise exactly for general cases, but we can develop an intuitive understanding by unpacking specific terms of the gradient variance.

\setlength{\leftskip}{0.5cm}
\paragraph{Variance from the importance ratio $\vomega$.}

The variance characterisation from either Equation~\eqref{eq:score_fn_var_abscont} or
\eqref{eq:hcr_bound} shows that the
importance ratio $\omega(\theta,h)$ directly affects the variance. The simplest
way to see this effect is to consider the contributions to the variance using
the form of the gradient in Equation \eqref{eq:scorefn_abscont_simplify}, for a
fixed $h$. The first term in that equation is the quadratic term
\begin{equation}
\expect{\dist(\vx; \theta)}{(\omega(\theta,h)-1)^2 \cost(\vx)^2},
\label{eq:score_function_var_weight}
\end{equation} %
where $\omega(\theta,h)$ is the importance ratio we identified in Equation
\eqref{eq:scorefn_abscont_simplify}. We will obtain finite variance
gradients when the integral in \eqref{eq:score_function_var_weight} is finite,
i.e.~when the conditions for absolute continuity are satisfied. We saw previously
that failures of absolute continuity can lead to biased gradients. In practice,
complete failure will be rare and we will instead face \textit{near}-failures in
maintaining absolute continuity, which as the above expression shows, will lead
to an increase in the variance of the Monte Carlo estimator of the gradient.

\paragraph{Variance due to input dimensionality.}

Assume for simplicity that the distribution factorises over the dimensions of
$\vx \in \mathbb{R}^D$ so that $p(\vx; \theta) = \prod_d p(x_d; \theta)$, and
again consider a scalar parameter $\theta$. In expectation, the importance
weights have the property that for any dimension $D$, $\prod_{d=1}^{D}
\expect{p(x_d;\theta)}{ \tfrac{\dist(x_d;\theta + h)}{\dist(x_d; \theta)}}= 1.$
The importance weight and its logarithm are given by $\omega(\theta,h) =
\prod_{d=1}^D \omega_d(\theta, h) = \prod_{d=1}^D\tfrac{\dist(x_d; \theta +
	h)}{\dist(x_d; \theta)}$ and $\log\omega(\theta, h) = \sum_{d=1}^D \log
\omega_d(\theta, h) = \sum_{d=1}^D \log \tfrac{\dist(x_d; \theta +
	h)}{\dist(x_d; \theta)} $, which we can use to study the behaviour of the
importance weight as the dimensionality of $\vx$ changes.

If we follow an
argument due to \citet{glynn1989importance} and \citet[p.
259]{glasserman2013monte}, and assume that the expectation of the log-ratio is bounded, i.e.~$\expect{p(\vx; \theta)}{\log \omega(\theta,h)} < \infty$,
then using the strong law of large numbers we can show that this expectation
converges to a constant, $\expect{p(\vx; \theta)}{\log \omega(\theta, h)} = c$.
Using Jensen's inequality, we know that $c \leq \log \expect{p(\vx;
	\theta)}{\prod_{d=1}^D \tfrac{p(x_d; \theta+h)}{p(x_d; \theta)}} = 0$, with equality only when
$p(x_d; \theta+h) = p(x_d; \theta)$, meaning $c<0$.  As a result, the  sum of
many such terms has a limiting behaviour:
\begin{equation} \lim_{d\to\infty} \sum_d \log \frac{p(x_d;
	\theta+h)}{p(x_d; \theta)} = -\infty \implies \lim_{d\to\infty} \omega(\theta,h) = \lim_{d\to\infty} \prod_d
	\frac{p(x_d;
	\theta+h)}{p(x_d; \theta)} = 0,
\end{equation}
where we reach the limit in the first term since it is a sum of negative terms,
and the second expression is obtained by exponentiation.  As the dimensionality
increases, we find that the importance weights converge to zero, while at the
same time their expectation is one for all dimensions. This difference between the instantaneous and average behaviour of $\omega(\theta, h)$ means that in high
dimensions the importance ratio can become highly skewed, taking large values
with small probabilities and leading to high variance as a consequence.
\paragraph{Variance from the cost.} %

\begin{figure}[t]
	\centering
	\begin{subfigure}[t]{0.45\textwidth}
		\centering
	\includegraphics[width=\textwidth]{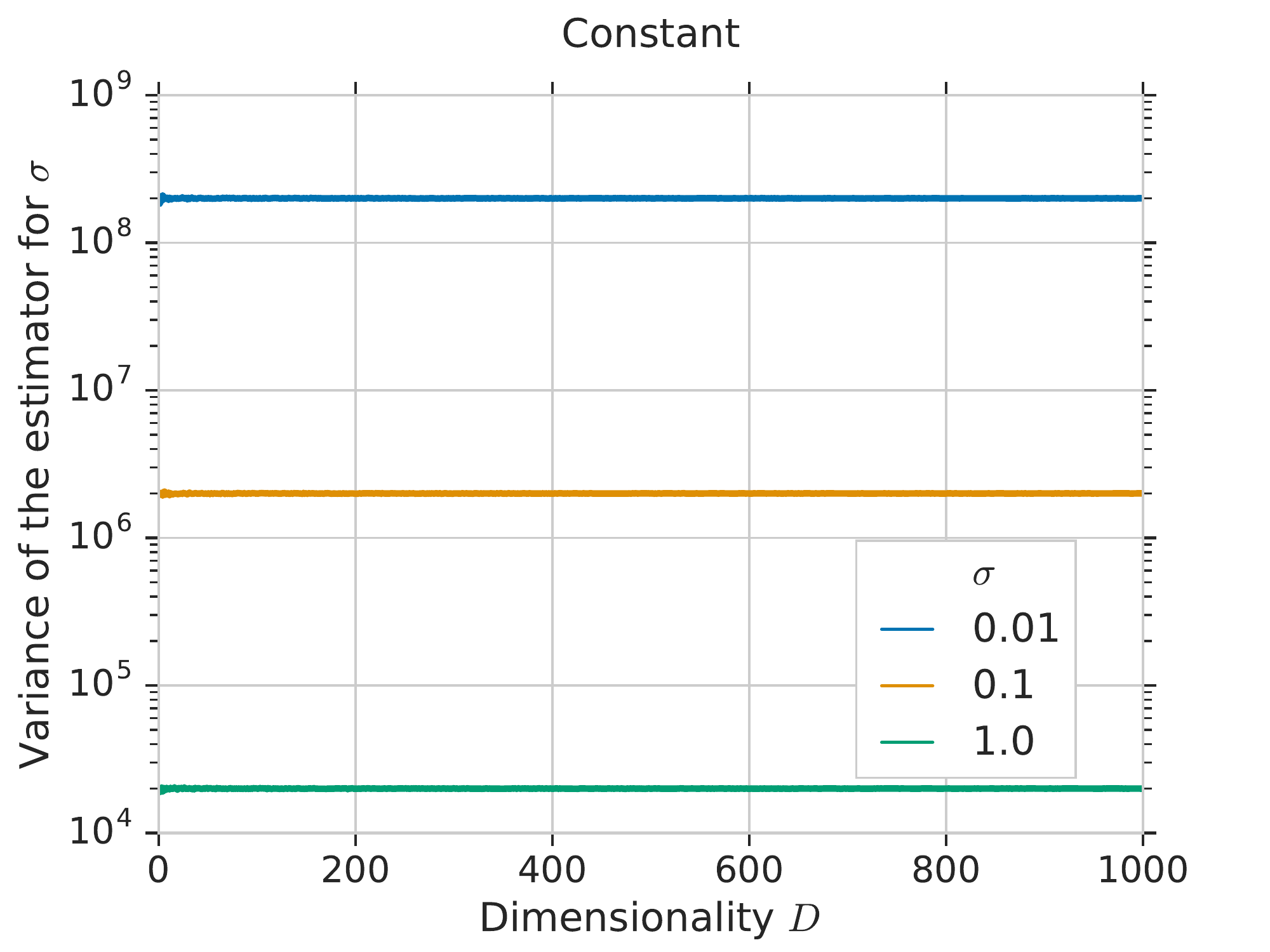}
	\end{subfigure}%
	\hspace{1cm}
	\begin{subfigure}[t]{0.45\textwidth}
	\centering
	\includegraphics[width=\textwidth]{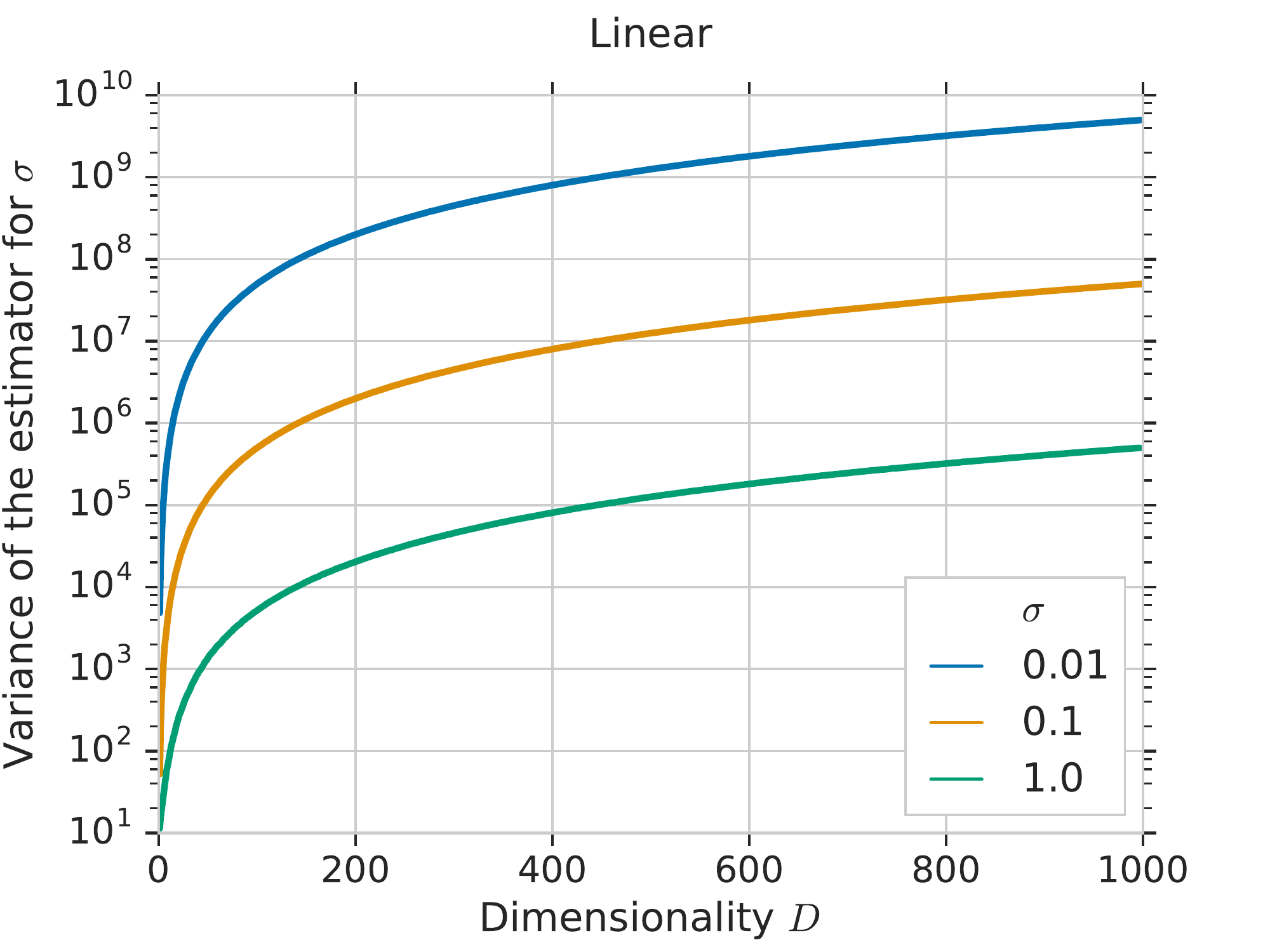}
\end{subfigure}%
\caption{Variance of the score function estimator for a Gaussian measure $\mathcal{N}(\vx | \tfrac{1}{2},\sigma^2\vI_D)$
	and two cost functions: a constant one $f(\vx)=100$ and a linear one $f(\vx)= \sum_d x_d$.
	The y-axis is the estimate of the average variance $\Var[\nabla_{\sigma}f(\mathbf{x})] $
	across parameter dimensions.}
\label{fig:reinforce_var}
\end{figure}

The cost function appears in all forms of the variance we wrote
\eqref{eq:score_fn_var_def}--\eqref{eq:hcr_bound} and is itself a significant
contributor to the estimator variance, since it is a multiplicative term in the
gradient. For example, if the cost function is a sum of $D$ terms, $f(\vx) =
\sum_k f(x_d)$, whose individual variance we assume is bounded, then the
variance of the score-function estimator $\mathbb{V}[\nabla_{\vtheta} \log p(\vx;
\vtheta) f(\vx)]$ will be of order $O(D^2)$. Because the cost function is a
black-box as far as the estimator is concerned, it will typically contain many
elements that do not directly influence the parameters, and hence will not
affect the gradient. But every extra term contributes to its variance: ideally,
before multiplying the cost with the score function, we would eliminate the
parts of it that have no influence on the  parameter whose derivative we are
computing. Alternatively, we can try to do this automatically by using the
gradient of the function itself (but only if it is available) to remove these
terms---this approach is explored in the next section.

To support this intuition, we explore the effect of the cost function and
measure on the variance properties of the score-function estimator using an example.
Figure~\ref{fig:reinforce_var} shows the estimator variance for
the gradient $\nabla_\sigma \expect{\mathcal{N}(\vx | 0.5,
  \sigma^2\vI_D)}{f(\vx)}$, for a constant cost $f(\vx) =100$ and a linear cost that
sums the dimensions of  $f(\vx) = \sum_d x_d$ for $\vx \in \mathbb{R}^D$.
For both cost functions, the expected value of the cost under the measure
has no dependence on the scale parameter $\sigma$. As we expected, for the linear cost, the variance
scales quadratically as the number of terms in the cost increases.

\setlength{\leftskip}{0pt}
This variance analysis is meant to emphasise the importance of understanding the
variance in our estimators. Whether because of the influence of the structure of
the cost function, or the dimensionality of the implied importance
weights in our gradients, we will need to counterbalance their effects and
remove excess variance by some method of \textit{variance reduction}. The
question of variance reduction will apply to every type of Monte Carlo gradient
estimator, but the specific solutions that are used will differ due to the
different assumptions and properties of the estimator. We will study
variance reduction in more detail in Section \ref{sect:var_reduction}.

\subsubsection{Higher-order Gradients}
Higher derivatives are conceptually simple to compute using the score
function estimator. The score of higher orders is defined as
 \begin{equation}
s^{(1)}(\vx) = \frac{\nabla_{\distParams}
	\dist(\vx; \distParams)}{\dist(\vx; \distParams)} = \nabla_{\distParams} \log
\dist(\vx; \distParams); \qquad s^{(k)}(\vx) = \frac{\nabla_{\distParams}^{(k)}
	\dist(\vx; \distParams)}{\dist(\vx; \distParams)},
\label{eq:higher_order_score}
\end{equation}
where $\nabla^{(k)}$ represents the $k$th-order gradient operator; unlike the first-order score, higher-order score functions are not defined in terms of higher derivatives of a log-probability. Using this
definition, the higher-order score-function gradient estimator is
\begin{equation} \veta^{(k)} =
\nabla^{(k)}_\distParams \expect{\dist(\vx; \distParams)}{\cost(\vx)} = \expect{\dist(\vx; \distParams)}{f(\vx)s^{(k)}(\vx)}.
\label{eq:higher_order_score_estimator}
\end{equation}
This simple mathematical shorthand hides the complexity of computing higher-order gradients. \citet{foerster2018dice} explore this further, and we expand on this discussion in Section \ref{sect:discussion} on computational graphs.
\subsubsection{Computational Considerations}
We can express the score-function gradient estimator (for a single parameter) in one other way, as
\begin{eqnarray}
	& \eta =
	\nabla_{\theta}\expect{\dist(\vx; \theta)}{\cost(\vx)} =
	\textrm{Cov}[\cost(\vx), \nabla_{\theta} \log \dist(\vx; \theta)], \\
	& \textrm{Cov}[\cost(\vx), \nabla_{\theta} \log
	\dist(\vx; \theta)]^2 \leq \mathbb{V}_{p(\vx; \theta)}[\cost(\vx)] \mathbb{V}_{p(\vx;
		\theta)}[\nabla_{\theta}
	\log
	\dist(\vx; \theta)].
\end{eqnarray}
The first identity shows that the score function gradient can be interpreted as
a measure of covariance between the cost function and the score function (and is
true because the expectation of the score is zero, which will remove the second
term that the covariance would introduce, \citet{pflug1996optimization} pp.
234). The second identity is the Cauchy-Schwartz
inequality, which bounds the squared covariance. Computationally, this shows
that the variance of the cost function is related to the magnitude and range of
the gradient. A highly-variable cost function can result in highly-variable
gradients, which is undesirable for optimisation. It is for this reason that we
will often constrain the values of the cost function by normalising or bounding
its value in some way, e.g., by clipping.

The score-function gradient is considered to be general-purpose because it is
computed using only the final value of the \gradCost in its computation. It
makes no assumptions about the internal structure of the cost function, and can
therefore be used with any function whose outcome we can simulate; many functions are then
open to us for use: known differentiable functions, discrete functions,
dynamical systems, or black-box simulators of physical/complex systems or graphics
engines. Overall the computational cost of the score function estimator is low;
it is of the order $\mathcal{O}(N(D + L))$ for $D$-dimensional
distributional parameters $\vtheta$, where $L$ is the cost of evaluating the
cost function, and $N$ is the number of samples used in the estimator.

Taking into account the exposition of this section, points for consideration
when using the score function estimator are:
\vspace{-4mm}
\begin{itemize}[noitemsep]
	\item \textit{Any type of cost function can be used}, allowing for the use of simulators and other
	black-box systems, as long as we are able to evaluate them easily.
	\item  The \textit{measure must be differentiable }with respect to its
	parameters, an assumption we make throughout this paper.
	\item We must be able to \textit{easily sample from the measure}, since the gradient estimator is computed
	using samples from it.
	 \item It is applicable to\textit{ both discrete and continuous distributions}, which adds to its generality.
	\item  The estimator can be implemented using only
	a \textit{single sample if needed}, i.e.~using $N=1$. Single-sample estimation is
	applicable in both univariate and multivariate cases, making the estimator
	computationally efficient, since we can deliberately control the number of times
	that the cost function is evaluated.
	\item Because there are many factors that affect the variance of the gradient estimator, it will be important to use some form of \textit{variance reduction} to obtain
	competitive performance.
\end{itemize}

\subsection{Research in Score Function Gradient Estimation}
We are the lucky inheritors of a rich body of work specifically devoted to the
score function gradient estimator. Given its simplicity and generality, this
estimator has found its way to many parts of computational science, especially
within operations research, computational finance, and machine learning. We
describe a subset of this existing work, focusing on the papers that provide deeper
theoretical insight and further context on the score-function estimator's use in
practice.

\paragraph{Development of the estimator.}
The score function estimator is one of
the first types of estimators to be derived, initially by several different
groups in the 1960s, including \citet{miller1967monte} in the design of stable
nuclear reactors, and \citet{rubinstein1969some} in the early development of
Monte Carlo optimisation. Later the estimator was derived and applied by
\citet{rubinstein1983one} for discrete-event systems in operations research,
where they began to refer to the estimator as the score function method, the
name we use in this paper. The estimator was again developed by
\citet{glynn1987likelilood, glynn1990likelihood} and by
\citet{reiman1989sensitivity}, where they referred to it as the likelihood ratio
method, and part of the study of queueing systems and regenerative stochastic
processes. Concise descriptions are available in several books, such as
those by \citet{rubinstein1993discrete} and \citet{fu2012conditional}, as well as in
two books we especially recommend by \citet[sect. 4.2.1]{pflug1996optimization}
and \citet[sect 7.3]{glasserman2013monte}.

\paragraph{Interchange of integration and differentiation.}
The two basic questions for the application of the score-function estimator are
differentiability of the stochastic system that is being studied, and
subsequently, the validity of the interchange of integration and
differentiation. For many simple systems, differentiability can be safely
assumed, but in discrete-event systems that are often studied in operations
research, some effort is needed to establish differentiability, e.g., like that
of the stochastic recursions by \citet{glynn1995likelihood}. The validity of the
interchange of differentiation and integration for score function estimators is
discussed by \citet{kleijnen1996optimization} and specifically in the note by
\citet{lecuyer1995note}.

\paragraph{In machine learning.}
In reinforcement learning, the score-function gradient estimator was developed as
the REINFORCE algorithm by \citet{williams1992simple}. In such settings, the
gradient is the fundamental quantity needed for policy search methods and is the
basis of the policy gradient theorem and the subsequent development of
actor-critic reinforcement learning methods \citep{sutton2000policy}.
Approximate Bayesian inference methods based on variational inference deployed
the score function gradient estimator to enable a more general-purpose,
black-box variational inference; one that did not require tedious manual
derivations, but that could instead be more easily combined with automatic
differentiation tools. The appeal and success of this approach has been shown by several
authors, including \citet{paisley2012variational}, \citet{wingate2013automated},
\citet{ranganath2014black}, and \citet{mnih2014neural}. Variance reduction is
essential for effective use of this estimator, and has been explored in several
areas, by \citet{greensmith2004variance} in reinforcement learning,
\citet{titsias2015local} in variational inference,
\citet{capriotti2008reducing} in computational finance, and more recently by
\citet{walder2019new}. We describe variance reduction techniques in more detail
in Section \ref{sect:var_reduction}. Finally, as machine learning has sought to
automate the computation of these gradients, the implementation of the score
function estimator within wider stochastic computational graphs has been
explored for the standard estimator \citep{schulman2015gradient} and its
higher-order counterparts \citep{foerster2018dice}; we will provide more discussion
on computational graphs in Section \ref{sect:discussion}.

\section{Pathwise Gradient Estimators} \label{sect:pathwise_estimators}
We can develop a very different type of gradient estimator if, instead of relying on the
knowledge of the score function, we take advantage of the structural characteristics
of the problem \eqref{eq:gradient}. One such structural property is the specific
sequence of transformations and operations that are applied to the sources of randomness
as they pass through the measure and into the cost function, to
affect the overall objective. Using this sampling path will lead us to a second
estimator, the pathwise gradient estimator, which as its name implies, is in the
class of derivatives of paths. Because we need information about the path
underlying a probabilistic objective, this class of gradient estimator will be less
general-purpose than the score-function estimator, but in losing this
generality, we will gain several advantages, especially in terms of lower
variance and ease of implementation.

The pathwise derivative is as fundamental to sensitivity analysis and stochastic
optimisation as the score function estimator is, and hence also appears under
several names, including: the process derivative \citep{pflug1996optimization}, as
the general area of perturbation analysis and specifically infinitesimal perturbation
analysis \citep{ho2012perturbation, glasserman1991gradient}, the
pathwise derivative \citep{glasserman2013monte}, and more recently as the reparameterisation trick 
and stochastic backpropagation \citep{rezende2014stochastic,
	kingma2014auto, titsias2014doubly}. Another way to view the pathwise approach is
as a process of pushing the parameters of interest, which are part of the
measure, into the cost function, and then differentiating the newly-modified
cost function. For this reason, the pathwise estimator is also called
the `push-in' gradient method \citep{rubinstein1992sensitivity}.

\subsection{Sampling Paths} 
Continuous distributions have a simulation property that allows both a direct and
an indirect way of drawing samples from them, making the  following sampling
processes equivalent:
\begin{equation} \hat{\vx} \sim \dist(\vx; \distParams) \quad \equiv \quad \hat{\vx}
= g(\hat\vepsilon, \distParams), \quad \hat\vepsilon \sim p(\vepsilon),
\end{equation} 
and states that an alternative way to generate samples $\hat{\vx}$ from the
distribution $\dist(\vx; \distParams)$ is to sample first from a simpler base
distribution $p(\vepsilon)$, which is independent of the parameters
$\distParams$, and to then transform this variate through a deterministic
\textit{path} $g(\vepsilon; \distParams)$; we can refer to this procedure as
either a sampling \textit{path} or sampling \textit{process}. For invertible paths, this
transformation is described by the rule for the change of variables for
probability
\begin{equation} \dist(\vx; \distParams)	= \dist(\vepsilon) \left| 
\nabla_{\vepsilon} g(\vepsilon; \distParams) \right|^{-1}.
\label{eqn:density_change_of_variable} 
\end{equation} 
There are several classes of transformation methods available \citep{devroye2006nonuniform}:
\vspace{-4mm}
\begin{itemize}[noitemsep] 
	\item \textbf{Inversion methods.} For univariate distributions, we will always
be able to find an equivalent base distribution and sampling path by using the
uniform distribution and inverse cumulative distribution function (CDF),
respectively. This method can often be difficult to use directly, however, since
computing the inverse of the CDF and its derivative can be computationally
difficult, restricting this approach to univariate settings. We will return
to this method later, but instead explore methods that allow us to use the CDF
instead of its inverse.
	
	\item \textbf{Polar transformations.} It is sometimes possible, and
	more efficient, to  generate a pair of random variates $(y, z)$ from the target
	distribution $p(x)$. We can map this pair to a representation in polar form $(r
	\cos \theta,r \sin \theta)$, which exposes other mechanisms for sampling, e.g., 
	the famous Box-Muller transform for sampling Gaussian variates is
	derived in this way. 
	
	\item \textbf{One-liners.} In many cases there are simple
	functions that transform a base distribution into a richer form. One
	widely-known example is sampling from the multivariate Gaussian  $p(\vx;
	\distParams) = \mathcal{N}(\vx | \vmu, \vSigma)$, by first sampling from the %
	standard Gaussian $p(\vepsilon)=\mathcal{N}(\vzero,\vI)$, and then applying
	the location-scale transformation $g(\vepsilon, \distParams) = \vmu +
	\vL\vepsilon$, with $\vL\vL^\top = \vSigma$. Many such transformations
	exist for common distributions, including the Dirichlet, Gamma, and
	Exponential. \citet{devroye1996random} refers  to these types of
	transformations as \textit{one-liners} because they can often be implemented in
	one line of code. 
\end{itemize} 
With knowledge of these transformation methods, we can invoke the
\textit{Law of the Unconscious Statistician} (LOTUS)
\citep{grimmett2001probability} 
\begin{equation} \expect{p(\vx;\vtheta)}{f(\vx)} =
\expect{p(\vepsilon)}{f(g(\vepsilon; \vtheta))},
\label{eq:pathwise_lotus}
\end{equation} 
which states that we can compute the expectation of a function of a random
variable $\vx$ without knowing its distribution, if we know its corresponding
sampling path and base distribution. LOTUS  tells us that in probabilistic
objectives, we can replace expectations over any random variables $\vx$
wherever they appear, by the transformation $g(\vepsilon; \distParams)$ and
expectations over the base distribution $p(\vepsilon)$. This is a way to
reparameterise a probabilistic system; it is often used in Monte Carlo methods,
where it is referred to as the \textit{non-centred parameterisation}
\citep{papaspiliopoulos2007general} or as the \textit{reparameterisation trick}
\citep{kingma2014auto}.

\subsection{Deriving the Estimator} 
Equipped with the pathwise simulation property of continuous distributions and LOTUS, we
can derive an alternative estimator for the sensitivity analysis problem
\eqref{eq:gradient} that exploits this additional knowledge of continuous
distributions. Assume that we have a distribution $\dist(\vx;
\vtheta)$ with known \textit{differentiable}
sampling path $g(\vepsilon; \distParams)$ and base
distribution $p(\vepsilon)$. The sensitivity analysis problem \eqref{eq:gradient} can then be
reformulated as
\begin{subequations} \begin{align} 
	\veta &= \nabla_\distParams
	\expect{\dist(\vx; \distParams)}{\cost(\vx)} 
	 = \nabla_\distParams \int{\dist(\vx; \distParams)\cost(\vx) d\vx}
	\label{eq:pathwise_grad_problem} \\ 
	& = \nabla_\distParams \int p(\vepsilon)
	\cost(g(\vepsilon; \distParams)) d\vepsilon \label{eq:pathwise_lotus_apply} \\ %
	& = \expect{p(\vepsilon)}{\nabla_\distParams \cost(g(\vepsilon; \distParams))} 
	\label{eq:pathwise_gradient}.\\ 
	\bar{\veta}_N & =
	\frac{1}{N}\sum_{n=1}^{N} \nabla_\distParams \cost(g(\hat\vepsilon^{(n)};
	\distParams)); \quad \hat\vepsilon^{(n)} \sim p(\vepsilon). 
	\label{eq:pathwise_estimator} 
\end{align} \end{subequations} 
In Equation \eqref{eq:pathwise_grad_problem} we first expand the definition of
the expectation. Then, using the law of the unconscious statistician, and knowledge of
the sampling path $g$ and the base distribution for $p(\vx; \distParams)$, we
reparameterise this integral \eqref{eq:pathwise_lotus_apply} as one over the variable
$\vepsilon$. The parameters $\distParams $ have now been \textit{pushed} into
the function making the expectation free of the parameters. This allows us to,
without concern, interchange the derivative and the integral
\eqref{eq:pathwise_gradient}, resulting in the pathwise gradient
estimator \eqref{eq:pathwise_estimator}. %

\subsection{Estimator Properties and Applicability}
The gradient \eqref{eq:pathwise_gradient} shows that we can compute gradients of
expectations by first pushing the parameters into the cost function and then
using standard differentiation, applying the chain rule. This is a natural way
to think about these gradients, since it aligns with our usual understanding of
how deterministic gradients are computed, and is the reason this approach is so
popular. Since the sampling path $g$ need not always be invertible,
the applicability of the estimator goes beyond the change of variable formula for
probabilities~\eqref{eqn:density_change_of_variable}.
Its simplicity, however, belies several distinct properties that impact its use
in practice.

\subsubsection{Decoupling Sampling and Gradient Computation} 
\label{sect:pathwise_decoupling}
The pathwise estimator \eqref{eq:pathwise_gradient}, as we derived it, is
limited to those distributions for which we simultaneously have a differentiable
path, and use this same path to generate samples. We could not compute gradients
with respect to parameters of a Gamma distribution in this way, because sampling
from a Gamma distribution involves rejection sampling which does not provide a
differentiable path. The process of sampling from a distribution and the process
of computing gradients with respect to its parameters are coupled in the
estimator \eqref{eq:pathwise_gradient}; we can expand the applicability of the
pathwise gradient by decoupling these two processes.

The pathwise estimator can be rewritten in a more general form:
\begin{subequations} \begin{align} 
	\veta = \nabla_\distParams
	\expect{\dist(\vx; \distParams)}{\cost(\vx)} & =
	\expect{p(\vepsilon)}{\nabla_\distParams \cost(\vx)|_{\vx = g(\vepsilon;
		\distParams)}}
	= \int p(\vepsilon) \nabla_\vx \cost(\vx)|_{\vx = g(\vepsilon; \distParams)} \nabla_{\distParams} g(\vepsilon;\distParams) \, d\vepsilon
	\label{eq:pathwise_implicit_chain}\\
	& = \int
	\dist(\vx; \distParams) \nabla_\vx \cost(\vx) \nabla_{\distParams} \vx \, d\vx =
	\expect{\dist(\vx; \distParams)}{\nabla_\vx \cost(\vx) \nabla_{\distParams} \vx}.
	\label{eq:pathwise_implicit} %
	\end{align} \end{subequations}
In the first line \eqref{eq:pathwise_implicit_chain}, we recall the gradient
that was formed as a result of applying the law of the unconscious statistician
in Equation \eqref{eq:pathwise_gradient}, and then apply the chain rule to write
the gradient w.r.t.~$\distParams$. Using the LOTUS \eqref{eq:pathwise_lotus} again, 
this time for the expectation
of $\nabla_\distParams \cost(\vx)=\nabla_\vx \cost(\vx) \nabla_{\distParams} \vx$
and in the reverse direction, we obtain \eqref{eq:pathwise_implicit}.
This derivation shows that sampling
and gradient estimation can be decoupled: we can generate samples using
any sampler for the original distribution $p(\vx; \distParams)$ (e.g., using a sampling path, rejection 
sampling, or Markov chain Monte Carlo), and compute the gradient
using the chain rule on the function by finding some way to compute the term
$\nabla_\distParams \vx$.

One way to compute $\nabla_\distParams \vx$ is to use $\nabla_{\distParams}
g(\vepsilon; \distParams)$ as we did for \eqref{eq:pathwise_gradient}. In
practice, this form is not always convenient. For example, if the sampling path
is an inverse cumulative density function (inverse CDF), which is often computed with
root-finding methods, then evaluating its derivative may require numerically
unstable finite difference methods. Instead, we can find another way of writing
$\nabla_\distParams \vx$ that makes use of the inverse of the path $g^{-1}(\vx;
\distParams)$. We can think of $g^{-1}(\vx; \distParams)$ as the `standardisation path'
of the random variable---that is the 
transformation that removes the dependence of the sample on the distribution parameters, standardising it 
to a zero mean unit variance-like form. In the
univariate case, instead of using the inverse CDF, we can use the standardisation
path given by the CDF. Consider the equation $\vepsilon = g^{-1}(\vx;
\distParams)$ as an implicit function for $\vx$. Evaluating the total derivative (TD) on both sides---using
implicit differentiation---and expanding we find that 
\begin{subequations}
\begin{eqnarray} 
& \vepsilon = g^{-1}(\vx; \vtheta) \implies \nabla^{\mathrm{TD}}_\distParams \vepsilon =
\nabla^{\mathrm{TD}}_{\distParams} g^{-1}(\vx; \distParams) \\
& \therefore \vzero =
\nabla_\vx g^{-1}(\vx; \distParams) \nabla_{\distParams} \vx +
\nabla_\distParams g^{-1}(\vx; \distParams) \\ 
& \nabla_{\distParams} \vx = -(
\nabla_\vx g^{-1}(\vx; \distParams) )^{-1} \nabla_\distParams g^{-1}(\vx;
\distParams).
\label{eq:pathwise_implicit_gradx} 
\end{eqnarray} 
\end{subequations}
Equation \eqref{eq:pathwise_implicit_gradx} gives us the expression needed to fully evaluate 
\eqref{eq:pathwise_implicit}. Equation \eqref{eq:pathwise_implicit_gradx} is the
form in which \citet{ho1983optimization} initially introduced the estimator, and
it corresponds to the strategy of differentiating both sides of the transformation noted by
\citet{glasserman2013monte}. \citet{figurnov2018implicit} refer to this approach
to as an implicit reparameterisation gradient, because of its use of implicit
differentiation.  In this form, we are now able to apply pathwise gradient
estimation to a far wider set of distributions and paths, such as for the Beta,
Gamma, and Dirichlet distributions. In Section \ref{sect:case_studies} we look at
this decoupling in other settings,  and in the discussion (Section
\ref{sect:opt_transport}) look at optimal transport as another way of computing
$\nabla_{\vtheta}\vx$.

\begin{gradexample}{Univariate Gaussian} 
For univariate Gaussian distributions $\mathcal{N}(x | \mu,
	\sigma^2)$, with $\vtheta = \{\mu, \sigma\}$, the location-scale transform is the
	natural choice of the path: $x = g(\epsilon; \vtheta) = \mu + \sigma\epsilon$ for
	$\epsilon \sim \mathcal{N}(0,1)$. The inverse path is then $\epsilon = g^{-1}(x,
	\vtheta) = \frac{x-\mu}{\sigma}$. The standard pathwise derivatives are
	$\frac{dx}{d\mu} = 1$ and $\frac{dx}{d\sigma} = \epsilon$. Equation
	\eqref{eq:pathwise_implicit_gradx} is then 
	\begin{equation} \frac{dx}{d\mu} = -
	\frac{\sfrac{\partial g^{-1}(x, \theta)}{\partial\mu}}{\sfrac{\partial g^{-1}(x,
			\theta)}{\partial x}} = -\frac{-\sfrac{1}{\sigma}}{\sfrac{1}{\sigma}} = 1; \qquad
	\frac{dx}{d\sigma} = - \frac{\sfrac{\partial g^{-1}(x,
			\theta)}{\partial\sigma}}{\sfrac{\partial g^{-1}(x, \theta)}{\partial x}} =
	-\frac{-\sfrac{(x-\mu)}{\sigma^2}}{\sfrac{1}{\sigma}} = \frac{x-\mu}{\sigma} =
	\epsilon. \end{equation} 
	We see that the two approaches provide the same gradient. %
\end{gradexample}

\begin{gradexample}{Univariate distributions}
	As we discussed, for univariate distributions $p(x; \vtheta)$ we can use the sampling
	path given by the inverse CDF: $x = g(\epsilon; \vtheta) = F^{-1}(\epsilon; \vtheta)$, where
	$\epsilon \sim \mathcal{U}[0, 1]$. Computing the derivative
	$\nabla_{\vtheta} x = \nabla_{\vtheta} F^{-1}(\epsilon; \vtheta)$ is often complicated and expensive.
	We can obtain an alternative expression for $\nabla_{\vtheta} x$ by considering the
	inverse path, which is given by the CDF $g^{-1}(x; \vtheta) = F(x; \vtheta)$.
	From Equation \eqref{eq:pathwise_implicit_gradx} we have
	\begin{equation}
		\nabla_{\vtheta} x = -\frac{\nabla_{\vtheta} F(x; \vtheta)}{\nabla_x F(x; \vtheta)} =
		-\frac{\nabla_{\vtheta} F(x; \vtheta)}{p(x; \vtheta)}.
	\label{eq:pathwise_implicit_univariate}
	\end{equation}
	In the final step, we used the fact that the derivative of the CDF w.r.t.~$x$
	is the density function. This expression allows efficient computation of pathwise gradients
	for a wide range of continuous univariate distributions, including Gamma, Beta, von Mises,
	Student's $t$, as well as univariate mixtures \citep{figurnov2018implicit,jankowiak2018pathwise}.
\end{gradexample}

\subsubsection{Bias and Variance Properties}
In deriving the pathwise estimator \eqref{eq:pathwise_estimator} we again
exploited an interchange of differentiation and integration. If this interchange
is valid, then the resulting application of the chain rule to compute the
gradient will be valid as well, and the resulting estimator will be unbiased. We can
always ensure that this interchange applies by ensuring that the cost functions
we use are differentiable. The implication of this is that we will be unable to use
the pathwise gradient estimator for discontinuous cost functions. We can be more rigorous about the conditions for
unbiasedness, and we defer to  \citet[sect. 7.2.2]{glasserman2013monte} for this
more in-depth discussion.

The variance of the pathwise estimator can be shown to be bounded by the squared Lipschitz
constant of the cost function \citep[sect. 7.2.2]{glasserman2013monte}. This result is also shown by \citet[sect. 10]{fan2015fast} for the case of gradients \eqref{eq:gradient} with Gaussian measures, instead using the properties of sub-Gaussian random variables, deriving a dimension-free bound for the variance in terms of the squared Lipschitz constant. This insight has two important implications for the use of the pathwise gradient. Firstly, the variance bounds that exist are
independent of the dimensionality of the parameter space, meaning that we can
expect to get low-variance gradient estimates, even in high-dimensional
settings. Because we are able to differentiate through the cost function itself,
only the specific path through which any individual parameter influences the
cost function is included in the gradient (automatic provenance tracking); unlike the score-function estimator,
the gradient does not sum over terms for which the parameter
has no effect, allowing the estimator variance to be much lower.
Secondly, as Figures \ref{fig:grad_var_quadratic} and \ref{fig:grad_var_exp_cos} show, as
the cost function becomes highly-variable, i.e.~its Lipschitz constant
increases, we enter regimes where, even with the elementary functions we
considered, the variance of the pathwise estimator can be higher than that of
the score-function method. The pathwise gradient, when it is applicable, will
not always have lower variance when compared to other methods since its variance
is directly bounded by the cost function's Lipschitz constant. In such
situations, variance reduction will again be a powerful accompaniment to the
pathwise estimator. Since most of the functions we will work with will not be
Lipschitz continuous, \citet{xu2018variance} use a different set of simplifying
assumptions to develop an understanding of the variance of the pathwise estimator that reinforces the
general intuition built here.

\subsubsection{Higher-order Gradients} 
Computation of higher-order gradients using the pathwise method is also possible
and directly involves higher-order differentiation of the function, requiring cost
functions that are differentiable at higher-orders.
This approach has been explored by
\citet{fu1993second} and \citet{fan2015fast}. Another common strategy for
computing higher-order gradients is to compute the first-order gradient using
the pathwise methods, and the higher-order gradients using the score-function method.

\subsubsection{Computational Considerations}
The pathwise gradient estimator is restricted to differentiable cost functions.
While this is a large class of functions, this limitation makes the pathwise estimator
less broadly applicable than the score-function estimator. For some types of discontinuous
functions it is possible to smooth the function over the discontinuity and
maintain the correctness of the gradient; this approach is often referred to as
smoothed perturbation analysis in the existing literature
\citep{glasserman1991gradient}.

Often there are several competing approaches for computing the gradient.
The choice between them will need to be made considering the associated
computational costs and ease of implementation. The case of the univariate Gaussian
measure $\mathcal{N}(x | \mu, \sigma^2)$ highlights this. This distribution has
two equivalent sampling paths: one given by the location-scale transform:
$x = \mu + \sigma \epsilon_1,\ \epsilon_1 \sim \mathcal{N}(0, 1)$ and another
given by the inverse CDF transform:
$x = F^{-1}(\epsilon_2; \mu, \sigma),\ \epsilon_2 \sim \mathcal{U}[0, 1]$.
While there is no theoretical difference between the two paths, the first path
is preferred in practice due to the simplicity of implementation.

The same analysis we pointed to relating to  the variance of the
Gaussian gradient earlier, also provides a
tail bound with which to understand the convergence of the Monte Carlo estimator
\citep{fan2015fast}. Rapid convergence can be obtained even
when using only a single sample to compute the gradient, as is often done
in practice. There is a trade-off between the number of samples used and the Lipschitz constant of the cost function, and may require  more samples to be used for functions with higher Lipschitz constants.
This consideration is why we will find that regularisation that promotes
smoothness of the functions we learn is important for successful
applications. Overall, the computational cost of the pathwise estimator is the
same as the score function estimator and is low, of the order
$\mathcal{O}(N(D + L))$, where $D$ is the dimensionality of the distributional parameters
$\vtheta$,  $L$ is the cost of evaluating the cost function and its gradient, and
$N$ is the number of samples used in the estimator.

Taking into account the exposition of this section, points for consideration
when using the pathwise derivative estimator are:
\vspace{-4mm}
\begin{itemize}[noitemsep]
	\item Only cost functions that are \textit{differentiable} can be used.
	\item When using the pathwise estimator \eqref{eq:pathwise_gradient}, we do \textit{not} need to know 
	the measure explicitly. Instead we must know its corresponding \textit{deterministic and differentiable 
	sampling path and a base distribution} that is easy to sample from.
	\item If using the implicit form of the estimator, we require a \textit{method of sampling from the original 
	distribution} as well as a way of computing the derivative of the inverse (standardisation) path.
	\item The estimator can be implemented using only a \textit{single sample if needed}, i.e.~using $N=1$. Single-sample estimation is
	applicable in both univariate and multivariate cases, making the estimator
	computationally efficient, since we can deliberately control the number of times
	that the cost function is evaluated. 
	\item We might need to \textit{control the smoothness} of the function during learning to avoid large variance, and may need to employ variance reduction.
	\end{itemize}

\subsection{Research in Pathwise Derivative Estimation}
\paragraph{Basic development.} 
This pathwise estimator was initially developed by \citet{ho1983optimization}
under the name of infinitesimal perturbation analysis (IPA), by which it is still
commonly known. Its convergence properties were analysed by
\citet{heidelberger1988convergence}. Other variations of
perturbation analysis expand its applicability, see
\citet{suri1988perturbation} and the books by \citet{glasserman1991gradient}
and \citet{ho2012perturbation}. The view as a push-in technique was expanded on
by \citet{rubinstein1992sensitivity}. Again, the books by
\citet[sect. 7.2]{glasserman2013monte} and \citet[sect.
4.2.3]{pflug1996optimization} provide a patient exploration of this method using
the names of the pathwise derivative and process derivative estimation,
respectively.

\paragraph{Appearance in machine learning.} 
The pathwise approach to gradient estimation has seen wide application in
machine learning, driven by work in variational inference. An early instance of
gradients of Gaussian expectations in machine learning was developed by
\citet{opper2009variational}. \citet{rezende2014stochastic} called their use of
the pathwise estimator stochastic backpropagation, since the gradient reduced to
the standard gradient computation by backpropagation averaged over an
independent source of noise. As we mentioned in Section
\ref{sect:stoch_optim_subsect}, \citet{titsias2014doubly} used the pathwise
estimator under the umbrella of doubly stochastic optimisation, to recognise
that there are two distinct sources of stochasticity that enter when using the
pathwise estimator with mini-batch optimisation. The substitution of the random
variable by the path in the estimator \eqref{eq:pathwise_estimator} led
\citet{kingma2014efficient, kingma2014auto} to refer to their use of the substitution as the
reparameterisation trick, and by which it is commonly referred to at present in
machine learning.

\paragraph{Use in generative models and reinforcement learning.}
Recognising that the pathwise gradients do not require knowledge of the final
density, but only a sampling path and base distribution, \citet{rezende2015variational} 
developed normalising flows for variational
inference, which allows learning of complex probability distributions that exploit
this property of the pathwise estimator.
Since the pathwise estimator does not require invertible sampling paths,
it is very suited for uses with very expressive but not invertible function
approximators, such as deep neural networks.
This has been used to optimise model parameters in implicit generative models, such those in generative
adversarial networks \citep{goodfellow2014generative, mohamed2016learning}.
To learn behaviour
policies in continuous action spaces, the pathwise estimator has also found
numerous uses in reinforcement learning for continuous control.
\citet[sect. 7.2]{williams1992simple} invoked this approach as an alternative to the score function 
when 
using
Gaussian distributions, and has also been used for learning value gradients by
\citet{heess2015learning} and \citet{lillicrap2015continuous}. We find that the
pathwise derivative is now a key tool for computing information-theoretic
quantities like the mutual information \citep{alemi2016deep}, in Bayesian
optimisation \citep{wilson2017reparameterization} and in probabilistic
programming \citep{ritchie2016deep}.

\paragraph{Derivative generalisations.} 
 \citet{gong1987smoothed} introduced smoothed perturbation analysis as one
variant of the pathwise derivative to address cases where the interchange of
differentiation and integration is not applicable, such as when there are
discontinuities in the cost function; and several other extensions of the
estimator appear in the perturbation analysis literature
\citep{glasserman1991gradient}. In the variational inference setting,
\citet{lee2018reparameterization} also look at the non-differentiable case by
splitting regions into differentiable and non-differentiable components. The
implicit reparameterisation gradients for univariate distributions were
developed by \citet{salimans2013fixed}. \citet{hoffman2015stochastic} used the
implicit gradients to perform backpropagation through the Gamma distribution
using a finite difference approximation of the CDF derivative.
\citet{graves2016stochastic} derived the implicit reparameterisation gradients
for multivariate distributions with analytically tractable CDFs, such as
mixtures. The form of implicit reparameterisation that we described here was
developed by \citet{figurnov2018implicit} and clarifies the connections to
existing work and provides further practical insight. Other generalisations
like that of \citet{ruiz2016generalized},
\citet{naesseth2017reparameterization}, and \citet{parmas18a}, combine the score
function method with the pathwise methods; we come back to these hybrid methods
in Section \ref{sect:case_studies}.

\section{Measure-valued Gradients}
\label{sect:weak_deriv_estimators}
A third class of gradient estimators, which falls within the class of derivatives of measure, is known simply
as the measure-valued gradient estimators, and has received little attention in
machine learning. By exploiting the underlying measure-theoretic properties of
the probabilities in the sensitivity analysis problem \eqref{eq:gradient}---the properties
of signed-measures in particular---we will obtain a class of unbiased and
general-purpose gradient estimators with favourable variance properties. This
approach is referred to interchangeably as either the weak derivative method
\citep{pflug1996optimization} or as the measure-valued derivative
\citep{heidergott2000measure}.

\subsection{Weak Derivatives}
Consider the derivative of a density $p(\vx; \vtheta)$ with respect to a single parameter $\theta_i$, with $i$ the index on the set of distributional parameters. The derivative $\nabla_{\theta_i} p(\vx; \vtheta)$ is itself not a
density, since it may have negative values and does not integrate to one. However, using the properties of signed measures, we
can always decompose this derivative into a difference of two densities, each multiplied
by a constant \citep{pflug1989sampling}:
\begin{equation}
\nabla_{\theta_i} p(\vx; \vtheta) = c^{+}_{\theta_i} \ppos(\vx; \vtheta) - c^{-}_{\theta_i} \pneg(\vx; \vtheta),
\label{eq:weak_deriv_two_cs}
\end{equation}
where $\ppos, \pneg$ are densities, referred to as the positive and negative
components of $p$, respectively. By integrating both sides of
\eqref{eq:weak_deriv_two_cs}, we can see that $c^{+}_\theta = c^{-}_\theta$:
\begin{equation}
\textrm{LHS:} \int \!\nabla_{\theta_i} p(\vx; \vtheta) \intd \vx = \nabla_{\theta_i}\! \int\! \! p(\vx; \vtheta) \intd\vx = 0; \!\!\quad
\textrm{RHS:} \! \int\!\! \left(c^{+}_{\theta_i} \ppos(\vx; \vtheta) - c^{-}_{\theta_i} \pneg(\vx;
\vtheta)\right) \!\intd\vx =
c^{+}_{\theta_i} \!- c^{-}_{\theta_i}.
\nonumber
\end{equation}
Therefore, we can simply use one constant that we denote $c_{\theta_i}$, and the decomposition
becomes
\begin{equation}
\nabla_{\theta_i} p(\vx; \vtheta) = c_{\theta_i} \left(\ppos(\vx; \vtheta) - \pneg(\vx; \vtheta)\right).
\label{eq:weak_deriv}
\end{equation}
The triple $\left(c_{\theta_i}, \ppos, \pneg \right)$ is referred to as  the ($i$th)
\textit{weak derivative} of $p(\vx; \vtheta)$. We restrict our definition here to
the univariate parameter case; the multivariate parameter case extends this definition to
form a vector of triples, with one triple for each dimension. We list the weak
derivatives for several widely-used distributions in Table
\ref{tab:mvd_triples}, and defer to \citet{heidergott2003measure} for
their derivations.
\begin{table}[tb]
	\centering
	\begin{tabular}{@{}lcll@{}}
		\toprule
		Distribution $\dist(x; \theta)$       & Constant $c_\theta$            & Positive part
		$\ppos(x)$           &
		Negative part $\pneg(x)$          \\ \midrule
		Bernoulli($\theta$)  & 1                              & $\delta_1$                            &
		$\delta_0$                            \\
		Poisson($\theta$)                & 1                              & $\mathcal{P}(\theta) + 1$             &
		$\mathcal{P}(\theta)$                 \\
		Normal($\theta, \sigma^2$)       & $\sfrac{1}{\sigma\sqrt{2\pi}}$ & $\theta + \sigma
		\mathcal{W}(2,
		0.5)$ & $\theta - \sigma \mathcal{W}(2, 0.5)$ \\
		Normal($\mu, \theta^2$)          & $\sfrac{1}{\theta}$            & $\mathcal{M}(\mu,
		\theta^2)$            &
		$\mathcal{N}(\mu, \theta^2)$            \\
		Exponential($\theta$)            & $\sfrac{1}{\theta}$            &
		$\mathcal{E}(\theta)$                 &
		$\theta^{-1}\mathcal{E}r(2)$          \\
		Gamma($a, \theta$)               & $\sfrac{a}{\theta}$            & $\mathcal{G}(a,
		\theta)$              &
		$\mathcal{G}(a+1, \theta)$            \\
		Weibull($\alpha, \theta$)        & $\sfrac{1}{\theta}$            & $\mathcal{W}(\alpha,
		\theta)$         &
		$\mathcal{G}(2, \theta)^{1/\alpha}$   \\ \bottomrule
	\end{tabular}
\caption{Weak derivative triples $\left(c_\theta, \ppos, \pneg \right)$ for the parameters of
	common distributions; we use $\mathcal{N}$ for the Gaussian density,
	$\mathcal{W}$ for the Weibull, $\mathcal{G}$ for the Gamma, $\mathcal{E}$ for
	the exponential, $\mathcal{E}r$ for the Erlang, $\mathcal{M}$ the double-sided
	Maxwell, and $\mathcal{P}$ for the Poisson. We always use $\theta$ to denote the parameter the derivative triple applies to. See Appendix \ref{app:distribution} for the forms of these
	distributions.}
	\label{tab:mvd_triples}
\end{table}

The derivative is weak because we do not require
the density $p$ to be differentiable on its domain, but rather require that
integrals of the decomposition $\ppos, \pneg$ against sets of test
functions converge. For example, the decomposition of mass functions of discrete distributions
results in positive and
negative components that are delta functions, which are integrable against
continuous functions. The weak derivative is not unique, but always exists and
can be obtained using the Hahn-Jordan decomposition of a signed measure into
two measures that have complementary support \citep{billingsley2008probability}.
Like the score function and the sampling path, the weak derivative is a tool we will use to study the
sensitivity analysis problem \eqref{eq:gradient}, although it has uses in other settings. The
connections between the weak derivative and functional analysis, as well as a
general calculus for weak differentiability is described by
\citet{heidergott2010weak}.

\subsection{Deriving the Estimator}
Using our knowledge of weak derivatives of probability measures, we can now
derive a third estimator for the gradient problem \eqref{eq:gradient}.  This
will lead us to an unbiased gradient estimator, which intuitively computes the
gradient using a weighted difference of two expectations. For $D$-dimensional
parameters $\vtheta$, we can rewrite the gradient for the $i$th parameter
$\theta_i$ as
\begin{subequations}
\begin{align}
\eta_i & = \nabla_{\theta_i} \expect{p(\vx; \vtheta)}{f(\vx)}  = \nabla_{\theta_i} \int p(\vx; \vtheta) f(\vx)
\intd\vx = \int \nabla_{\theta_i} p(\vx; \vtheta) f(\vx) \intd\vx
\label{eq:wd_interchange} \\
& = c_{\theta_i} \left(\int \cost(\vx) \ppos_{i}(\vx; \vtheta) \intd\vx - \int \cost(\vx) \pneg_{i}(\vx;
\vtheta)
\intd\vx \right) \label{eq:wd_decomposition}\\
& = c_{\theta_i} \left( \expect{\ppos_{i}(\vx; \vtheta)}{\cost(\vx)} - \expect{\pneg_{i}(\vx;
\vtheta)}{\cost(\vx)} \right) \label{eq:wd_deriv}. \\
\bar{\eta}_{i,N} &= \frac{c_{\theta_i}}{N} \left(\sum_{n=1}^N f(\dot{\vx}^{(n)}) - \sum_{n=1}^N
f(\ddot{\vx}^{(n)})\right);
\quad \dot{\vx}^{(n)} \sim  \ppos_{i}(\vx; \vtheta),\quad \ddot{\vx}^{(n)} \sim  \pneg_{i}(\vx;
\vtheta).
\label{eq:wd_estimator}
\end{align}
\end{subequations}
In the first line \eqref{eq:wd_interchange} we expanded the expectation as an
integral and then interchanged the order of differentiation and integration. In
the second line \eqref{eq:wd_decomposition}, we substituted the
density-derivative with its weak derivative representation
\eqref{eq:weak_deriv}.  We use the symbols $\ppos_{i}, \pneg_{i}$ to remind
ourselves that the positive and negative components may be different depending
on which parameter of the measure the derivative is taken with respect to, and
$c_{\theta_i}$ to indicate that the constant will also change depending on the
parameter being differentiated. We write the gradient using the expectation
operators we have used throughout the paper \eqref{eq:wd_deriv}, and finally
obtain an estimator in Equation \eqref{eq:wd_estimator}.

The triple $\left(c_{\theta_i}, \ppos_{i}(\vx; \vtheta), \pneg_{i}(\vx;
\vtheta)\right)$ is the measure-valued gradient of \eqref{eq:gradient}
\citep{pflug1989sampling, heidergott2003measure}. We now have a third unbiased
estimator of the gradient that applies to any type of \gradCost functions $f$,
differentiable or not, since no knowledge of the function is used other than our
ability to evaluate it for different inputs.

\begin{gradexample}{Bernoulli measure-valued gradient}
We can illustrate the generality of the measure-valued derivatives by
considering the case where $p(x)$ is the Bernoulli distribution.
	\begin{equation}
	\nabla_\theta \int \dist(x; \theta) \cost(x) dx = \nabla_\theta (\theta \cost(1) +
	(1-\theta)\cost(0)) = \cost(1)
	- \cost(0).
	\end{equation}
	The weak derivative is given by the triple $(1, \delta_1, \delta_0)$, where
	$\delta_x$ denotes the Dirac measure at $x$. This example is also interesting, since in this case, the measure-valued estimator is the same as the estimator that would be derived using the score-function approach.
\end{gradexample}

\begin{gradexample}{Gaussian measure-valued gradient}
We now derive the measure-valued derivative for the probabilistic objective \eqref{eq:expectation_function} w.r.t.~the
mean parameter of a Gaussian measure; we will do this from first principles, rather than relying on the result in Table
\ref{tab:mvd_triples}. We will make use of the change of variables $x = \sigma y + \mu$,
with differentials $\sigma \intd y = \intd x$ in this derivation.
\begin{subequations}
\begin{align}
\eta_{\mu}   &=	\nabla_\mu \expect{\mathcal{N}(x; \mu, \sigma^2)}{f(x)} =  \int \nabla_\mu
\mathcal{N}(x;
	\mu, \sigma^2) \cost(x) \intd x \label{eq:mvd_gauss_mean_deriv_ex}\\
	&=  \int f(x) \tfrac{1}{\sigma\sqrt{2\pi}}\tfrac{1}{\sigma} \exp\left(-\tfrac{1}{2}\left(\tfrac{x -
		\mu}{\sigma} \right)^2
	\right) \left( \mathds{1}_{\{x \geq \mu\}} \tfrac{x - \mu}{\sigma} - \mathds{1}_{\{x < \mu\}}
	\tfrac{ \mu -x}{\sigma} \right) \intd x \label{eq:mvd_ex_gaussian_deriv}\\
	&= \tfrac{1}{\sigma\sqrt{2\pi}} \left( \int_{\mu}^{\infty} f(x) \tfrac{1}{\sigma}
	\exp\left(-\tfrac{1}{2}\left(\tfrac{x -
		\mu}{\sigma} \right)^2 \right) \tfrac{x - \mu}{\sigma} \intd x \right.
 \left. - \int_{-\infty}^{\mu} f(x) \tfrac{1}{\sigma} \exp\left(-\tfrac{1}{2}\left(\tfrac{\mu-x}{\sigma}
	\right)^2 \right) \tfrac{\mu - x}{\sigma} \intd x \right) \\
	&= \tfrac{1}{\sigma\sqrt{2\pi}} \left( \int_{\mu}^{\infty} f(x) \tfrac{1}{\sigma}
	\exp\left(-\tfrac{1}{2}\left(\tfrac{x -
		\mu}{\sigma} \right)^2 \right) \tfrac{x - \mu}{\sigma} \intd x \right.
	\left. - \int_{-\mu}^{\infty} f(-x) \tfrac{1}{\sigma} \exp\left(-\tfrac{1}{2}\left(\tfrac{\mu+x}{\sigma}
	\right)^2 \right) \tfrac{\mu + x}{\sigma} \intd x \right) \\
	&=  \tfrac{1}{\sigma\sqrt{2\pi}} \left( \int_{0}^{\infty} f(\mu + \sigma y)y \exp \left( -
	\tfrac{y^2}{2}\right)  \intd y -
	\int_0^\infty f(\mu - \sigma y)y \exp \left(- \tfrac{y^2}{2}\right) \intd y \right)
	\label{eq:mvd_ex_var_change}\\
	&= \tfrac{1}{\sigma\sqrt{2\pi}} \left( \expect{\mathcal{W}(y|2, 0.5)}{f(\mu + \sigma y)} -
	\expect{\mathcal{W}(y|2, 0.5)}{f(\mu - \sigma y)} \right).
	\label{eq:mvd_gaussian_weak_deriv}
	\end{align}
\end{subequations}
In the first line we exchange the order of integration and differentiation, and
in the second line \eqref{eq:mvd_ex_gaussian_deriv} differentiate the Gaussian density
with respect to its mean $\mu$.  The derivative is written in a form that decomposes the integral
around
the mean $\mu$, and in this way introduces the minus sign needed for the form of the weak
derivative. In the third line, we split the integral over the domains implied by the
decomposition. The fourth line changes the limits of the second integral to the domain of $(-\mu,
\infty)$, which is used in the next line \eqref{eq:mvd_ex_var_change} along with the change of
variables to rewrite the integral more compactly in terms of $y$. In the final line
\eqref{eq:mvd_gaussian_weak_deriv} we recognise the Weibull distribution with
zero-shift as the new measure under which the integral is evaluated and write the expression using
 the expectation notation we use throughout the paper. We give the
form of the Weibull and other distributions in Appendix \ref{app:distribution}.

Equation \eqref{eq:mvd_gaussian_weak_deriv} gives the
measure-valued derivative as the difference of two expectations under the
Weibull distribution weighted by a constant, and can be described by the triple and estimator
\begin{eqnarray}
&	\left(\tfrac{1}{\sigma\sqrt{2\pi}}, \mu + \sigma \mathcal{W}(2, \tfrac{1}{2}), \mu - \sigma
	\mathcal{W}(2, \tfrac{1}{2})\right); \nonumber \\
&	\bar{\eta}_{\mu,N} = \frac{1}{N\sigma\sqrt{2\pi}} \sum_{n=1}^{N} \left(f(\mu + \sigma\dot{y}^{(n)})-f(\mu -
	\sigma\ddot{y}^{(n)})\right); \quad \dot{y}^{(n)} \sim \mathcal{W}(2, \tfrac{1}{2}), \ddot{y}^{(n)} \sim \mathcal{W}(2,
	\tfrac{1}{2}).
	\label{eq:mvd_ex_estimator}
\end{eqnarray}
In Equation \eqref{eq:mvd_ex_estimator} the samples used for positive and negative parts
are independent. A simple variance reduction is possible by
\textit{coupling} the two Weibull distributions with a set of common random
numbers; we explore variance reduction by coupling
in more detail later.
\end{gradexample}

While we have restricted ourselves to scalar quantities in these derivations,
the same properties extend to the vector case, where the measure-valued
derivative is a vector of triples; as \citet{pflug1996optimization} says, `the
vector case is more complicated in notation, but not in concept'. In Equation
\eqref{eq:wd_interchange}, if the measure is a factorised distribution $p(\vx;
\vtheta) = \prod_{d} p(x_d | \theta_d)$ then the positive and negative
components of the weak derivative will itself factorise across dimensions. For
the positive component, this decomposition will be $\ppos_i(\vx; \vtheta) =
p(\vx_{\neg i})\ppos_i(x_i; \theta_i)$, which is the product of the original
density for all dimensions except the $i$th dimension, and the positive density
decomposition; the same holds for the negative component. For the derivative
with respect to the mean when the measure is a diagonal Gaussian in
\eqref{eq:mvd_gauss_mean_deriv_ex}, this decomposition is the product of a
Weibull distribution for the $i$th component, and a Gaussian
distribution for all other dimensions, which are easy to sample from when
forming the estimator. Full-covariance Gaussian measures will follow a similar
decomposition, sampling from the multivariate Gaussian and replacing the $i$th
component with a Weibull variate.

\subsection{Estimator Properties and Applicability}
\subsubsection{Domination}
The two derivatives-of-measure---the score function and the measure-valued
derivative---differ in the ways that they establish the correctness of the
interchange of differentiation and integration. For the score function
estimator, we achieved this in Section
\ref{sect:score_func_estimators_properties} by invoking the dominated convergence theorem and
assuming the gradient estimator is dominated (bounded) by a constant. We
explored one example where we were unable to ensure domination (using the
example of bounded support  in Section
\ref{sect:score_func_estimators_properties}), because no bounding constant
applies at the boundaries of the domain. For the weak derivative, the
correctness of the interchange of differentiation and integration is guaranteed
by its definition: the \textit{fundamental property of weak derivatives} states
that if the triple $(c, \ppos, \pneg)$ is a weak derivative of the probability
measure $p(x)$, then for every bounded continuous function $f$
\citep{pflug1996optimization, heidergott2000measure}
\begin{equation}
\nabla_\theta \int f(x)p(x; \theta) \intd x = c_\theta \left[
\int f(x)\ppos(x; \theta) \intd x - \int f(x)\pneg(x; \theta) \intd x\right].
\end{equation}
We should therefore expect the pathology in the example on bounded support not to
appear.

\begin{gradexample}{Bounded support revisited}
We revisit our earlier example from Section \ref{sect:abs_cont} in which we compute the gradient of
the objective
\eqref{eq:gradient} using the cost function $\cost(x) =x$ and distribution
$\dist_{\theta}(x) = \frac{1}{\theta}\mathds{1}\{0<x<\theta\}$, which is
differentiable in $\theta$ when $x \in (0, \theta)$. The measure-valued
derivative is
\begin{subequations}
\begin{align}
\nabla_\theta \int f(x)\mathcal{U}_{[0, \theta]}(x) \intd x & = \nabla_\theta \left(  \frac{1}{\theta} \int_0^\theta
f(x)\intd x \right) = \frac{1}{\theta}f(\theta) - \frac{1}{\theta^2}\int_0^\theta f(x)\intd x \\
& = \frac{1}{\theta} \left(\int f(x) \delta_{\theta}(x) \intd x - \int f(x)\mathcal{U}_{[0, \theta]}(x) \intd x \right).
\label{eq:mvd_bounded_support}
\end{align}
\end{subequations}
In the first line, we fixed the limits of integration using the support of the
uniform distribution and then applied the product rule for differentiation to
obtain the two terms. In the second line \eqref{eq:mvd_bounded_support}, we
rewrite the terms to reflect the triple form with which we have been
communicating the weak derivative. The measure-valued derivative is given by the
triple $\left( \tfrac{1}{\theta}, \delta_{\theta}, \mathcal{U}_{[0, \theta]}
\right)$; the positive part is a discrete measure whereas
the negative part is continuous in the range $(0, \theta)$. Using this result
for our specific gradient problem, we find:
\begin{subequations}
\begin{eqnarray}
\textrm{True gradient: }& \nabla_\theta \expect{\dist_{\theta}(x) }{x} =
\nabla_\theta \left( \left.
\frac{x^2}{2\theta}
\right|_0^\theta\right) = \frac{1}{2};\\
\textrm{Measure-valued gradient: }&  \tfrac{1}{\theta}\left(\expect{\delta_{\theta}}{x} - \expect{
\mathcal{U}_{[0, \theta]}}{x}\right) = \tfrac{1}{\theta}\left(\theta - \tfrac{\theta}{2}\right) = \frac{1}{2}.
\label{eq:mvd_uniform_dist_example}
\end{eqnarray}
\end{subequations}
By not requiring domination, the measure-valued derivative achieves the correct value.
\end{gradexample}

\subsubsection{Bias and Variance Properties}
By using the fundamental property of weak derivatives,
which requires bounded and continuous cost functions $f$, we can show that the measure-valued
gradient \eqref{eq:mvd_gaussian_weak_deriv}
provides an unbiased estimator of the gradient.  Unbiasedness can also be shown
for other types of cost functions, such as  those that are unbounded, or those
with sets of discontinuities that have measure zero under the positive and
negative components \citep[sect 4.2]{pflug1996optimization}.

The variance of the measure-valued derivative estimator is
\begin{equation}
\mathbb{V}_{p(\vx; \theta)}[\eta_{N=1}] = \mathbb{V}_{\ppos(\vx; \theta)}[f(\vx)] +
\mathbb{V}_{\pneg(\vx; \theta)}[f(\vx)] - 2\textrm{Cov}_{\ppos(\vx'; \theta)\pneg(\vx;
\theta)}[f(\vx'),f(\vx)].
\label{eq:wd_variance}
\end{equation}
From this, we see that the variance of the gradient estimator
depends on the choice of decomposition of the weak derivative into its
positive and negative components. An orthogonal decomposition using the
Hahn-Jordan decomposition is suggested to give low variance in general \citep{pflug1989sampling}.
Furthermore, we see that if the random variables can be `coupled' in some way,
where they share the same underlying source of randomness, this can reduce
gradient variance by increasing the covariance term in \eqref{eq:wd_variance}.
The most common coupling scheme is to sample the variables $\dot\vx$ and
$\ddot\vx$ using common random numbers---this is what we used in
Figures~\ref{fig:grad_var_quadratic} and \ref{fig:grad_var_exp_cos}.
From the figures, we see that the measure-valued derivative estimator with coupling
often provides lower variance than the alternative estimators.
\citet[example 4.21]{pflug1996optimization} discusses variance reduction for
measure-valued derivatives and approaches for coupling in detail, and we expand on this in Section
\ref{sect:var_reduction_coupling}.

\begin{figure}[t]
	\centering
	\begin{subfigure}[t]{0.45\textwidth}
		\centering
	\includegraphics[width=\textwidth]{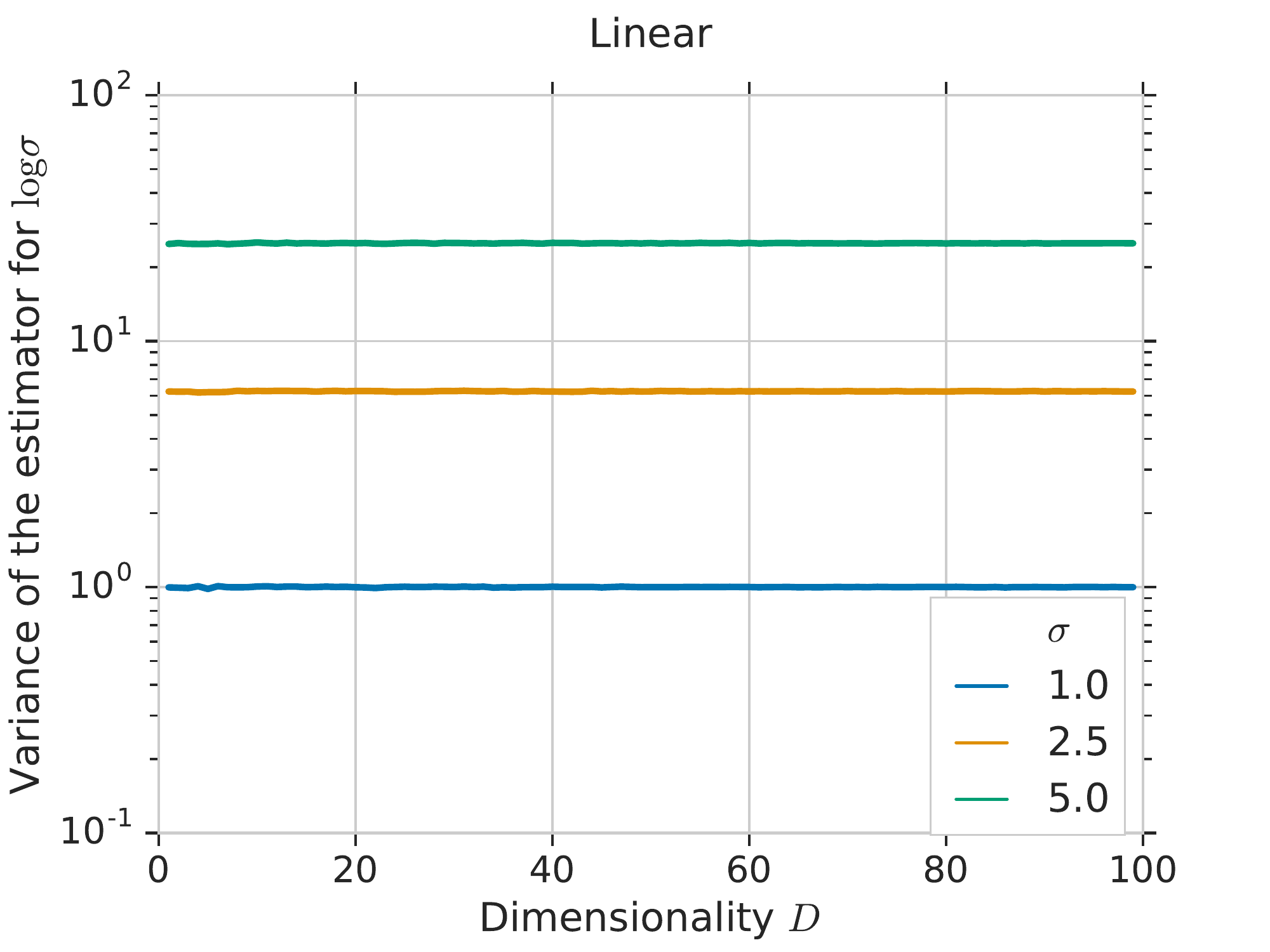}
	\end{subfigure}%
	\hspace{1cm}
	\begin{subfigure}[t]{0.45\textwidth}
	\centering
	\includegraphics[width=\textwidth]{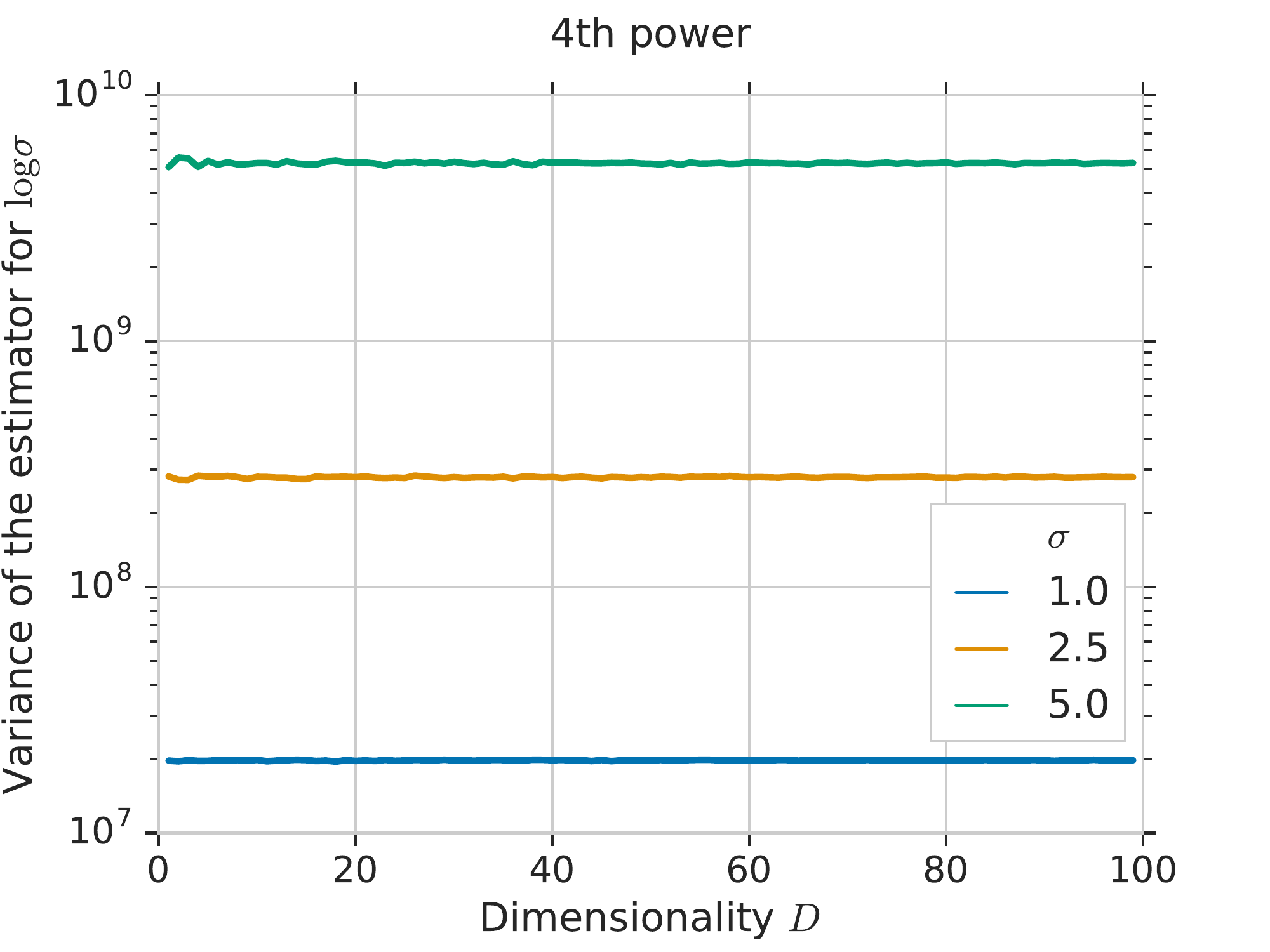}
\end{subfigure}%
\caption{Variance of the coupled measure valued estimator for a Gaussian measure $\mathcal{N}(\vx | 10,\sigma^2\vI_D)$
	and two cost functions: a linear cost $f(\vx)= \sum_d x_d$ and a fourth-order cost $f(\vx)= \sum_d x^4_d$.
	The y-axis is the estimate of the average variance $\Var[\nabla_{\sigma}f(\mathbf{x})] $
	across parameter dimensions.}
\label{fig:measure_valued_var}
\end{figure}

To examine the effect of the number of measure parameters, data dimensionality, and cost function on
the variance of the coupled measure valued gradient estimator, Figure~\ref{fig:measure_valued_var}
shows the estimator variance for
the gradient $\nabla_\sigma \expect{\mathcal{N}(\vx | 10,
  \sigma^2\vI_D)}{f(\vx)}$, for a linear cost
$f(\vx) = \sum_d x_d$ for $\vx \in \mathbb{R}^D$ as well as
 the forth-order cost $f(\vx) = \sum_d x_d^4$. We observe that the measure valued estimator
 is not sensitive to the dimensionality of the measure parameters, unlike the
 score function estimator (see Figure~\ref{fig:reinforce_var}). It is however sensitive
 to the magnitude of the function, and the variance of the measure, which is reflected in the higher variance for the positive and negative components.

\subsubsection{Higher-order Derivatives}
The problem of computing higher-order derivatives involves computing $\int
\cost(x)\nabla_\theta^{(k)} p(x; \theta) \intd x$. The second derivative
$\nabla_\theta^2p(x; \theta)$ is also a signed measure, which means that it
too has a weak derivative representation. For higher-order derivatives we will
again be able to compute the gradient using a weighted difference of expectations, using the
weak-derivative triple $(c(\theta), {\ppos}^{(n)}, {\pneg}^{(n)} )$,
where ${\ppos}^{(n)}$ and ${\pneg}^{(n)}$ are
the positive and negative components of the decomposition for the $n$th order
density-derivative $\nabla_\theta^{(n)}p(x; \theta)$.

\subsubsection{Computational Considerations}

Measure-valued gradients are much more computationally expensive than the
score-function or pathwise gradients. This is because the gradient we computed
in \eqref{eq:wd_estimator} is the gradient for a single parameter: for every
parameter we require two evaluations of the cost function to compute its
gradient. It is this structure of adapting the underlying sampling
distributions for each parameter that leads to the low variance of the estimator
but at the same time makes its application to high-dimensional parameter spaces
prohibitive. For problems that do not have a large number of parameters, and for
which we can evaluate the cost $f$ frequently, e.g., being in the form of a
simulator rather than a costly interactive system, this gradient may prove
beneficial. Overall the computational cost is $\mathcal{O}(2NDL)$, for $N$
samples drawn from each of the positive and negative components of the weak
derivative, $D$-dimensional parameters $\vtheta$, and a cost $L$ of evaluating
the cost function. %

Taking into account the exposition of this section, points for consideration
when using the measure-valued derivatives are:
\vspace{-4mm}
\begin{itemize}[noitemsep]
	\item The measure-valued derivative can be \textit{used with any type of cost function}, differentiable or
	not, as long as we can evaluate it repeatedly for different inputs.
	\item It is applicable to\textit{ both discrete and continuous distributions}, which adds to its generality.
	\item The estimator is \textit{computationally expensive in high-dimensional parameter spaces} since
	it requires two cost function evaluations for every parameter.
	\item We will need methods to \textit{sample from the positive and negative measures}, e.g., the double-sided Maxwell distribution for the gradient of the Gaussian variance.
	\item Using the weak derivative requires \textit{manual derivation of the decomposition} at first, although for many common distributions the weak-derivative decompositions are known.
\end{itemize}

\subsection{Research in Measure-valued Derivatives}
The weak derivative was initially introduced by \citet{pflug1989sampling}, with
much subsequent work to establish its key properties, including behaviour under
convex combinations, convolution, transformation, restriction, and the use of
$L_1$ differentiability of the gradient problem as a sufficient condition for
unbiasedness. In most papers, the details of how to derive the weak derivative
were omitted; this important exposition was provided for many common
distributions by \citet{heidergott2003measure}. The connections between
perturbation analysis (pathwise estimation) and measure-valued differentiation
are explored by \citet{heidergott2003measure}, with specific exploration of
Markov processes and queuing theory applications.
\citet{heidergott2008sensitivity} also provide a detailed exposition for
derivatives with Gaussian measures, which is one of the most commonly
encountered cases. In machine learning, \citet{buesing2016stochastic} discuss
the connection between weak derivatives,  finite differences, and other
Monte Carlo gradient estimation methods, and \citet{rosca2019measure} provide an empirical comparison for problems in approximate Bayesian inference.

\section{Variance Reduction Techniques}
\label{sect:var_reduction}
From the outset, we listed low variance as one of the crucial properties
of Monte Carlo estimators. The three classes of
gradient estimators we explored each have different underlying conditions and
properties that lead to their specific variance properties. But in all cases the
estimator variance is something we should control, and ideally reduce, since
it will be one of the principal sources of performance issues.
As we search for the lowest variance gradient estimators, it is important to recognise
a key tension that will develop, between a demand for
computationally-efficient and effective variance reduction versus
low-effort, generic, or black-box variance reduction tools. We focus on four
common methods in this section---large-samples, coupling, conditioning, and
control variates. These methods will be described separately, but they can also be
combined and used in concert to further reduce the variance of an estimator.
Variance reduction is one of the largest
areas of study in Monte Carlo methods, with important expositions on this topic
provided by \citet[ch. 4]{robert2013monte}, \citet[ch. 8-10]{owen2013} and
\citet[ch. 4]{glasserman2013monte}.

\subsection{Using More Samples}

The simplest and easiest way to reduce the variance of a Monte Carlo estimator
is to\textit{ increase the number of samples} $N$ used to compute
the estimator \eqref{eq:mce_definition}. The variance of such estimators shrinks
as $O(\sfrac{1}{N})$, since the samples are independent, while the computational
cost grows linearly in $N$. For a more complex variance reduction method to be
appealing, it should compare favourably to this reference method in terms of
efficiency of variance reduction. If evaluating the cost function involves only
a computational system or simulator, then using more samples can be appealing,
since the computational cost can be substantially reduced by parallelising the
computation. For some problems, however, increasing the number of Monte Carlo
samples will not be an option, typically in problems where evaluating the cost
function involves a real-world experiment or interaction. We restricted our
discussion to the standard Monte Carlo method, which evaluates the average
\eqref{eq:mce_definition} using independent sets of random variates, but other approaches can provide
opportunities for improved estimation. Quasi Monte Carlo methods generate a set of
samples that form a minimum discrepancy sequence, which are both more efficient in the use of the
samples and can lead to reduced variance, with more to explore in this area (\citet[chp.
5]{glasserman2013monte}, \citet{leobacher2014introduction},
\citet{buchholz2018quasi}).

\subsection{Coupling and Common Random Numbers}
\label{sect:var_reduction_coupling}
We will often encounter problems that take the form of a difference between two expectations of a
function $\cost(\vx)$ under different but closely-related distributions $\dist_1(\vx)$ and
$\dist_2(\vx)$:
\begin{equation}
\eta =  \expect{\dist_1(\vx)}{\cost(\vx)} - \expect{\dist_2(\vx)}{\cost(\vx)}.
\end{equation}
We saw this type of estimation problem in the derivation of the measure-valued gradient
estimator~\eqref{eq:wd_estimator}. The direct approach to computing the difference would be to
estimate each expectation separately using $N$ independent samples and to then take their difference:
\begin{align}
\bar{\eta}_\text{ind} & = \frac{1}{N}\sum_{n=1}^N\cost\left(\hat\vx_1^{(n)}\right) -
\frac{1}{N}\sum_{n=1}^N\cost\left(\hat\vx_2^{(n)}\right)
= \frac{1}{N}\sum_{n=1}^N \left( \cost\left(\hat\vx_1^{(n)}\right) - \cost\left(\hat\vx_2^{(n)}\right)
\right),
\label{eq:diff_ind}
\end{align}
where $\hat\vx_1^{(n)}$ and $\hat\vx_2^{(n)}$ are i.i.d.~samples from
$\dist_1(\vx)$ and $\dist_2(\vx)$, respectively. We can achieve a simple form of
variance reduction by \textit{coupling} $\hat\vx_1^{(n)}$ and $\hat\vx_2^{(n)}$
in Equation \eqref{eq:diff_ind}, so that each pair $(\hat\vx_1^{(n)},
\hat\vx_2^{(n)})$ is sampled from some joint distribution $\dist_{12}(\vx_1,
\vx_2)$ with marginals $\dist_1(\vx)$ and $\dist_2(\vx)$. The variance of the
resulting coupled estimator $\bar{\eta}_\text{cpl}$ for $N=1$ is
\begin{align}
\mathbb{V}_{\dist_{12}(\vx_1, \vx_2)}\left[\bar{\eta}_\text{cpl}\right] 
    & = \mathbb{V}_{\dist_{12}(\vx_1, \vx_2)}\left[\cost(\vx_1)-\cost(\vx_2)\right] \nonumber \\
        & = \mathbb{V}_{\dist_{1}(\vx_1)}\left[\cost(\vx_1)\right] + \mathbb{V}_{\dist_{2}(\vx_2)}\left[\cost(\vx_2)\right] - 2\Cov_{\dist_{12}(\vx_1,
        \vx_2)}\left[\cost(\vx_1), \cost(\vx_2)\right] \nonumber \\
        & = \mathbb{V}_{\dist_1(\vx_1)\dist_2(\vx_2)}\left[\bar{\eta}_\text{ind}\right] - 2\Cov_{\dist_{12}(\vx_1,
        \vx_2)}\left[\cost(\vx_1), \cost(\vx_2)\right].
\end{align}
Therefore, to reduce variance we need to choose a coupling $\dist_{12}(\vx_1, \vx_2)$
such that $\cost(\vx_1)$ and $\cost(\vx_2)$ are positively correlated. In
general, finding an optimal coupling can be difficult, even in the univariate
case since it involves the inverse CDFs of $\dist_1(\vx_1)$ and $\dist_2(\vx_2)$
(see e.g.~\citet[p.~225]{pflug1996optimization}).
The use of \textit{common random numbers} is one simple and popular coupling technique that can
be
used when $\dist_1(\vx_1)$ and
$\dist_2(\vx_2)$ are close or in a related family of distributions.
Common random numbers involve sharing, in some way, the underlying random numbers used in
generating the random variates $\hat\vx_1$ and $\hat\vx_2$. For example, in the univariate
case, we can do this by sampling $u \sim \mathcal{U}[0, 1]$ and applying the inverse CDF
transformations: $x_1 = CDF_{p_1}^{-1}(u)$ and $x_2 = CDF_{p_2}^{-1}(u)$.

\begin{figure}[t]
  \centering
  \includegraphics[width=\textwidth]{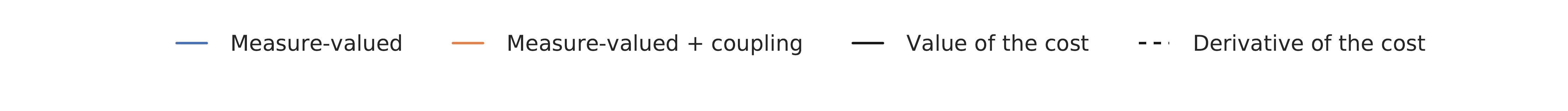} \\
  \vspace{-0.5cm}
  \begin{subfigure}[t]{0.45\textwidth}
    \centering
    \includegraphics[width=\linewidth]{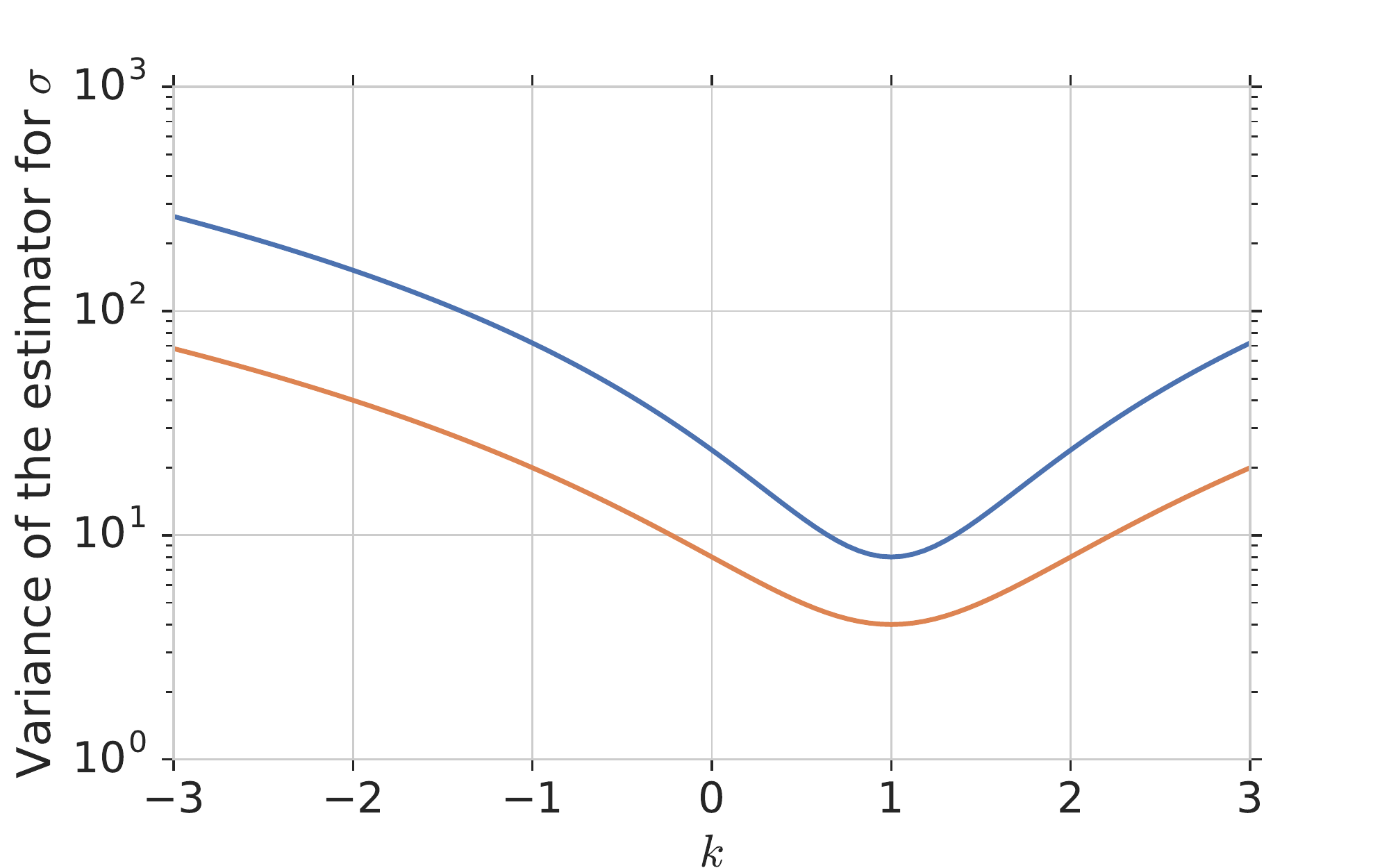} \\
    \includegraphics[width=0.9\linewidth,right]{costs_square}
  \end{subfigure}%
  \hspace{1cm}
  \begin{subfigure}[t]{0.45\textwidth}
    \centering
    \includegraphics[width=\linewidth]{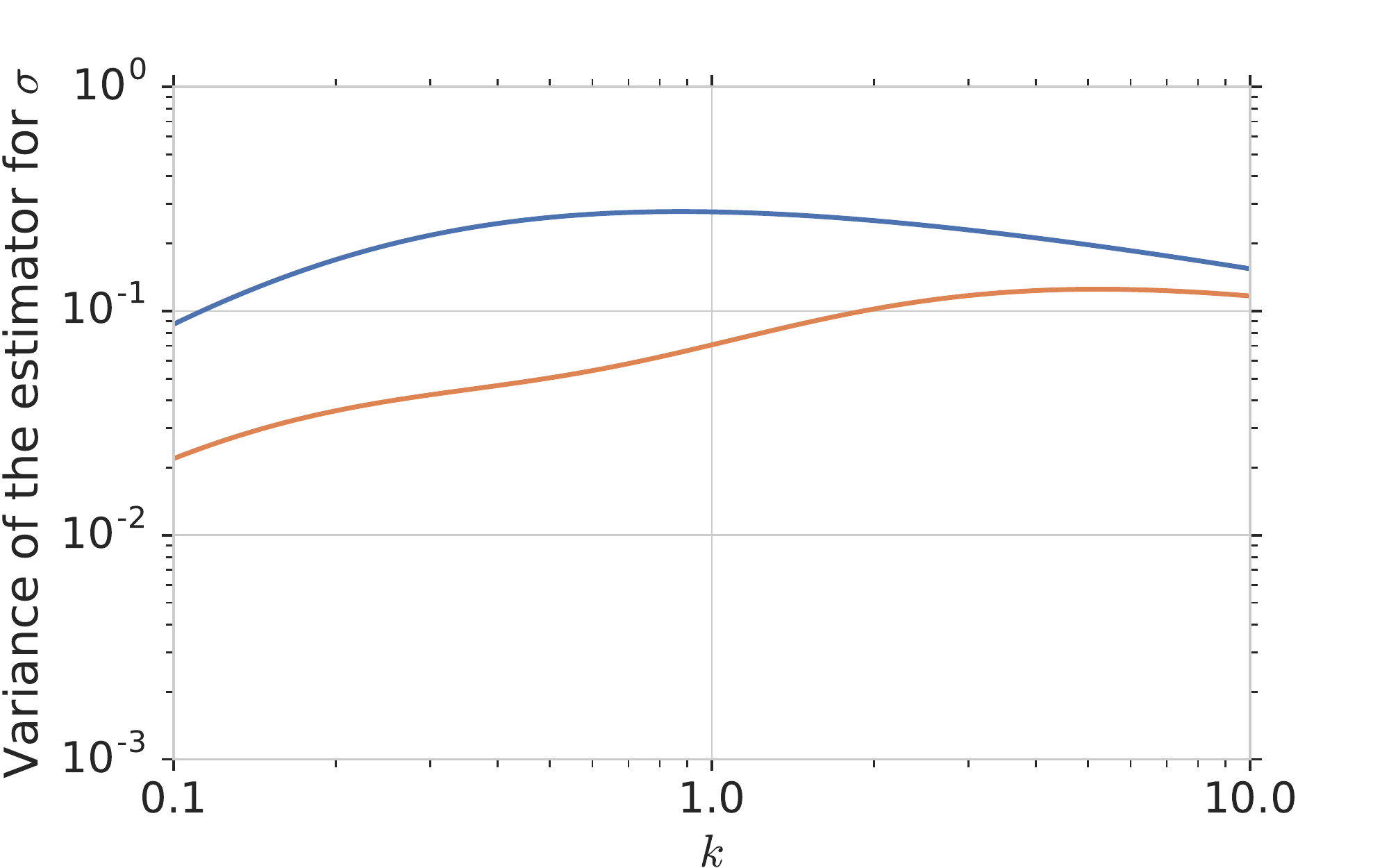} \\
    \includegraphics[width=0.9\linewidth,right]{costs_exp}
  \end{subfigure}
  \caption{The effect of Maxwell-Gaussian coupling on the variance of the stochastic estimates of
  the measure valued estimator for $\nabla_{\sigma} \expect{\mathcal{N}(x | \mu,
      \sigma^2)}{\cost(x; k)}$ for $\mu = \sigma = 1$ as a function of $k$.
      Left:  $\cost(x; k)=(x-k)^2$; right:
      $\cost(x; k)=\exp(-kx^2)$. The graphs in the bottom row
      show the function (solid) and its gradient (dashed).}
  \label{fig:coupling_no_coupling_sigma}
\end{figure}

\begin{gradexample}{Maxwell-Gaussian Coupling}
Consider Gaussian measures~$\mathcal{N}(x| \mu, \sigma^2)$ in the setting of
Equation \eqref{eq:gradient} and the task of computing the gradient with
respect to the standard deviation $\sigma$.	The measure-valued gradient, using
Table \ref{tab:mvd_triples}, is given by the triple $\left(\tfrac{1}{\sigma},
\mathcal{M}(x | \mu, \sigma^2), \mathcal{N}(x | \mu, \sigma^2) \right)$, where
$\mathcal{M}$ is the double-sided Maxwell distribution with location $\mu$ and
scale $\sigma^2$, and $\mathcal{N}$ is the Gaussian distribution with mean $\mu$
and variance $\sigma^2$; see Appendix \ref{app:distribution} for the densities for these distributions.
We can couple the Gaussian and the Maxwell distribution by  exploiting their
corresponding sampling paths with a common random number: if we can generate
samples $\dot{\varepsilon} \sim \mathcal{M}(0, 1)$ then, by first sampling from
the Uniform distribution $\dot{u} \sim \mathcal{U}[0,1]$ and reusing the Maxwell
samples, we can generate $\mathcal{N}(0, 1)$ distributed samples
$\ddot{\varepsilon}$ via $\ddot{\varepsilon} = \dot{\varepsilon} \dot{u}$. Then,
we can perform a location-scale transform to obtain the desired Maxwell and
Normal samples: $\dot{x} = \mu + \sigma \dot{\varepsilon}, \quad \ddot{x} = \mu
+ \sigma \ddot{\varepsilon}$. The distributions are coupled because they use the
same underlying Maxwell-distributed variates
\citep{heidergott2008sensitivity}. The variance reduction effect of
Maxwell-Gaussian  coupling for the measure valued gradient estimator
can be seen in Figure~\ref{fig:coupling_no_coupling_sigma}.
\end{gradexample}

\begin{figure}[t]
  \centering
  \includegraphics[width=\textwidth]{measure_valued_estimator_and_costs} \\
  \vspace{-0.5cm}
  \begin{subfigure}[t]{0.45\textwidth}
    \centering
    \includegraphics[width=\linewidth]{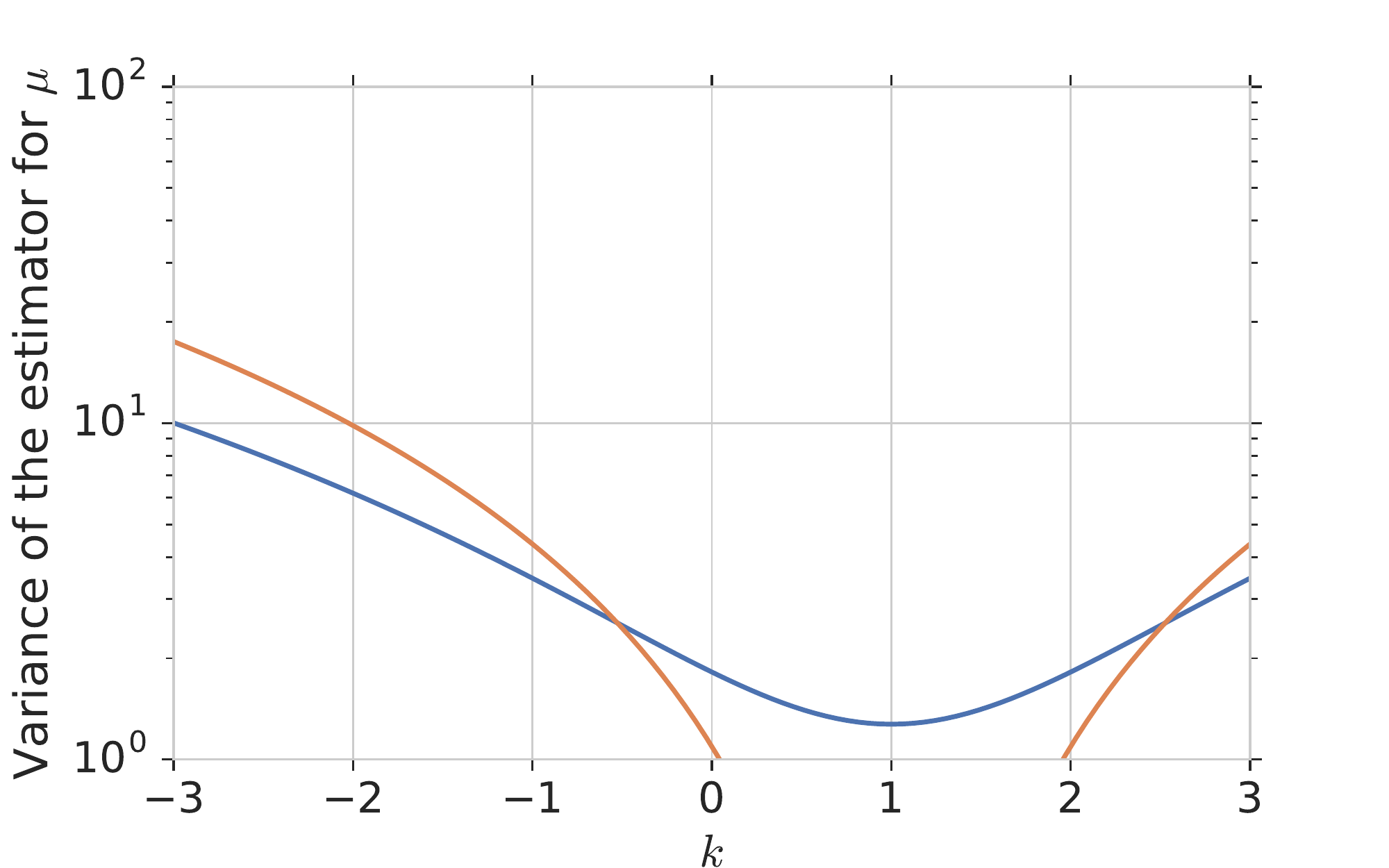} \\
    \includegraphics[width=0.9\linewidth,right]{costs_square}
  \end{subfigure}%
  \hspace{1cm}
  \begin{subfigure}[t]{0.45\textwidth}
    \centering
    \includegraphics[width=\linewidth]{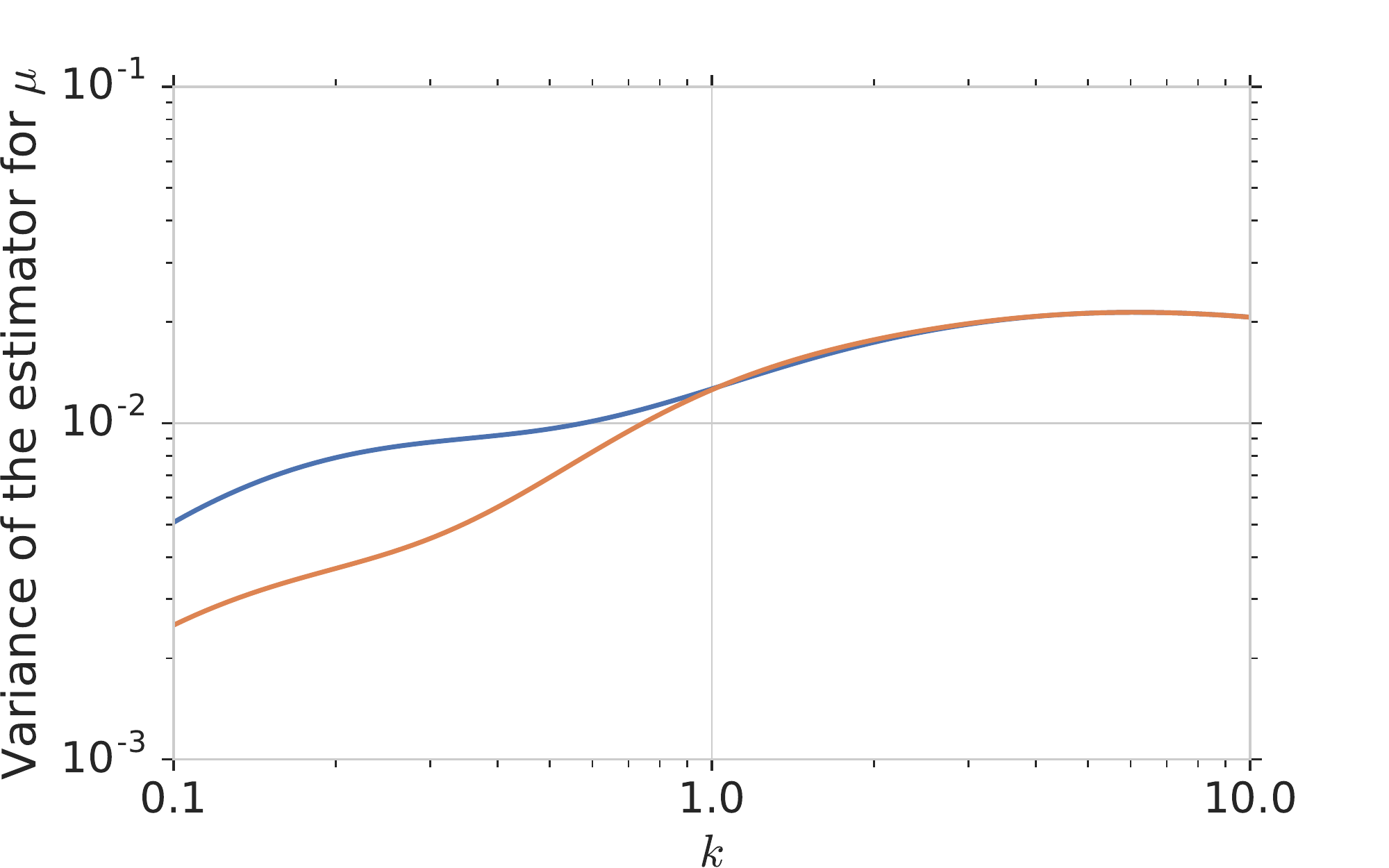} \\
    \includegraphics[width=0.9\linewidth,right]{costs_exp}
  \end{subfigure}
    \caption{The effect of Weibull-Weibull coupling on the variance of the stochastic estimates of
   the measure valued estimator for $\nabla_{\mu} \expect{\mathcal{N}(x | \mu,
      \sigma^2)}{\cost(x; k)}$ for $\mu = \sigma = 1$ as a function of $k$.
      Left:  $\cost(x; k)=(x-k)^2$; right:
      $\cost(x; k)=\exp(-kx^2)$. The graphs in the bottom row
      show the function (solid) and its gradient (dashed).}
  \label{fig:coupling_no_coupling_mu}
\end{figure}

\begin{gradexample}{Weibull-Weibull Coupling}
Consider Gaussian measures~$\mathcal{N}(x| \mu, \sigma^2)$ in the setting of
Equation \eqref{eq:gradient} and the task of computing the gradient with
respect to the mean $\mu$. The measure-valued gradient, using
Table \ref{tab:mvd_triples}, is given by the triple $\left(\sfrac{1}{\sigma\sqrt{2\pi}},
\theta + \sigma \mathcal{W}(2, 0.5), \theta - \sigma \mathcal{W}(2, 0.5)\right)$
where $\mathcal{W}$ is the Weibull distribution; see Appendix \ref{app:distribution} for its density function.
We can apply coupling by reusing the Weibull samples when computing the positive and negative terms
of the estimator. Figure~\ref{fig:coupling_no_coupling_mu} shows that using Weibull-Weibull
coupling does not always reduce the variance of the measure valued estimator;
depending on the cost function, coupling can increase variance.
\end{gradexample}

The main drawback of using coupling is that implementing it can require
non-trivial changes to the code performing sampling, which can make it less
appealing than variance reduction methods that do not require any such changes.

\subsection{Variance Reduction by Conditioning}

Probabilistically conditioning our estimators on a subset of dimensions and integrating out
the remaining dimensions analytically is a powerful form of variance reduction in Monte Carlo estimators, sometimes
referred to
as Rao-Blackwellisation. Assume that the dimensions $\{1,...,D\}$ of $\vx$ are
partitioned into a set $\condSet$ and its complement $\condSetC =
\{1,...,D\} \setminus \condSet$ with the expectation $g(\vx_{\condSetC}) =
\expectGiven{\dist(\vx_\condSet)}{\cost(\vx)}{\vx_{\condSetC}}$ being
analytically computable for all values of $\vx_{\condSetC}$.  We can then estimate
$\expect{\dist(\vx)}{\cost(\vx)}$ by performing Monte Carlo integration over a smaller
space of $\vx_{\condSetC}$:
\begin{align}
    \bar{g}_N = \frac{1}{N}\sum_{n=1}^N g\left(\vxhn_{\condSetC}\right) = \frac{1}{N}\sum_{n=1}^N
    \expectGiven{\dist\left(\vxhn_\condSet\right)}{\cost\left(\vxhn\right)}{\vxhn_{\condSetC}} \text{ where } \vxhn_{\condSetC} \sim \dist(\vx_{\condSetC})
    \text{ for }n=1,...,N.
\end{align}
We can show that this conditional estimator has lower variance than its unconditional counterpart.
Following
\citet{owen2013}, by the law of total variance, we have
\begin{align}
    \Var_{\dist(\vx)}[\cost(\vx)] & = \expect{\dist(\vx_{\condSetC})}{\VarGiven{\dist(\vx_\condSet)}{\cost(\vx)}{\vx_{\condSetC}}} +
    \Var_{\dist(\vx_{\condSetC})}[\expectGiven{\dist(\vx_\condSet)}{\cost(\vx)}{\vx_{\condSetC}}]  \\
&  = \expect{\dist(\vx_{\condSetC})}{\VarGiven{\dist(\vx_\condSet)}{\cost(\vx)}{\vx_{\condSetC}}} +
\Var_{\dist(\vx_{\condSetC})}[g\left(\vx_{\condSetC}\right)]  \\
& \ge \Var_{\dist(\vx_{\condSetC})}[g(\vx_{\condSetC})],
\label{eq:lotv}
\end{align}
which shows that the conditional system has lower variance than than the unconditional system
$f$.
Thus the conditional estimator $\bar{g}_N$ is guaranteed to have lower variance than
unconditional one $\bar{f}_N = \frac{1}{N}\sum_{n=1}^N f\left(\vxhn\right)$:
\begin{align}
\Var_{\dist(\vx)}\left[\bar{f}_N\right] = \frac{1}{N} \Var_{\dist(\vx)}[f(\vx)] \ge \frac{1}{N}
   \Var_{\dist(\vx)}[g(\vx_{\condSetC})] = \Var_{\dist(\vx)}\left[\bar{g}_N\right].
\end{align}
Conditioning is the basis of several variance reduction schemes: smoothed perturbation analysis for
the pathwise
derivative \citep{gong1987smoothed}; the use of leave-one-out estimation
and Rao-Blackwellisation in hierarchical models by \citet{ranganath2017black}; and the
local-expectation gradients developed by \citet{titsias2015local} to improve on
the variance properties of score-function and pathwise estimators in variational
inference.

This technique is useful in practice only if we can compute the conditional
expectations involved efficiently, whether analytically or with numerical
integration. If evaluating these expectations is costly, then simply
increasing the number of samples in the Monte Carlo estimator over the original
space may prove easier and at least as effective.
\subsection{Control Variates}
We now consider a general-purpose technique for reducing the variance of any
Monte Carlo method, again looking at the general problem
$\expect{\dist(\vx;\distParams)}{\cost(\vx)}$ \eqref{eq:expectation_function},
which we can do since all the gradient estimators we developed in the preceding
sections were eventually of this form. The strategy we will take is to replace
the function $\cost(\vx)$ in the expectation \eqref{eq:expectation_function} by
a substitute function $\tilde{\cost}(\vx)$ whose expectation
$\expect{\dist(\vx;\distParams)}{\tilde\cost(\vx)}$ is the same, but whose variance is lower.
If we have a function $h(\vx)$ with a known
expectation $\expect{\dist(\vx; \distParams)}{h(\vx)}$, we can easily construct such a substitute function
along with the corresponding estimator as follows:
\begin{eqnarray}
& \tilde{\cost}(\vx) = \cost(\vx) - \beta(h(\vx) - \expect{\dist(\vx; \distParams)}{h(\vx)})
\label{eq:linear_control_variate} \\
& \bar\eta_N = \frac{1}{N}\sum_{n=1}^N
\tilde{\cost}(\hat\vx^{(n)}) = \bar{f} - \beta(\bar{h} -  \expect{\dist(\vx; \distParams)}{h(\vx)}),
\label{eq:linear_control_variate_estimator}
\end{eqnarray}
where $\hat\vx^{(n)} \sim \dist(\vx; \distParams)$  and we use the shorthand
$\bar{f}$ and $\bar{h}$ denote the sample averages. We refer to the function
$h(\vx)$ as a control variate,  and the estimator
\eqref{eq:linear_control_variate_estimator} a \textit{control variate
	estimator}. Control variates allow us to exploit information about quantities
with known expectations, to reduce the variance of an estimate of an unknown expectation.
The observed error $(h(\vx) - \expect{\dist(\vx; \distParams)}{h(\vx)})$ serves
as a control in estimating $\expect{\dist(\vx;\distParams)}{\cost(\vx)}$, and
$\beta$ is a coefficient that affects the strength of the control variate. We
expand on the specific choice of $h$ in Section \ref{sect:cv_designing}.
\citet{glynn2002some} describe the connections between control variates and other variance
reduction approaches such as antithetic and stratified sampling, conditional Monte Carlo, and
non-parametric maximum likelihood.

\subsubsection{Bias, Consistency and Variance}
We can show that if the variance of $h(\vx)$ is finite, both the unbiasedness and consistency of the estimator
\eqref{eq:linear_control_variate_estimator} are maintained:
\begin{align}
& \textrm{\textbf{Unbiasedness:} }\expect{\dist(\vx;\distParams)}{\tilde{f}(\vx; \beta)} =
\expect{}{\bar{f} -
\beta(\bar{h} -  \expect{}{h(\vx)})} = \expect{}{\bar{f}}
= \expect{\dist(\vx;\distParams)}{\cost(\vx)}, \\
& \textrm{\textbf{Consistency:} } %
\lim_{n\rightarrow \infty} \frac{1}{N}\sum_{n=1}^N \tilde{\cost}(\vxhn) =
\expect{\dist(\vx;\distParams)}{\tilde{\cost}(\vx)} =  \expect{\dist(\vx;\distParams)}{\cost(\vx)}.
\end{align}
Importantly, we can characterise the variance of the estimator (for $N=1$) as
\begin{equation}
  \Var[\tilde{\cost}] = \Var[\cost(\vx) - \beta(h(\vx) - \expect{\dist(\vx;
  \distParams)}{h(\vx)})] = \Var[\cost] - 2\beta \textrm{Cov}[f,h] + \beta^2\Var[h].
  \label{eq:variance_linear_cv}
\end{equation}
By minimising \eqref{eq:variance_linear_cv} we can find that the optimal value of the coefficient is
\begin{equation}
\label{eq:cv_coeff}
\beta^* = \frac{\textrm{Cov}[f,h]}{\Var[h]} = \sqrt {\frac{\Var[f]}{\Var[h]}}\textrm{Corr}(f,h),
\end{equation}
where we expressed the optimal coefficient in terms of the variance of
$f$ and $h$, as well as in terms of the correlation coefficient
$\textrm{Corr}(f,h)$. The effectiveness of the control variate can be
characterised by the ratio of the variance of the control variate estimator to
that of the original estimator: we can say our efforts have been effective if the
ratio is substantially less than 1. Using the optimal control coefficient in
\eqref{eq:variance_linear_cv}, we find that the potential variance reduction is
\begin{equation}
\frac{\Var[\tilde{\cost}(\vx)]}{\Var[\cost(\vx)]} = \frac{\Var[\cost(\vx) - \beta(h(\vx) -
\expect{p(\vx;\distParams)}{h(\vx)})]}{\Var[\cost(\vx)]} = 1 - \textrm{Corr}(\cost(\vx),h(\vx))^2.
\end{equation}
Therefore, as long as $\cost(\vx)$ and $h(\vx)$ are not uncorrelated, we can
always obtain a reduction in variance using control variates. However, for variance reduction to be
substantial, we need control variates that are strongly correlated with the
functions whose variance we wish to reduce. The sign of the correlation does not
matter since it is absorbed into $\beta^*$, and the stronger the correlation,
the larger the reduction in variance.

In practice, the optimal $\beta^*$ will not be known and so we will usually need to
estimate it empirically. Estimating $\bar{\beta}_N$ using the same
$N$ samples used to estimate $\bar{h}$ will introduce a bias because
$\bar{\beta}_N$ and $\bar{h}$ will no longer be independent. In practice, this
bias is often small \citep{owen2013, glasserman2013monte}, and can be controlled
since it decreases quickly as the number of samples $N$ is increased. This bias can also be eliminated by using a family of `splitting techniques', as described by  \citet{avramidis1993splitting}.
We refer to the discussion by \citet{nelson1990control} for additional analysis on this topic.

\subsubsection{Multiple and Non-linear Controls}
The control variates in \eqref{eq:linear_control_variate} are \textit{linear
	control variates}, and this form should be used as a default. A first
generalisation of this is in allowing for multiple controls, where we use a set of
$D$ controls rather than one, using
\begin{equation}
\tilde{\cost}(\vx) = \cost(\vx) - \vbeta^\top(\vh(\vx) - \expect{\dist(\vx; \distParams)}{\vh(\vx)}),
\label{eq:linear_control_variate_multiple} \\
\end{equation}
where $\vbeta$ is now a $D$-dimensional vector of coefficients, with $D$
usually kept low; the bias and variance analysis follows the single-control
case above closely. The use of multiple controls is best suited to cases when
the evaluation of the function $f$ is expensive and can be approximated by a linear
combination of inexpensive control variates.

We can also consider non-linear ways of introducing control variates. Three alternative (multiplicative)
forms are
\begin{equation}
\tilde{\cost} = \cost\frac{\expect{\dist(\vx;\distParams)}{h}}{h};
\quad
\tilde{\cost} =
\cost^{\frac{h}{\expect{\dist(\vx;\distParams)}{h}}};
\quad
\tilde{\cost} = \cost\exp(\expect{\dist(\vx;\distParams)}{h}-h).
\end{equation}
These multiplicative controls lead to consistent but biased estimates.  The
non-linear controls also do not require a coefficient $\beta$ since this is
implicitly determined by the relationship between $h$ and $f$. When we use a
large number of samples in the estimator, non-linear controls are no better than
linear controls, since asymptotically it is the linear terms that matter
\citep[sect. 8]{glynn1989indirect}. In most contemporary applications, we do not
use large samples---often we use a single sample in the estimator---and it is in
this small sample regime that non-linear controls may have applicability. The
practical guidance though, remains to use linear controls as the default.

\subsubsection{Designing Control Variates}
\label{sect:cv_designing}
The primary requirements for a control variate are that it must have a tractable
expectation and, ideally, be strongly correlated with the target. While control
variates are a very general technique that can be applied to any Monte Carlo estimator,
designing effective control variates is an art, and the best control variates
exploit the knowledge of the problem at hand. We can identify some general
strategies for designing controls:
\begin{description}
  \item[Baselines.] One simple and general way to reduce the variance of
  a score-function gradient estimator is to use the score function itself
  as a control variate, since its expectation under the
  measure is zero (Equation \eqref{eq:expected_score}). This simplifies to subtracting
  the coefficient $\beta$ from the cost in the score function estimator in
  \eqref{eq:score_estimation_mce}, giving the modified estimator
\begin{align}
    \bar{\eta}_N & = \frac{1}{N}\sum_{n=1}^N \left(\cost(\vxhn; \costParams) - \beta\right)
    \nabla_\distParams \log \dist(\vxhn;
    \distParams); \quad \vxhn \sim \dist(\vx; \distParams).
    \label{eq:score_estimation_mce_baseline}
\end{align}
  In reinforcement learning, $\beta$ is called a \emph{baseline}
\citep{williams1992simple} and has historically been estimated with a running
average of the cost. While this approach is easier to implement than
optimising $\beta$ to minimise variance, it is not optimal and does \textit{not}
guarantee lower variance compared to the vanilla score-function estimator
\citep{weaver2001optimal}. Baselines are among the most popular variance
reduction methods because they are easy to implement, computationally cheap,
and require minimal knowledge about the problem being solved. We explore the
baseline approach in an empirical analysis in Section
\ref{sect:blr_experiments}.
\item[Bounds.] We can use bounds on the cost function $f$ as ways of
specifying the form of the control variate $h$. This is intuitive because it
maintains a correlation between $f$ and $h$, and if chosen well, may be
easily integrable against the measure and available in closed form. This
approach requires more knowledge of the cost function, since we will need to
characterise the cost analytically in some way to bound it. Several bounding methods,
such as B\"{o}hning and Jaakkola's bounds \citep{khan2010variational}, and piecewise bounds
\citep{marlin2011piecewise}, have been developed in local variational inference.
\citet{paisley2012variational} show specific use of bounds as control
variates. In general, unless the bounds used are tight, they will not be
effective as control variates, since the gap between the bound and the true
function is not controllable and will not necessarily give the information
needed for variance reduction.
\item[Delta Methods.] We can create control variates by approximating the cost
function using a Taylor expansion. This requires a cost function that is
differentiable so that we can compute the second-order Taylor expansion, but can
be an effective and very general approach for variance reduction that allows
easy implementation \citep{paisley2012variational, owen2013}. We will develop a specific example in Section
\ref{sect:delta_cv}.

  \item[Learning Controls.] If we estimate the gradients repeatedly as a
part of an optimisation process, we can create a parameterised control variate
function and learn it as part of the overall optimisation algorithm. This
allows us to create a control variate that dynamically adapts over the course
of optimisation to help better match the changing nature of the cost function.
Furthermore, this approach makes it easy to condition on any additional
information that is relevant, e.g.,~by using a neural network to map the
additional information to the parameters of the control variate.  This
strategy has been used by \citet{mnih2014neural} to reduce the score-function
gradient variance substantially in the context of variational inference, by
defining the baseline as the output of a neural network that takes the observed data
point as its input. This approach is general, since it does not make
assumptions about the nature of the cost function, but introducing a new set
of parameters for the control function can make the optimisation process
more expensive.

Even in the simpler case of estimating the gradient only once, we can still
learn a control variate that results in non-trivial variance reduction as long
as we generate sufficiently many samples to train it and use a
sample-efficient learning algorithm. \citet{oates2017control}
demonstrated the feasibility of this approach using kernel-based
non-parametric control variates.
\end{description}

\section{Case Studies in Gradient Estimation}
\label{sect:case_studies}
We now bring together the different approaches for gradient estimation in
a set of case studies to explore how these ideas have been combined in
practice. Because these gradients are so fundamental, they appear in all the
major sub-areas of machine learning---in supervised, unsupervised, and
reinforcement learning. Not all combinations that we can now think of using what
we have reviewed have been explored, and the gaps we might find
point to interesting topics for future work.

\subsection{Delta Method Control Variates}
\label{sect:delta_cv}

This case study makes use of the delta method \citep{bickel2015mathematical},
which is a method for describing the asymptotic variance of a function of a
random variable. By making use of Taylor expansions, the delta method allows us
to design lower variance gradient estimators by using the differentiability, if
available, of the cost function. It can be used for variance reduction in both
the score-function estimator \citep{paisley2012variational} and the pathwise
estimator \citep{miller2017reducing}, and shows us concretely one approach for
designing control variates that was listed in Section \ref{sect:cv_designing}.

We denote the gradient and Hessian of the cost function $f(\vx)$ as $\vgamma(\vx) :=
\nabla_{\vx} f(\vx)^\top$ and $\vH(\vx) := \nabla^2_{\vx} f(\vx) $, respectively. The
second-order Taylor expansion of a cost function $f(\vx)$ expanded around
point $\vmu$ and its derivative are
\begin{align}
h(\vx) & = f(\vmu) + (\vx - \vmu)^\top\vgamma(\vmu) + \tfrac{1}{2} (\vx -
\vmu)^\top\vH(\vmu)(\vx - \vmu),  \\
\nabla_{\vx} h(\vx) & = \vgamma(\vmu)^\top  + (\vx - \vmu)^\top\vH(\vmu).
\end{align}
We can use this
expansion directly as a control variate for the score-function estimator, following the
approach
we
took to obtain Equation \eqref{eq:linear_control_variate}:
\begin{subequations}
\begin{align}
\veta_{SF} & = \nabla_{\vtheta} \expect{\dist(\vx;\distParams)}{\cost(\vx)} =
\nabla_{\vtheta} \expect{\dist(\vx;\distParams)}{\cost(\vx) - \vbeta^\top h(\vx)}
+ \vbeta^\top \nabla_{\vtheta} \expect{\dist(\vx; \distParams)}{h(\vx)}
\label{eq:case_delta_sf_def}\\
&  =\expect{\dist(\vx; \vtheta)}{ \left(  f(\vx) - \vbeta^\top
	h(\vx) \right) \nabla_{\vtheta} \log
	\dist(\vx; \vtheta) } +
\vbeta^\top\nabla_{\distParams} \expect{\dist(\vx; \distParams)}{h(\vx)}.
\label{eq:case_delta_sf}
\end{align}
\end{subequations}
In the first line, we again wrote the control-variate form for a Monte Carlo gradient,
and in the
second
line we wrote out the gradient using the score-function approach. This is the
control variate
that is proposed by \citet[eq.
(12)]{paisley2012variational} for Monte Carlo gradient estimation in variational
inference problems. In the Gaussian mean-field variational
inference that \citet{paisley2012variational} consider, the second term is known
in closed-form and hence does not require Monte Carlo approximation. Like in
Equation \eqref{eq:linear_control_variate_multiple}, $\vbeta$ is a multivariate
control coefficient and is estimated separately.

With the increased popularity of the pathwise gradient estimator in machine
learning, \citet{miller2017reducing} showed how delta method control
variates can be used with pathwise estimation in the context of 
variational inference. The approach is not fundamentally different from
that of \citet{paisley2012variational}, and we show here that they can be
derived from each other. If we begin with Equation \eqref{eq:case_delta_sf_def}
and instead apply the pathwise estimation approach, using a sampling path $\vx =
g(\vepsilon; \vtheta); \vepsilon \sim p(\vepsilon)$, we will obtain the gradient
\begin{subequations}
\begin{align}
\veta_{PD} & = \nabla_{\vtheta} \expect{\dist(\vx;\distParams)}{\cost(\vx)} =
\nabla_{\vtheta} \expect{\dist(\vx;\distParams)}{\cost(\vx) - \beta h(\vx)}
+ \beta \nabla_{\vtheta} \expect{\dist(\vx; \distParams)}{h(\vx)}   \label{eq:case_delta_pd_def}
\\
& = \nabla_{\vtheta} \expect{p(\vepsilon)}{\cost(g(\vepsilon; \vtheta)) - \beta h(g(\vepsilon;
\vtheta))}
	+ \beta \nabla_{\vtheta} \expect{\dist(\vx; \distParams)}{h(\vx)} \\
& =  \expect{p(\vepsilon)}{\nabla_{\vx} \cost(\vx) \nabla_{\vtheta} g(\vepsilon; \vtheta) -
	\beta \nabla_{\vx} h(\vx)  \nabla_{\vtheta} g(\vepsilon; \vtheta)}
	+ \beta\nabla_{\vtheta} \expect{\dist(\vx; \distParams)}{h(\vx)}.  \label{eq:case_miller_pathwise_cv}
\end{align}
\end{subequations}
The first line recalls the linear control variate definition
\eqref{eq:linear_control_variate} and expands it. In the second line, we use the
LOTUS to reparameterise the expectation in terms of the sampling path $g$ and
the base distribution $p(\vepsilon)$, but again assume that the final term is
known in closed-form and does not require stochastic approximation. In the final
line \eqref{eq:case_miller_pathwise_cv} we exchanged the order of integration and
differentiation, and applied the chain rule to obtain the form of the gradient
described by \citet{miller2017reducing}, though from a different
starting point.

As both \citet{paisley2012variational} and \citet{miller2017reducing}
demonstrate, we can achieve an appreciable variance reduction using the delta
method, if we are able to exploit the generic differentiability of the cost
function. Due to the widespread use of automatic differentiation, methods like
these can be easily implemented and allow for a type of automatic control
variate, since no manual derivation and implementation is needed.

\subsection{Hybrids of Pathwise and Score Function Estimators}
Decoupling of gradient estimation and sampling was an aspect of the pathwise
estimator that we expanded upon in Section \ref{sect:pathwise_decoupling}.
This decoupling can be achieved in other ways than with implicit differentiation that we used previously, and we explore two other options in this section.

Up to this point, we have always required that the sampling path consists of
a transformation $\vx=g(\vepsilon; \vtheta)$ and a corresponding base
distribution $p(\vepsilon)$ that is \textit{independent} of the parameters
$\vtheta$. In some cases, we might only be able to find a weaker sampling path that consists
of a transformation $g$ and a \textit{parameter-dependent base
	distributions} $p(\vepsilon; \vtheta)$. In this setting, we will obtain
a gradient estimator that is a hybrid of the pathwise and score-function estimators:
\begin{subequations}
\begin{align}
\nabla_{\distParams} \expect{\dist(\vx; \distParams)}{\cost(\vx)} &= \nabla_{\distParams}
\expect{p(\vepsilon; \distParams)}{\cost(g(\vepsilon; \distParams))} = \nabla_{\distParams} \int
p(\vepsilon; \distParams) \cost(g(\vepsilon; \distParams)) \intd\vepsilon  \\
& = \int p(\vepsilon; \distParams) \nabla_{\distParams} \cost(g(\vepsilon; \distParams)) \intd \vepsilon +
\int \nabla_{\distParams}  p(\vepsilon; \distParams) \cost(g(\vepsilon; \distParams)) \intd\vepsilon  \\
& = \underbrace{\expect{p(\vepsilon; \distParams)}{\nabla_{\distParams} \cost(g(\vepsilon;
\distParams))}}_{\text{pathwise gradient}} +
\underbrace{\expect{p(\vepsilon; \distParams)}{\cost(g(\vepsilon; \distParams))
\nabla_{\distParams} \log p(\vepsilon; \distParams)}}_{\text{score-function gradient}}.
\label{eq:hybrids}
\end{align}
\end{subequations}
In the first line, we expanded the definition of the gradient-expectation under
the new parameter-dependent base distribution $p(\vepsilon; \vtheta)$
and the path $g(\vepsilon; \vtheta)$. In the second line, we applied the chain
rule for differentiation, which results in two terms involving the gradient of the
function and the gradient of the distribution. In the final line \eqref{eq:hybrids}, we apply the
identity for score functions \eqref{eq:log_deriv_identity} to replace the
gradient of the probability with the gradient of the log-probability, which
leaves us with two terms that we recognise as the two major gradient estimators
we have explored in this paper. We note that estimators for both terms in \eqref{eq:hybrids} can
incorporate variance reduction techniques such as control variates.
We now explore two approaches that fit within this hybrid estimator framework.

\subsubsection{Weak Reparameterisation}
Sampling paths of the type we just described, which have parameter-dependent base
distributions $p(\vepsilon; \vtheta)$, are referred to by
\citet{ruiz2016generalized} as weak reparameterisations.

\begin{gradexample}{Gamma Weak Reparameterisation}
The Gamma distribution $\mathcal{G}(x; \alpha, \beta)$ with shape parameter $\alpha$
and rate $\beta$ can be represented using the weak reparameterisation \citep{ruiz2016generalized}
\begin{subequations}
\begin{equation}
\epsilon = g^{-1}(x; \alpha, \beta) = \frac{\log(x) - \psi(\alpha) + \log(\beta)}{\sqrt{\psi_1(\alpha)}};
\end{equation}
\begin{equation}
p(\epsilon; \alpha, \beta) = \frac{e^{\alpha
		\psi\left(\alpha\right)}\sqrt{\psi_1{(\alpha)}}}{\Gamma(\alpha)}
	\exp\left(\epsilon \alpha \sqrt{\psi_1(\alpha)} -
		\exp\left(\epsilon \sqrt{\psi_1(\alpha)} + \psi(\alpha)\right)
		\right), \label{eq:case_studies_gamma_weak}
\end{equation}
\end{subequations}
where $\psi$ is the digamma function, and $\psi_1$  is its derivative.
The final density \eqref{eq:case_studies_gamma_weak} can be derived by using
change of variable formula~\eqref{eqn:density_change_of_variable}, and shows
what is meant by weak dependency, since it depends on shape parameter $\alpha$
but not on the rate parameter $\beta$ .
\end{gradexample}

Using knowledge of the existence of a weak
reparameterisation for the measure we can rewrite \eqref{eq:hybrids}. We use two
shorthands:
$\ell(\vepsilon;
\distParams)  := \nabla_{\distParams} g(\vepsilon; \distParams)$ for the
gradient of the weak-path with respect to the parameters, and $ u(\vepsilon;
\distParams)  := \nabla_{\distParams} \log \detchangevar$ for the gradient of
the logarithm of the Jacobian-determinant. The pathwise term in \eqref{eq:hybrids} becomes
\begin{subequations}
\begin{align}
\int p_\distParams(\vepsilon) \nabla_{\distParams} \cost(g(\vepsilon; \distParams)) \intd \vepsilon & =
\int p_\distParams(g(\vepsilon; \distParams)) \detchangevar \nabla_{\distParams} \cost(g(\vepsilon;
\distParams)) \intd \vepsilon \\
& = \int p_\distParams(\vx) \nabla_{\distParams} \cost(\vx) \intd \vx =
\int p_\distParams(\vx) \nabla_{\vx} \cost(\vx) \nabla_{\distParams} \vx d\vx \\
&=  \expect{p_\distParams(\vx)}{\nabla_{\vx} \cost(\vx) \ell(g^{-1}(\vx; \distParams) ; \distParams)},
\end{align}
\end{subequations}
where we first applied the rule for the change of variables to rewrite the
integrals in terms of the variable $\vx$. Introducing the infinitesimals for
$\intd\vx$ cancels the determinant terms, after which we apply the chain rule and finally
rewrite it in the more familiar expectation form in the final line. Similarly, the score
function term in \eqref{eq:hybrids} can be written as
\begin{subequations}
\begin{align}
& \int p_\distParams(\vepsilon) \cost(g(\vepsilon; \distParams)) \nabla_{\distParams} \log p_\distParams(\vepsilon) \intd \vepsilon \\
&= \int p_\distParams(g(\vepsilon; \distParams)) \detchangevar \cost(g(\vepsilon; \distParams)) \nabla_{\distParams} \left( \log
p_\distParams(g(\vepsilon; \distParams)) + \log \detchangevar \right) \intd \vepsilon  \\
&= \int p_\distParams(g(\vepsilon; \distParams)) \detchangevar \cost(g(\vepsilon; \distParams)) \left( \nabla_{\vx} \log
p_\distParams(\vx) \nabla_{\distParams}g(\vepsilon; \distParams) + 
\nabla_{\distParams} \log p_\distParams\left(\vx|_{\vx = g(\vepsilon; \distParams)}\right) + u(\vepsilon;
\distParams) \right) \intd \vepsilon  \nonumber \\
&=\expect{ p_\distParams(\vx)}{\cost(\vx)\left(\nabla_{\vx} \log p_\distParams(\vx) \ell(g^{-1}(\vx; \distParams) ;
	\distParams) + \nabla_{\distParams} \log p_\distParams(\vx) + u(g^{-1}(\vx; \distParams) ; \distParams) \right) },
\end{align}
\end{subequations}
where we again use the rule for the change of variables in the second line wherever $p(\vepsilon)$
appears. On the next line we enumerate the two ways in which $p_\distParams(g(\vepsilon; \distParams))$
depends on $\distParams$ when computing the gradient w.r.t.~$\distParams$. We obtain the final line by
expressing the integral as an expectation w.r.t. $\vx$, keeping in mind that $\intd\vx = \detchangevar \intd\vepsilon$.

\subsubsection{Gradients with Rejection Sampling}
Rejection sampling offers another way of exploring the decoupling of sampling
from gradient estimation in pathwise estimation.
\citet{naesseth2017reparameterization} developed the use of rejection sampling in
the setting of the hybrid gradient \eqref{eq:hybrids}, which we will now present.

Using rejection sampling, we can always
generate samples from a distribution $\dist_\distParams(\vx)$  using a proposal
distribution $r_\distParams(\vx)$ for which $\dist_\distParams(\vx) <
M_\distParams r_\distParams(\vx)$ for a finite constant $M_\distParams$. If we
also know a parameterisation of the proposal distribution $r_\distParams(\vx)$ in terms
of
a sampling path $\vx = g(\vepsilon; \distParams)$ and a base distribution
$\vepsilon \sim p(\vepsilon)$, we can generate samples using a simple procedure:
we repeat the three steps of
drawing a sample $\vepsilon \sim p(\vepsilon)$, evaluating $\vx = g(\vepsilon; \distParams)$, and drawing a uniform variate
$u \sim \mathcal{U}[0, 1]$, until $u < \frac{\dist_\distParams(\vx)}{M_\distParams r_\distParams(\vx)}$, at which point
we accept the corresponding $\vx$ as a sample from $\dist_\distParams(\vx)$.

This algorithmic process of rejection sampling induces an effective base distribution
\begin{align}
\pi(\vepsilon) & = p(\vepsilon) \frac{\dist_\distParams\left(g(\vepsilon; \distParams)\right)}{r_\distParams\left(g(\vepsilon;
	\distParams)\right)}, \label{eq:case_rejection_effectivedist}
\end{align}
which intuitively reweights the original base distribution $p(\vepsilon)$ by the ratio of the measure $p_{\vtheta}$ and the proposal $r_{\vtheta}$; see \citet{naesseth2017reparameterization} for a detailed derivation.
We can now reparameterise $p_{\vtheta}(\vx)$ in our objective \eqref{eq:expectation_function} using the proposal's sampling path $g(\vepsilon; \vtheta)$ and the effective distribution \eqref{eq:case_rejection_effectivedist}:
\begin{align}
\expect{\dist_\distParams(\vx)}{\cost(\vx)} =
\int p(\vepsilon) \frac{\dist_\distParams\left(g(\vepsilon; \distParams)\right)}{r_\distParams\left(g(\vepsilon;
	\distParams)\right)}  \cost(g(\vepsilon; \distParams)) \intd \vepsilon =
\int \pi(\vepsilon) \cost\left(g(\vepsilon; \distParams)\right) \intd \vepsilon =
\expect{\pi(\vepsilon)}{\cost(g(\vepsilon; \distParams))}.
\label{eq:ar_expectations}
\end{align}
The gradient \eqref{eq:hybrids} can be rewritten using the effective
distribution $\pi(\vepsilon)$ instead, giving another approach for gradient
estimation:
\begin{subequations}
\begin{align}
\nabla_{\distParams} \expect{\dist_\distParams(\vx)}{\cost(\vx)} =
\nabla_{\distParams} \expect{\pi(\vepsilon)}{\cost(g(\vepsilon; \distParams))}
= \nabla_{\distParams} \int \pi(\vepsilon) \cost(g(\vepsilon; \distParams)) \intd \vepsilon \\
= \underbrace{\expect{\pi_\distParams(\vepsilon)}{\nabla_{\distParams}{\cost(g(\vepsilon;
\distParams))}}}_{\text{pathwise estimation}} +
\underbrace{\expect{\pi_\distParams(\vepsilon)}{{\cost(g(\vepsilon; \distParams))} \nabla_{\distParams}{\log
\frac{\dist_\distParams\left(g(\vepsilon; \distParams)\right)}{r_\distParams\left(g(\vepsilon;
\distParams)\right)}}  }}_{\text{score-function estimation}}.
\end{align}
\end{subequations}
To minimise the magnitude of the score-function term, the proposal distribution
$r_\distParams(\vx)$ should be chosen so that it yield a high acceptance probability.

\subsection{Empirical Comparisons}
\label{sect:blr_experiments}
Logistic regression remains one of the most widely-used statistical models, and
we use it here as a test case to explore the empirical performance of
gradient estimators and to comment on their implementation and interactions with
the optimisation procedure.
Bayesian logistic regression defines a probabilistic model for a target $y_i
\in \{-1,1\}$ given features (or covariates) $\vx_i \in \mathbb{R}^D$ for data points $i = 1,
\ldots, I$, using a set of parameters $\vw$. We denote the collection of all $I$
data points as $\{\vX, \vy\}$. Using a Gaussian prior on the parameters and a
Bernoulli likelihood, the probabilistic model is
\begin{equation}
p(\vw) = \mathcal{N}(\vw | \vzero, c\vI); \qquad p(y_i | \vx_i, \vw) = \sigma\left(y_i\vx_i^\top\vw\right),
\end{equation}
where $\sigma(z)= \left(1 + \exp(-z)\right)^{-1}$ is the logistic function, and
$c$ is a constant. We use variational Bayesian logistic regression, an
approximate Bayesian inference procedure, to determine the parameter posterior
distribution $p(\vw | \vX, \vy)$ \citep{jaakkola1997variational,
	murphy2012machine}. Variational inference was one of the five research areas
that we reviewed in Section \ref{sect:stoch_opt} to which gradient estimation is
of central importance, allows us to specify a bound on the
integrated likelihood of the logistic regression model, maximising which requires
gradients of the form of \eqref{eq:gradient}. Variational inference introduces a
variational distribution $q(\vw; \vtheta) = \mathcal{N}(\vw| \vmu, \vSigma)$,
with variational parameters $\vtheta = \{\vmu, \vSigma\}$ that are optimised
using a  variational lower bound objective \citep{jaakkola1997variational}
\begin{align}
& \mathcal{L}(\vtheta) = \sum_{i=1}^{I}\expect{q(\vw | \vmu, \vSigma)}{\log
\sigma\left(y_i\vx_i^\top\vw\right)} -
\KLpq{q(\vw | \vmu, \vSigma)}{ p(\vw)}.
\label{eqn:log-regression-elbo}
\end{align}
The first expectation is a problem of the form we have studied throughout this
paper, which is the expectation of a cost function, given by a log-likelihood
under a Gaussian measure; and we are interested in Monte Carlo gradients of the
variational parameters $\vtheta$. The second term is  the Kullback-Leibler (KL)
divergence between the variational posterior $q$ and the prior distribution $p$,
which is known in closed-form for Gaussian distributions;  in other models, the KL will not be known in closed-form, but its gradients can still be computed using the estimators we discuss in this paper.  We optimise the
variational bound using stochastic gradient descent, and compare learning of the
variational parameters using both the score-function and pathwise estimator. The
objective function we use is a Monte Carlo estimator of Equation \eqref{eqn:log-regression-elbo}:
\begin{align}
\bar{\mathcal{L}}(\vtheta) = \frac{I}{B} \sum_{i=1}^{B} \frac{1}{N} \sum_{n=1}^{N}{\log
\sigma\left(y_{p_i}\vx_{p_i}^\top\hat\vw_n\right)} -
&\KLpq{\mathcal{N}(\vw | \vmu, \vSigma)}{\mathcal{N}(\vw | \vzero, c\vI)}; \\
&\hat\vw_n \sim \mathcal{N}(\vw | \vmu, \vSigma),\ p_i \sim \{1, \dots, I\}.
\end{align}
where $B$ is the data batch size, $I$ is the size of the full data set,
$N$ is the number of samples taken from the
posterior distribution to evaluate the Monte Carlo gradient, and the posterior covariance is a
diagonal matrix $\vSigma = \textrm{diag}(\vs)$. We use stochastic gradient
descent for optimisation, with cosine learning rate decay~\citep{cosine_decay} unless stated
otherwise. We use the UCI Women's Breast Cancer data set~\citep{uci}, which has
$I = 569$ data points and $D=31$ features. For evaluation, we always use the
entire data set and $1000$ posterior samples. Using this setup, we tie together
the considerations raised in the previous sections, comparing gradient
estimators when used in the same problem setting, looking at the interaction between the
Monte Carlo estimation of the gradient and the mini-batch optimisation, and
reporting on the variance of estimators under different variance reduction
schemes. Generalised regression using the pathwise and score
function estimators have been explored in several papers including
\citet{paisley2012variational, kucukelbir2017automatic, salimans2013fixed,
	wang2013variance, miller2017reducing}.

\begin{figure}[tp]
	\centering
	\includegraphics[width=0.48\textwidth]{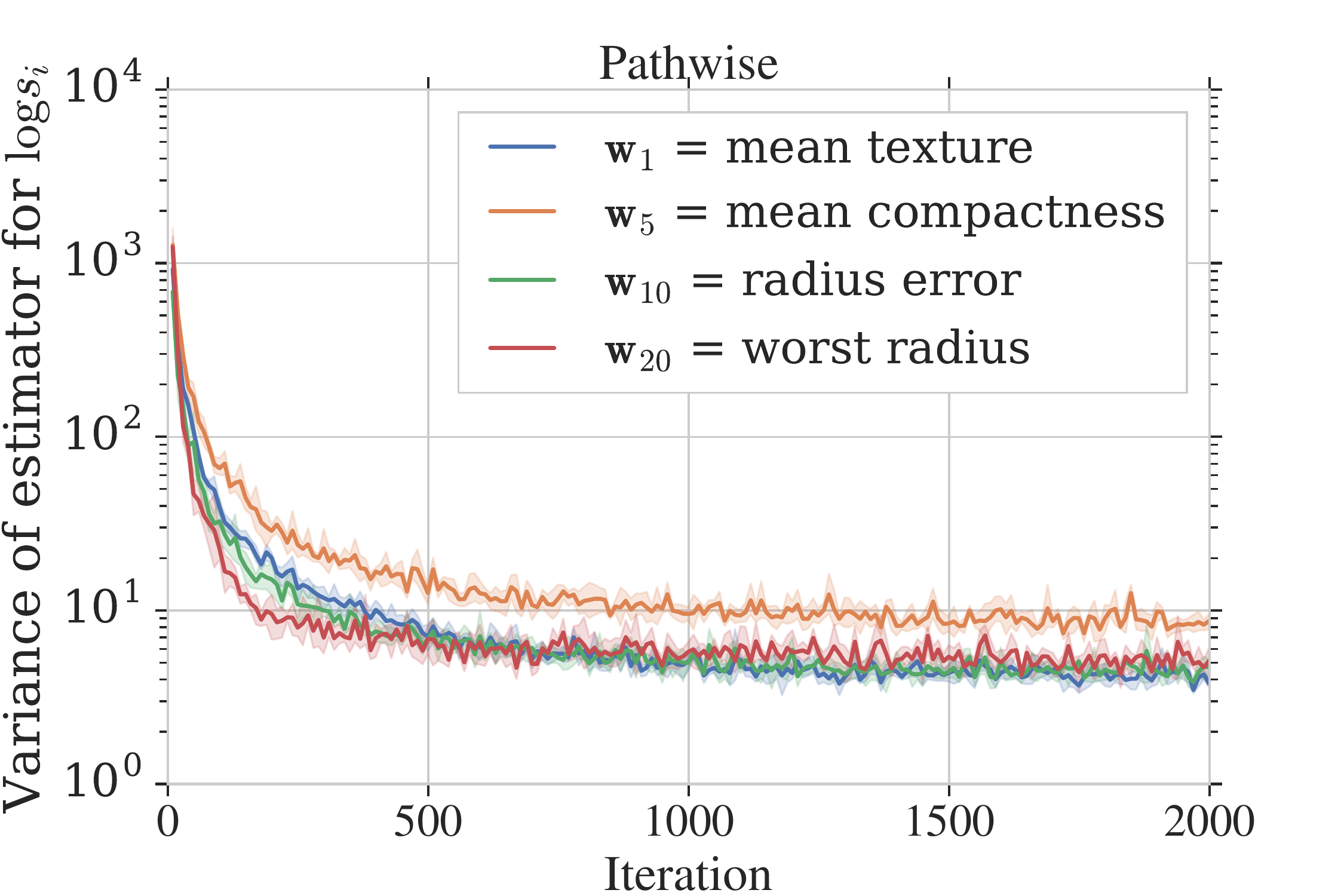}
	\includegraphics[width=0.48\textwidth]{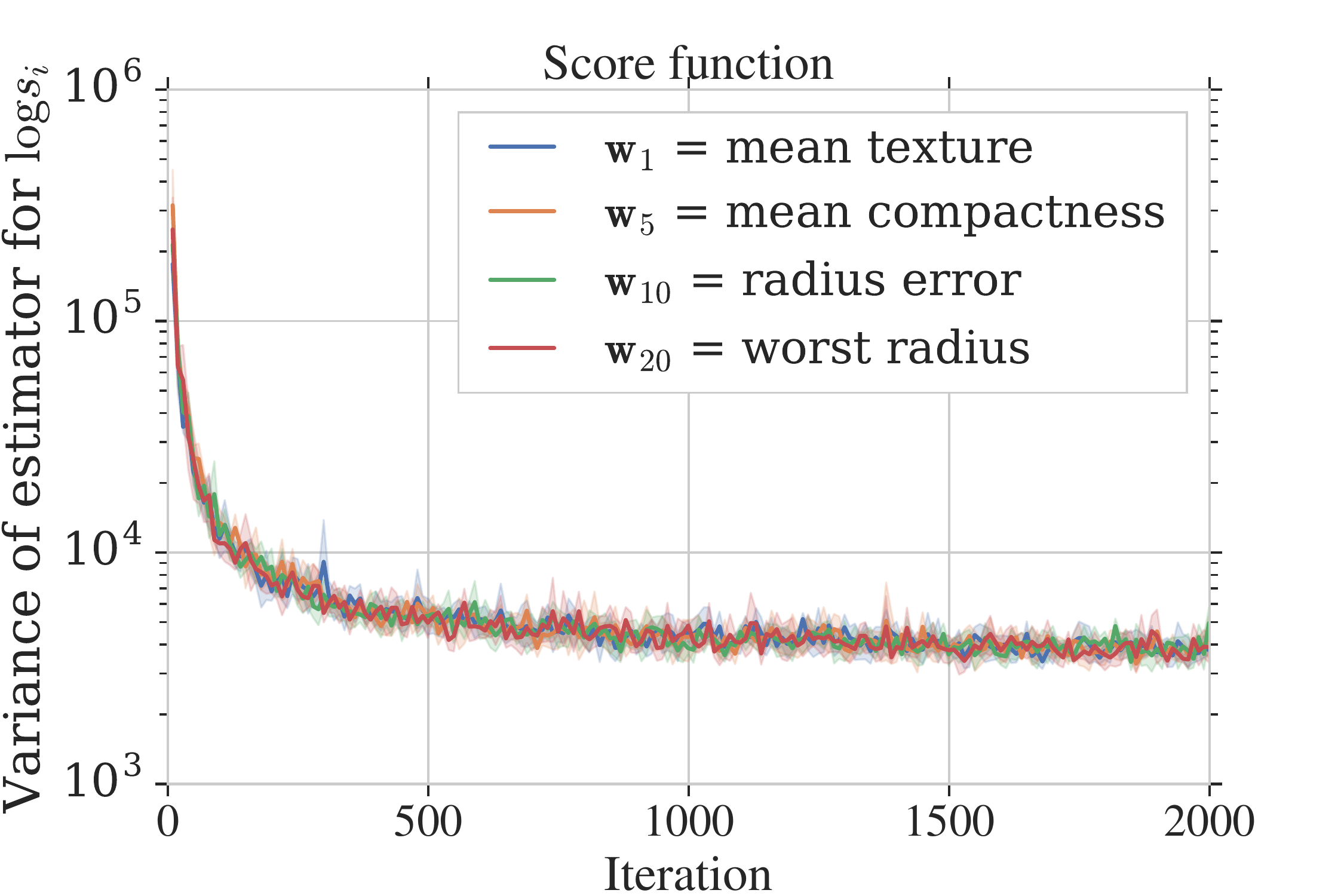}
	\caption{Progress of gradient variance $\Var_{q(\vw;
			\vs)}[\nabla_{\log s}f(\vw)]$ for different parameters $i$ as training
	progresses; batch size $B=32$, the number of samples $N=50$. The legend shows
	the feature index and its name in the data set. }
	\label{fig:blr_grad_var_time}
\end{figure}

\begin{figure}[t]
	\centering
	\includegraphics[width=0.48\textwidth]{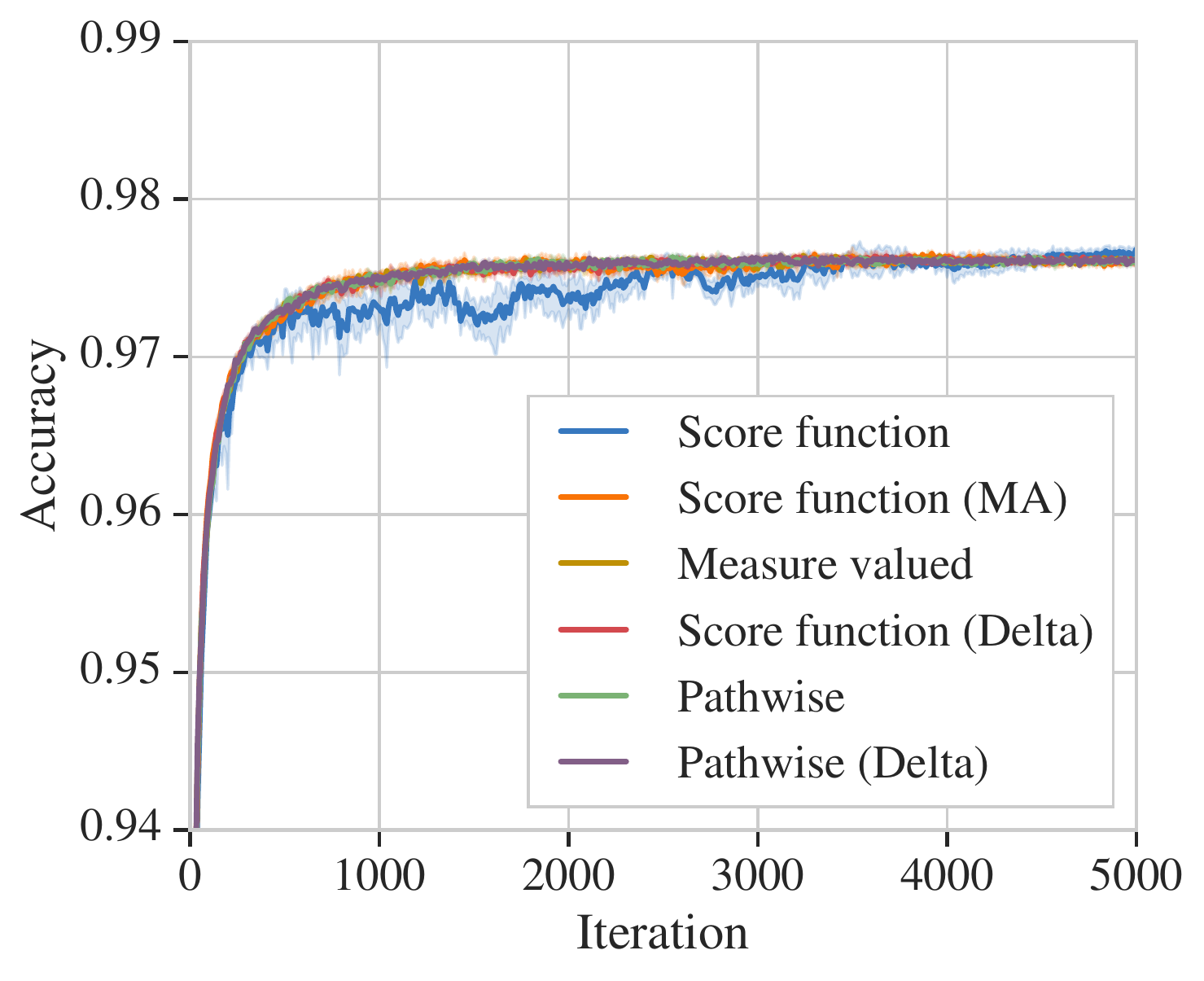}
	\includegraphics[width=0.48\textwidth]{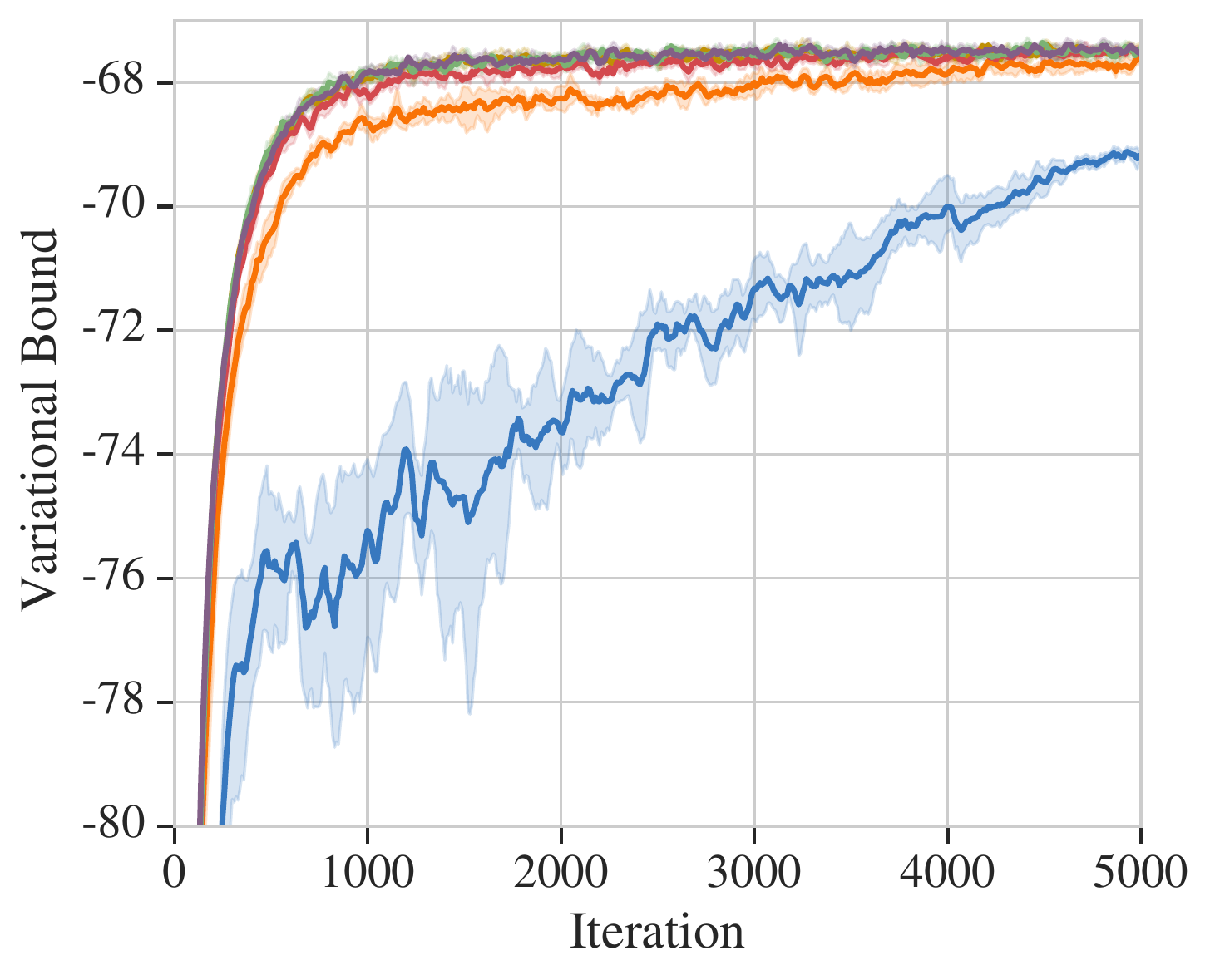}\\
	\includegraphics[width=0.48\textwidth]{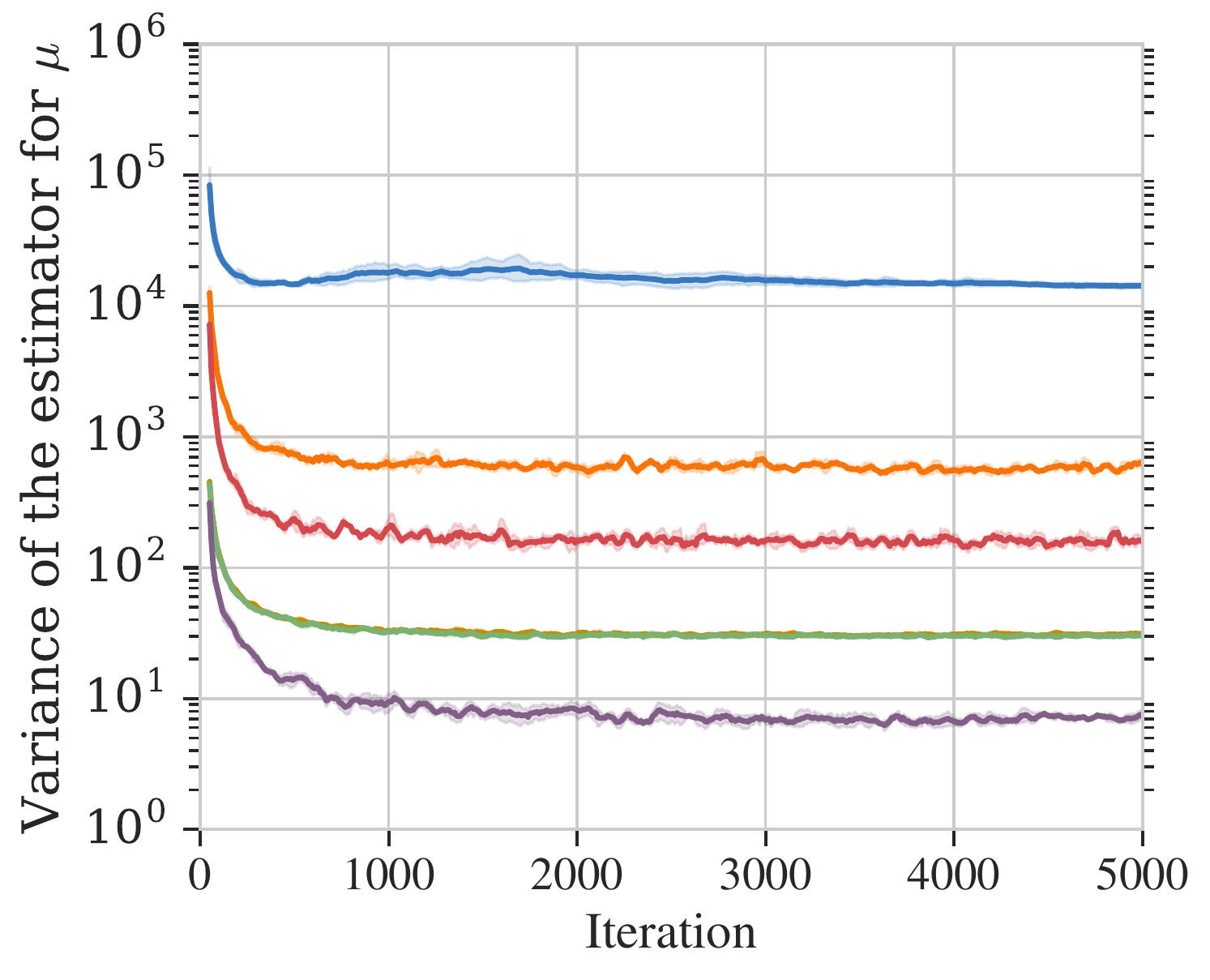}
	\includegraphics[width=0.48\textwidth]{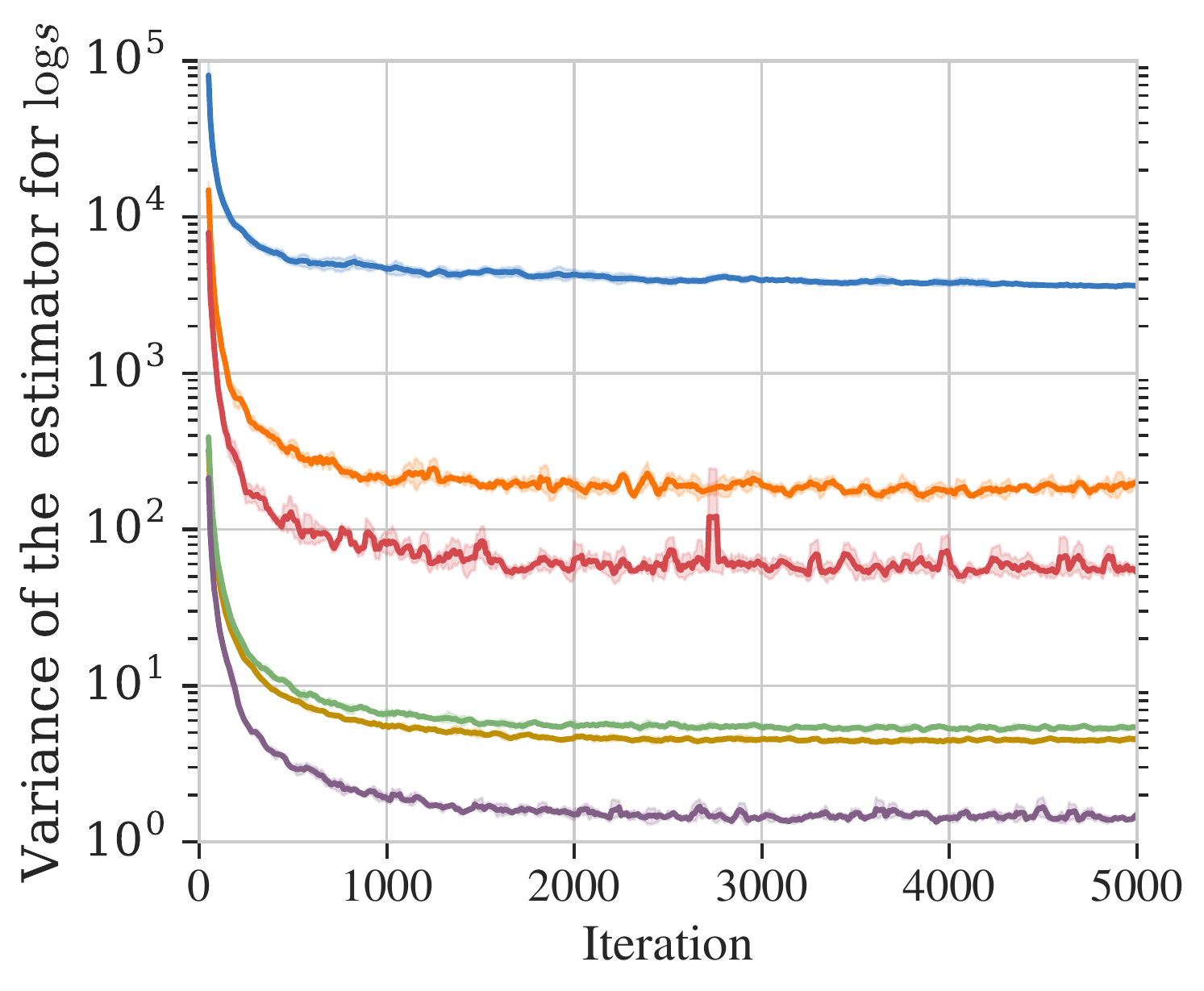}
	\caption{Comparison between the score-function, pathwise, and measure-valued
		estimators at a fixed learning rate ($10^{-3}$). $B=32, N=50$. When a control
		variate is used, it is marked in parenthesis: MA stands for moving average,
		Delta for the second-order delta method. For measure-valued derivatives, we always use coupling.
		In the second row, we show the
		variance of the estimator for the mean $\Var_{q(\vw| \vmu,
			\vs)}[\nabla_{\mu}f(\vw)]$ and the log-standard deviation $\Var_{q(\vw|
			\vmu,\vs)}[\nabla_{\log s}f(\vw)]$, averaged over parameter dimensions.}
	\label{fig:sc_path_1}
\end{figure}

\paragraph{Decreasing variance over training.} Figure
\ref{fig:blr_grad_var_time} shows the variance of the two estimators for the
gradient with respect to the log of the standard deviation $\log s_d$ for four input features. These plots establish the basic behaviour of the optimisation:
different dimensions have different gradient-variance levels; this
variance decreases over the course of the optimisation as the parameters get
closer to an optimum. The same behaviour is shown by both estimators, but the
magnitude of the variance is very different between them, with the pathwise
estimator having much lower variance compared to the score-function.

\begin{figure}[t]
	\centering
	\includegraphics[width=0.48\textwidth]{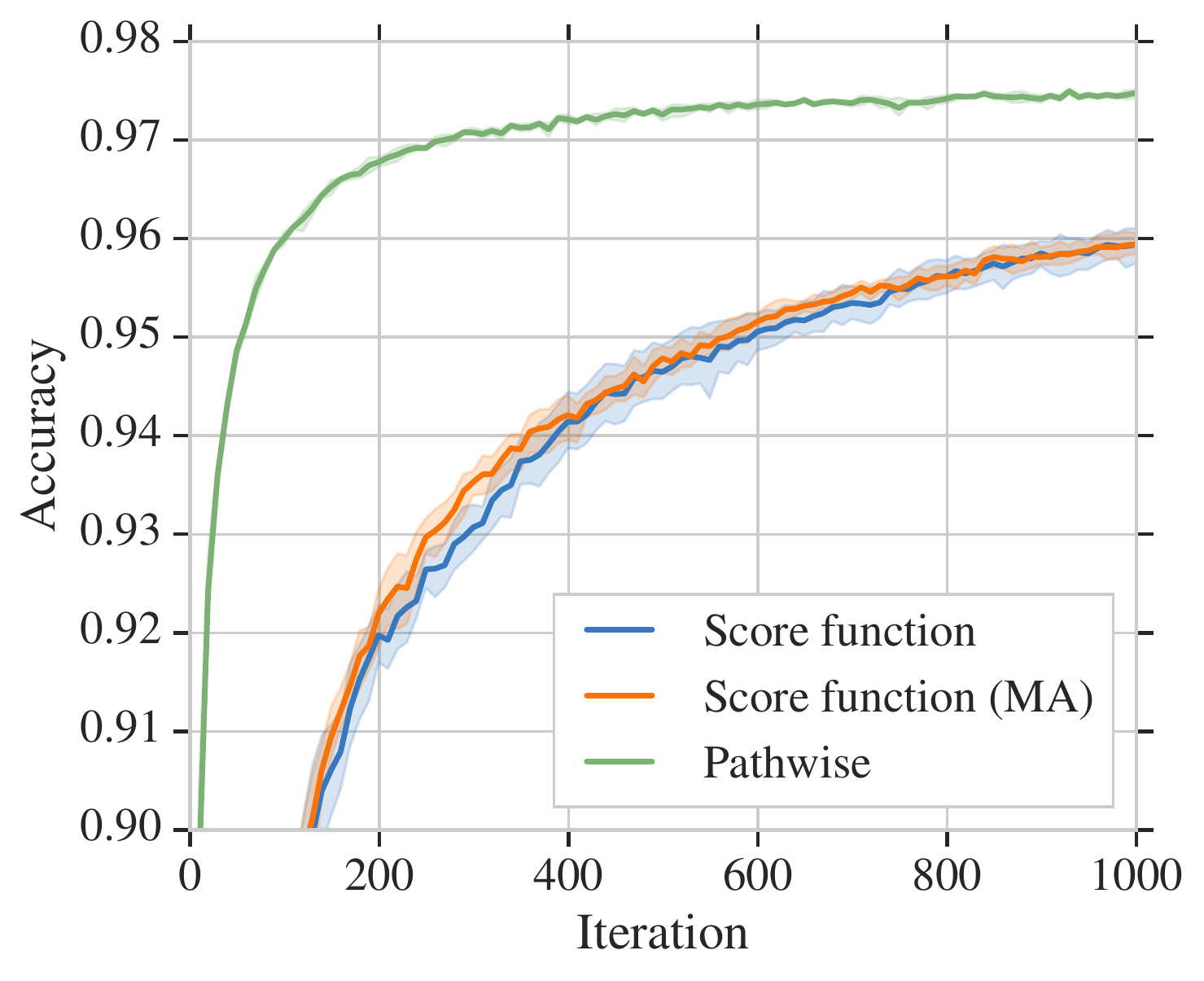}
	\includegraphics[width=0.48\textwidth]{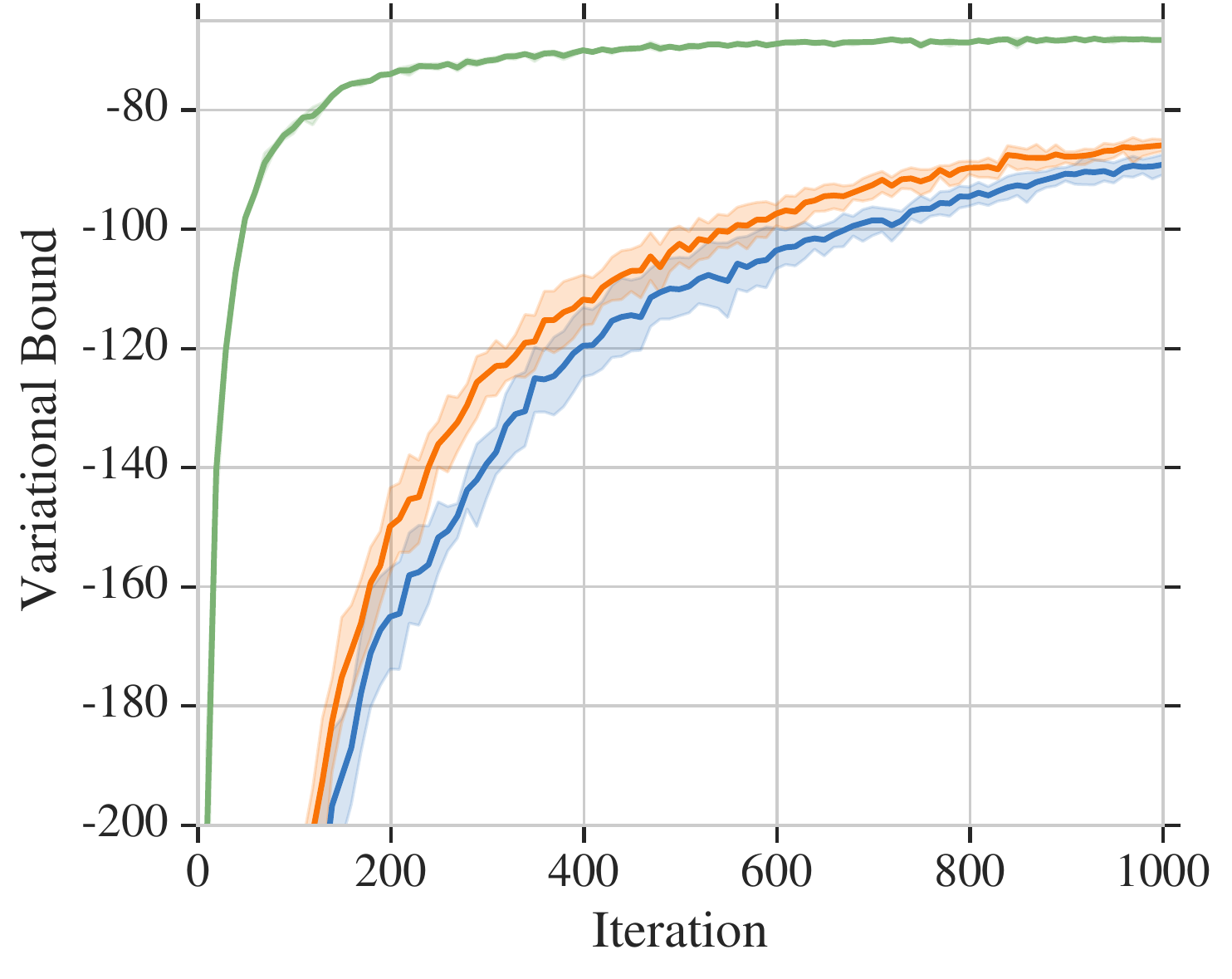}\\
	\includegraphics[width=0.48\textwidth]{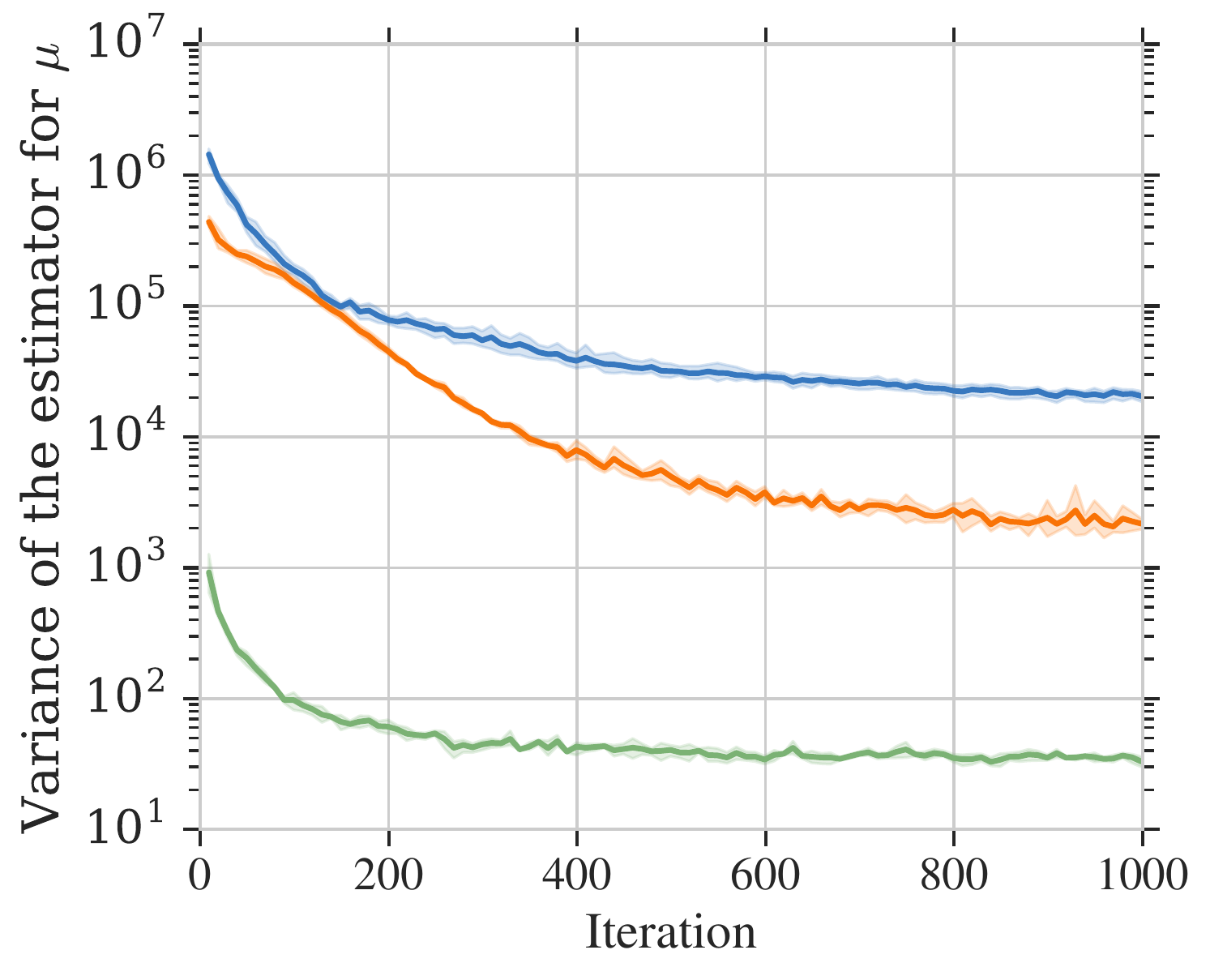}
	\includegraphics[width=0.48\textwidth]{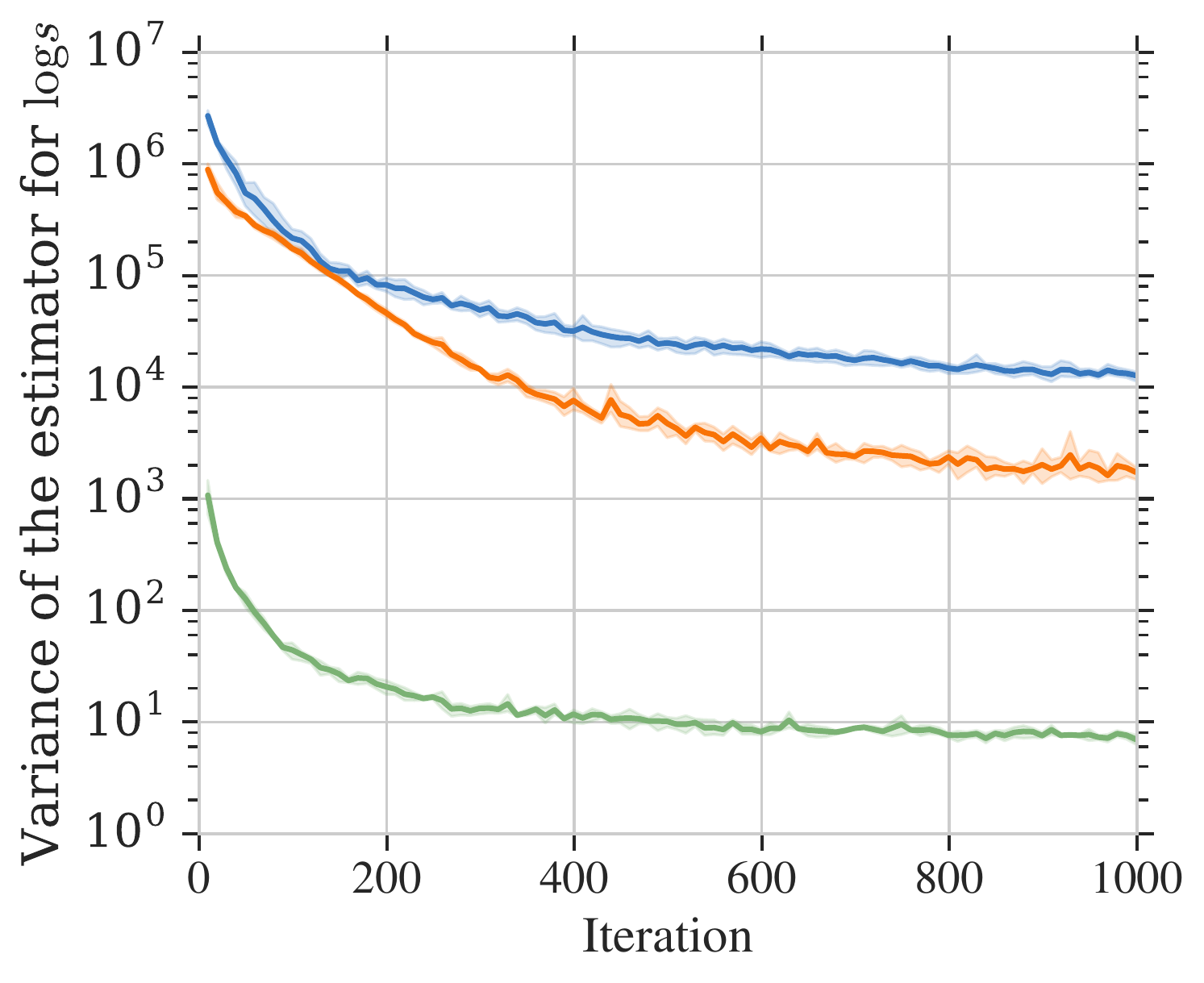}
	\caption{Comparison between the score-function estimator and pathwise
		estimators with a fixed batch size and number of posterior samples $B=32, N=5$.
		The pathwise estimator converges with a higher learning rate ($10^{-3}$), while
		the score function requires a smaller learning rate ($10^{-4}$) to avoid
		divergence due to higher gradient variance. We do not compare with control
		variates that require additional samples, such as the delta method control
		variate. In the second row, we show the variance of the estimator for the mean
		$\Var_{q(\vw| \vmu, \vs)}[\nabla_{\mu}f(\vw)]$ and the log-standard deviation
		$\Var_{q(\vw| \vmu,\vs)}[\nabla_{\log s}f(\vw)]$, averaged over parameter
		dimensions.}
	\label{fig:sc_path_2}
\end{figure}

\paragraph{Performance with fixed hyperparameters.}  We compare the
score-function and pathwise estimators by now fixing the hyperparameters of the
optimisation algorithm, using a batch size of $B=32$ data points, a learning rate of
$10^{-3}$ and estimating the gradient using $N=50$ samples from the measure.
Figure~\ref{fig:sc_path_1} shows the variational lower bound and the accuracy
that is attained in this setting. These graphs also show the control-variate
forms of the gradient estimators.  As control variates, we use the delta method
(Section ~\ref{sect:delta_cv}) and the moving average (MA) baseline
(Section~\ref{sect:cv_designing}). For the delta method control variate, we
always use 25 samples from the measure to estimate the control variate
coefficient using Equation~\eqref{eq:cv_coeff}.

Comparing the different estimators in Figure~\ref{fig:sc_path_1}, the pathwise
estimator has lower variance and faster convergence than the score-function
estimator. The score-function estimator greatly benefits from the use of control
variates, with both the moving average baseline and the delta method reducing
the variance. The delta method control variate leads to greater variance
reduction than a moving average baseline, and reduces variance both for the
score-function and pathwise estimators. We already have a sense from these
plots that where the pathwise estimator is applicable, it forms a good default
estimator since it achieves the same performance level but with lower variance
and faster convergence.

\paragraph{Using fewer samples.} Figure~\ref{fig:sc_path_2} shows a similar
setup to Figure~\ref{fig:sc_path_1}, with a fixed batch size $B=32$, but using
fewer samples from the measure, $N=5$. Using fewer samples affects the speed of
convergence. The score-function estimator converges slower than the pathwise
estimator, since it requires smaller learning rates. And like in the previous
figure, this instability at higher learning rates can be addressed using more
samples in the estimator, or more advanced variance reduction like the delta
method.

\begin{figure}[t]
	\centering
	\includegraphics[width=0.32\textwidth]{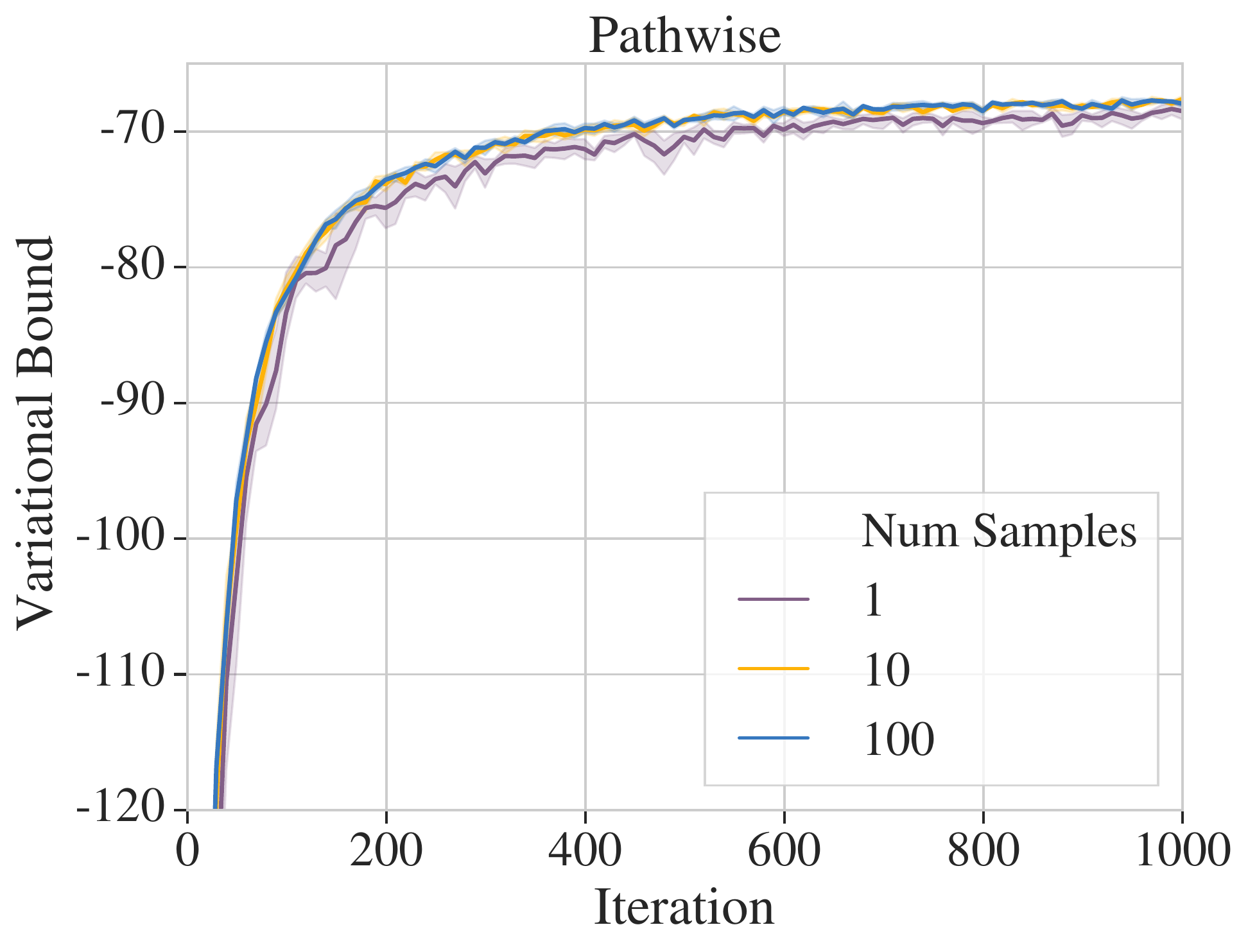}
	\includegraphics[width=0.32\textwidth]{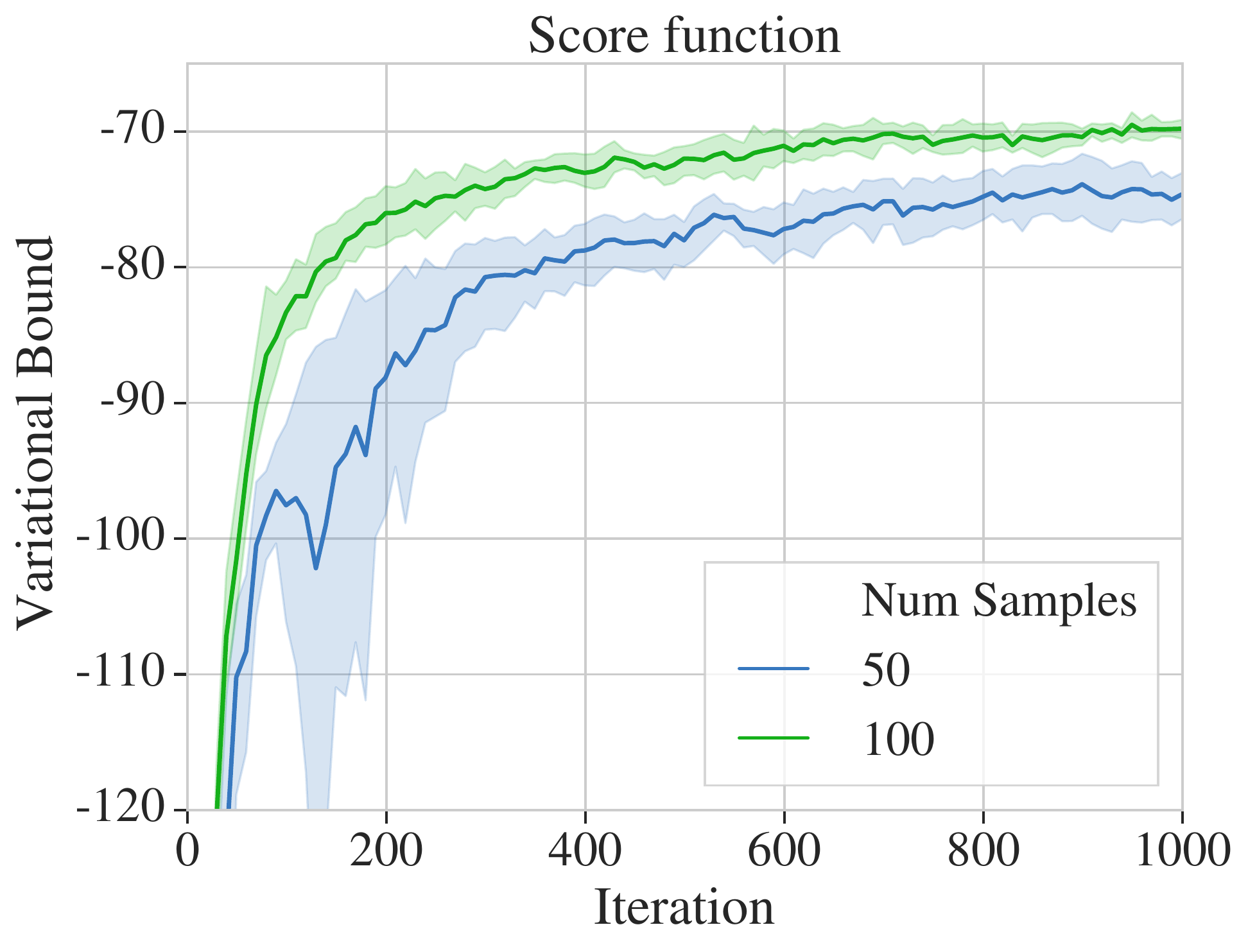}
	\includegraphics[width=0.32\textwidth]{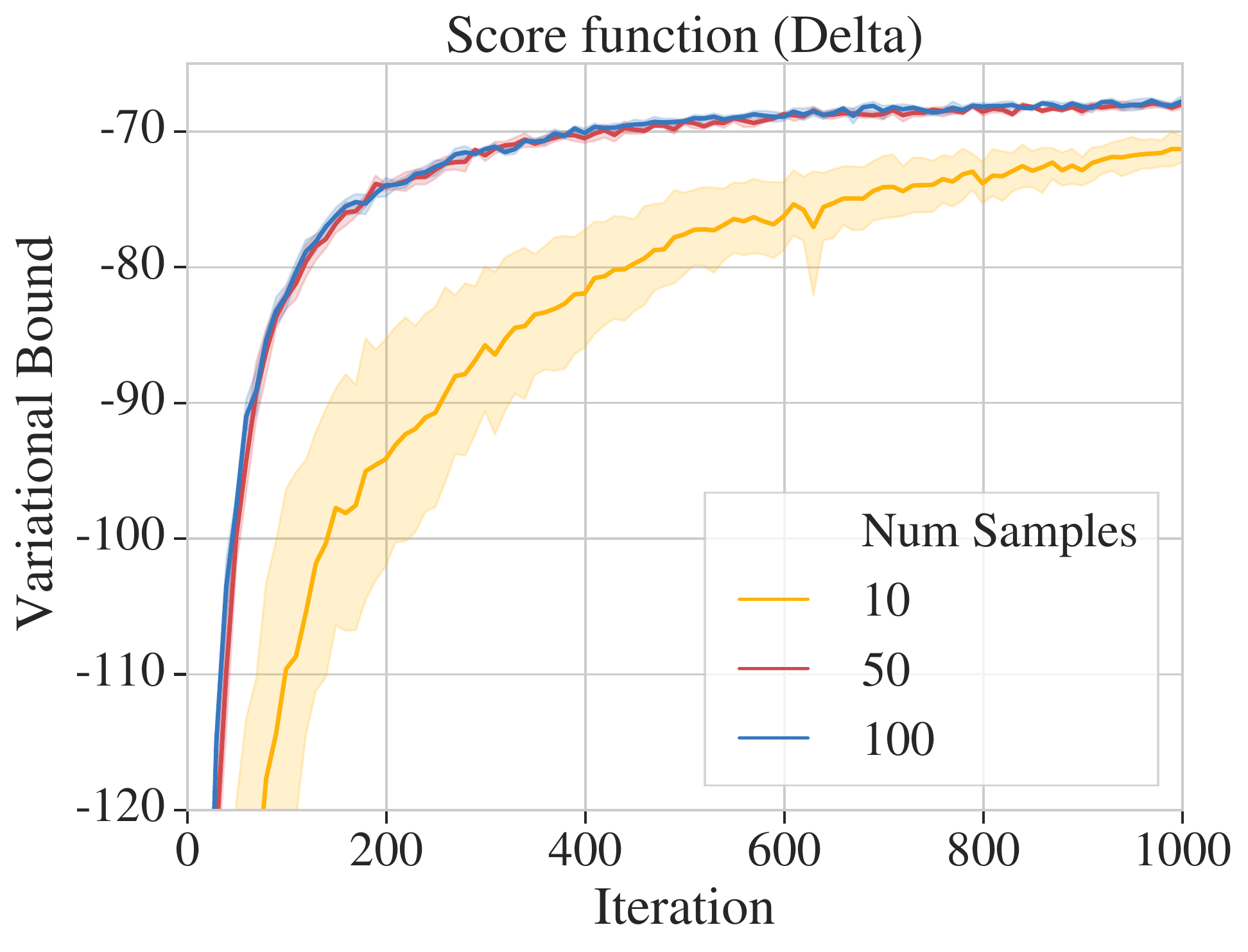}
	\caption{The effect of the number of posterior samples $N$ on the evidence lower
	bound
	for
	different estimators. $B=32$.}
	\label{fig:blr_num_samples}
\end{figure}

\begin{figure}[t]
	\centering
	\includegraphics[width=0.32\textwidth]{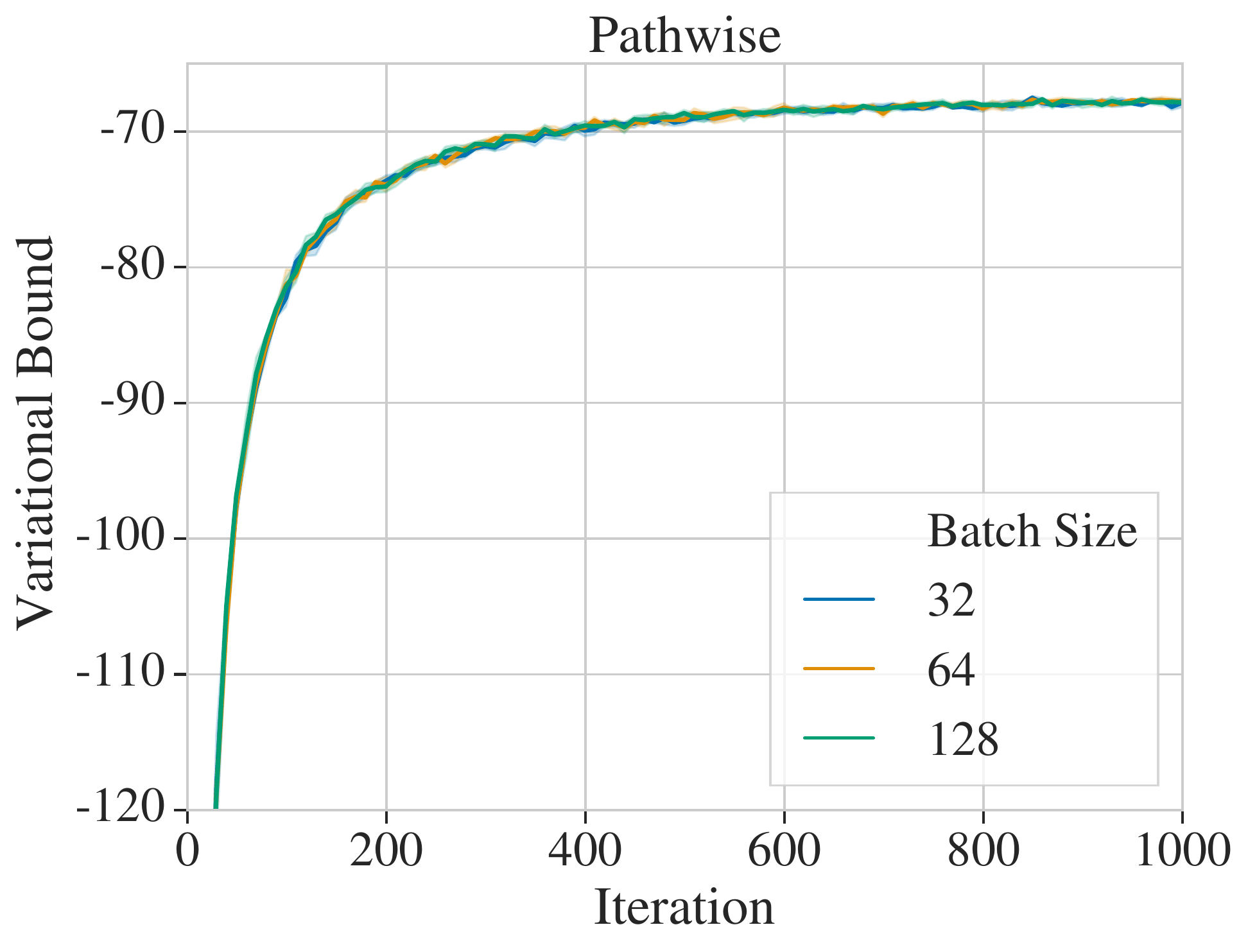}
	\includegraphics[width=0.32\textwidth]{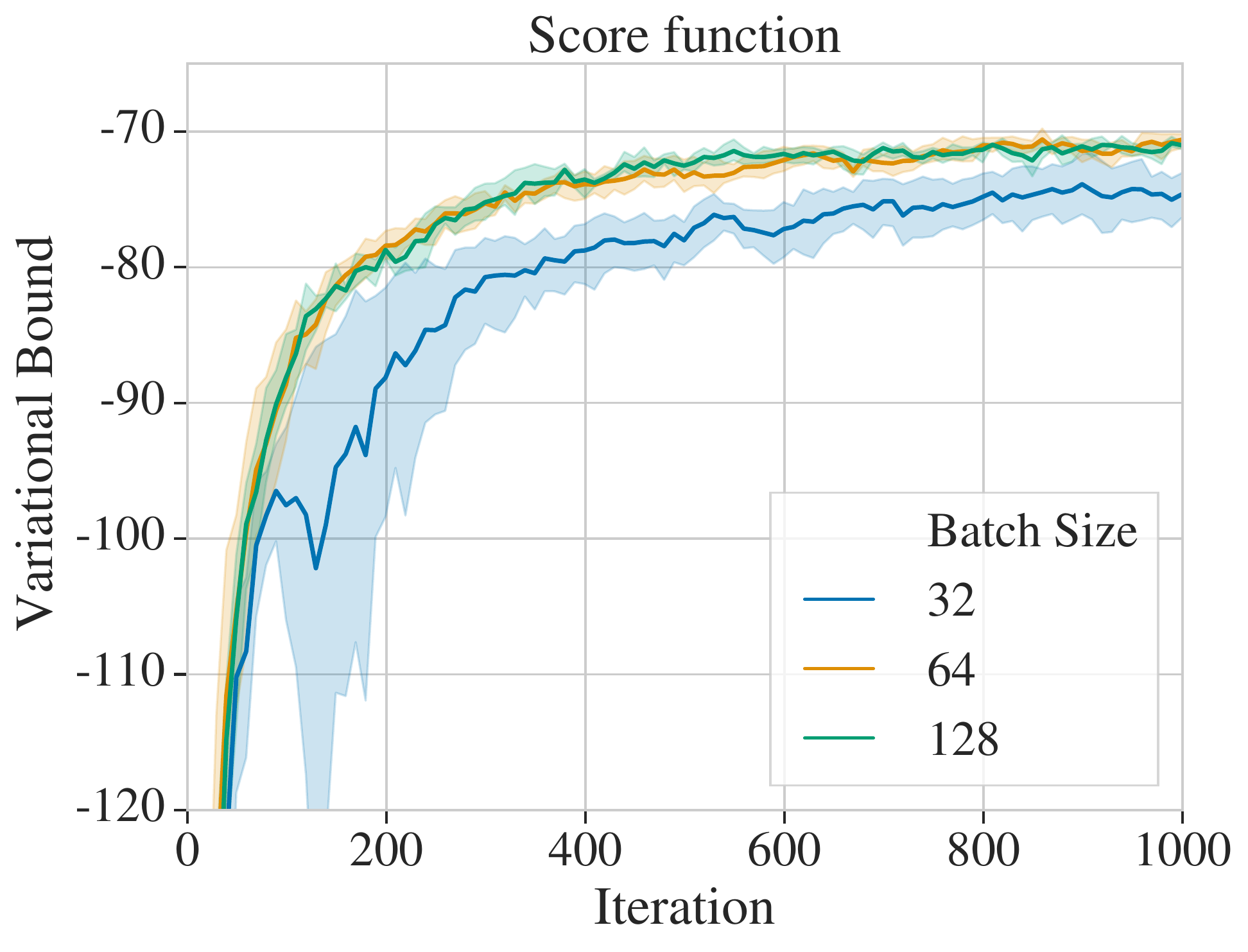}
	\includegraphics[width=0.32\textwidth]{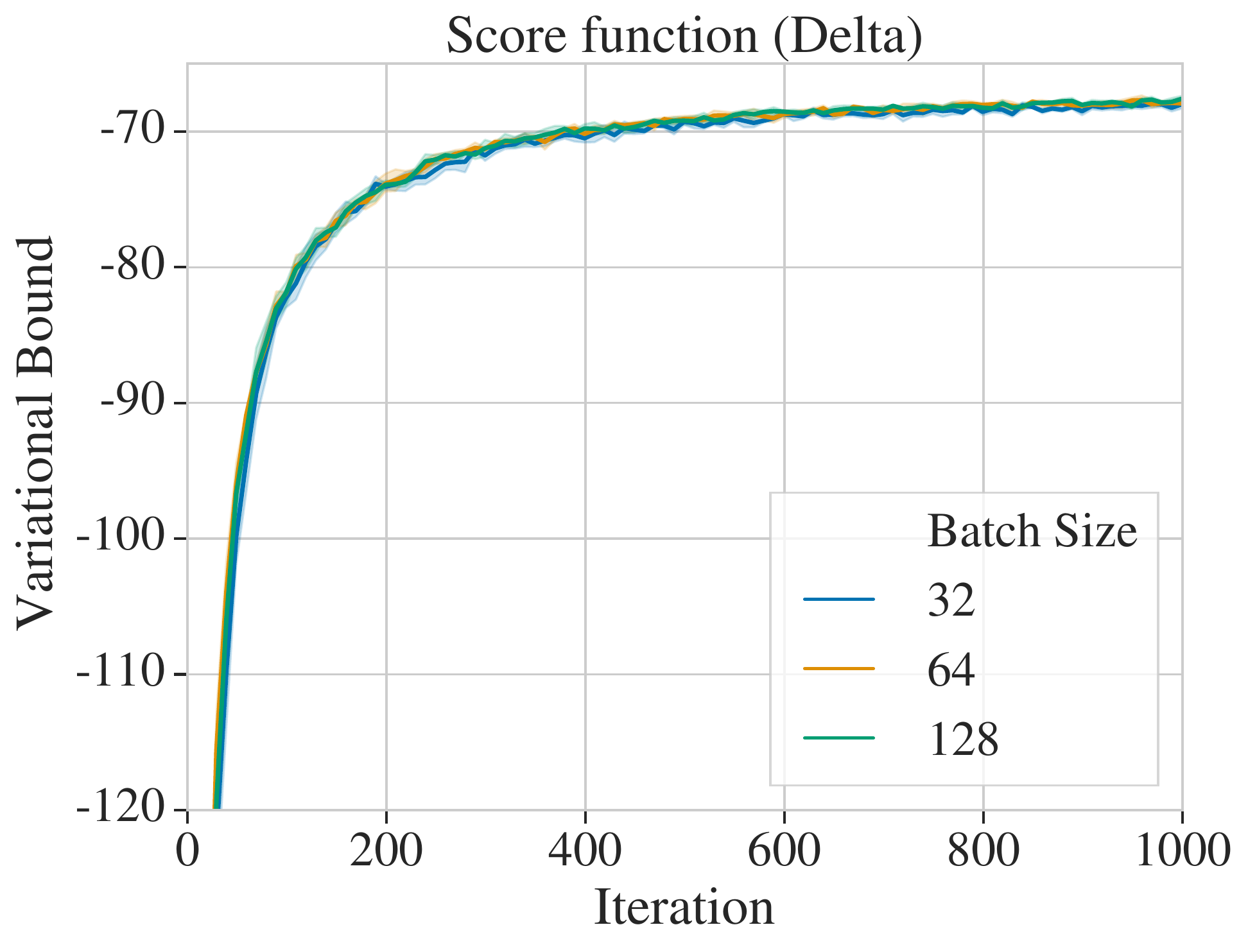}
	\caption{The effect of the number of the batch size $B$ on the evidence lower bound
	for
	different
	estimators with $N=50$.}
	\label{fig:blr_bs}
\end{figure}

\begin{figure}[t]
	\centering
	\includegraphics[width=0.48\textwidth]{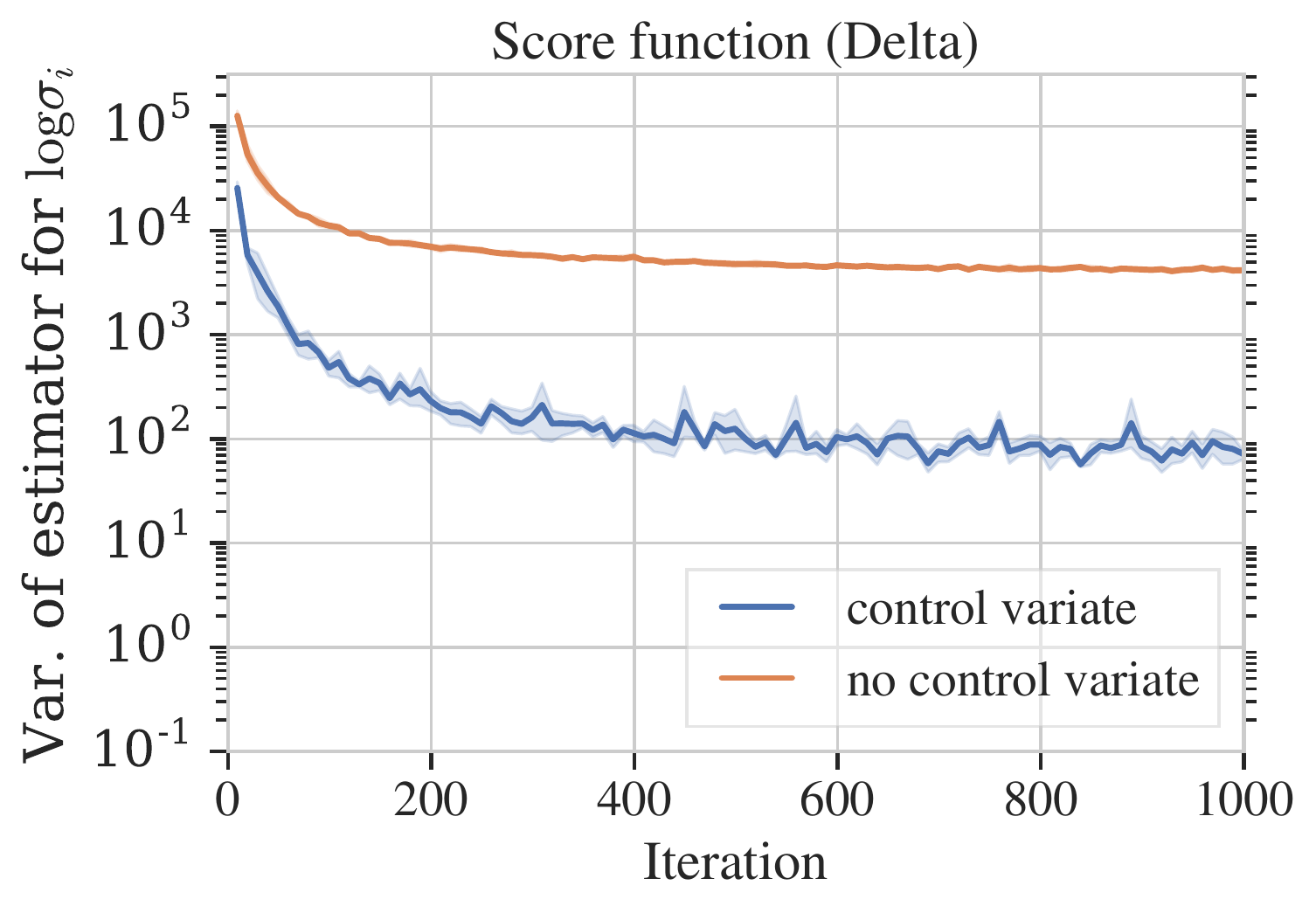}
	\includegraphics[width=0.48\textwidth]{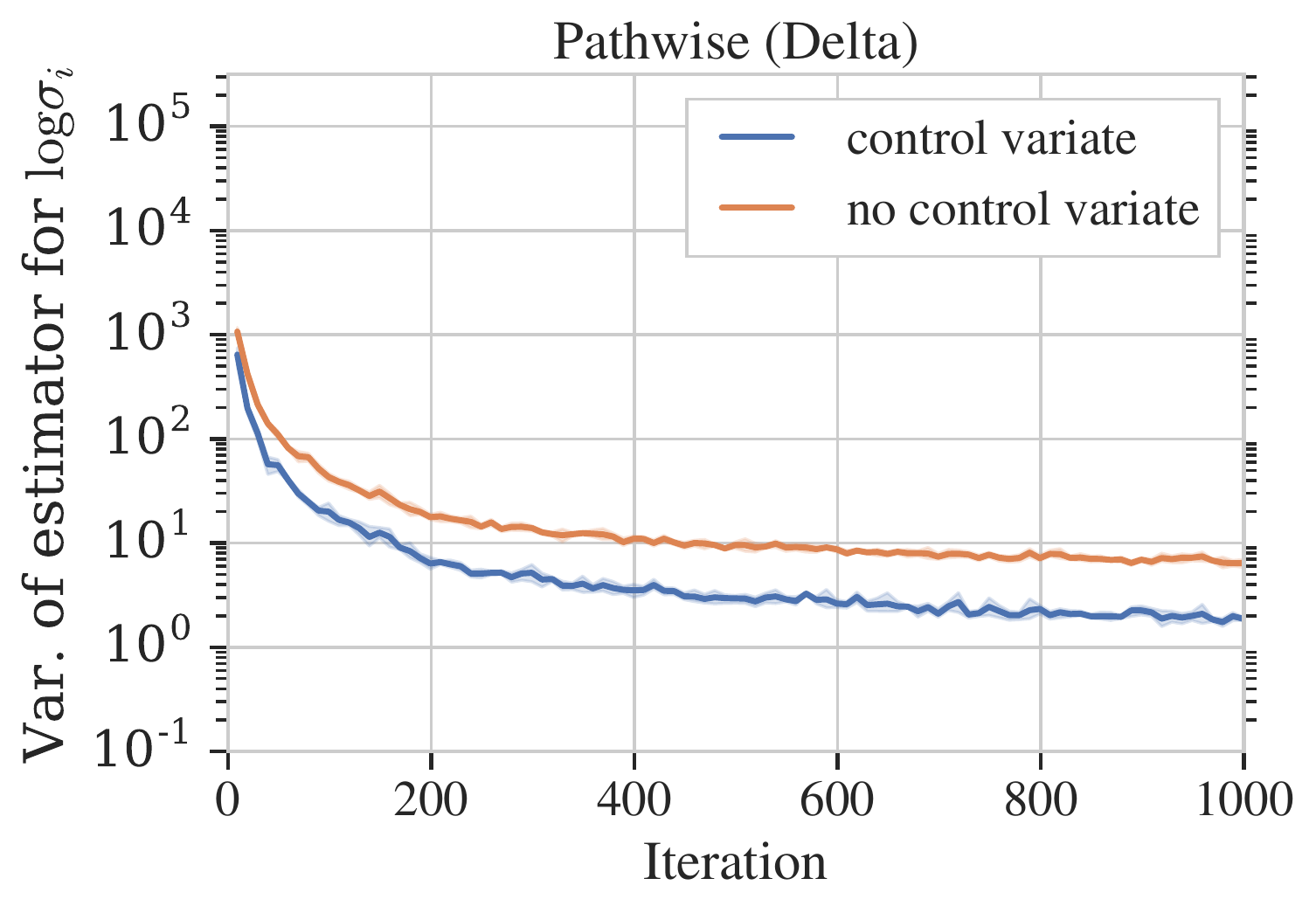}
	\includegraphics[width=0.48\textwidth]{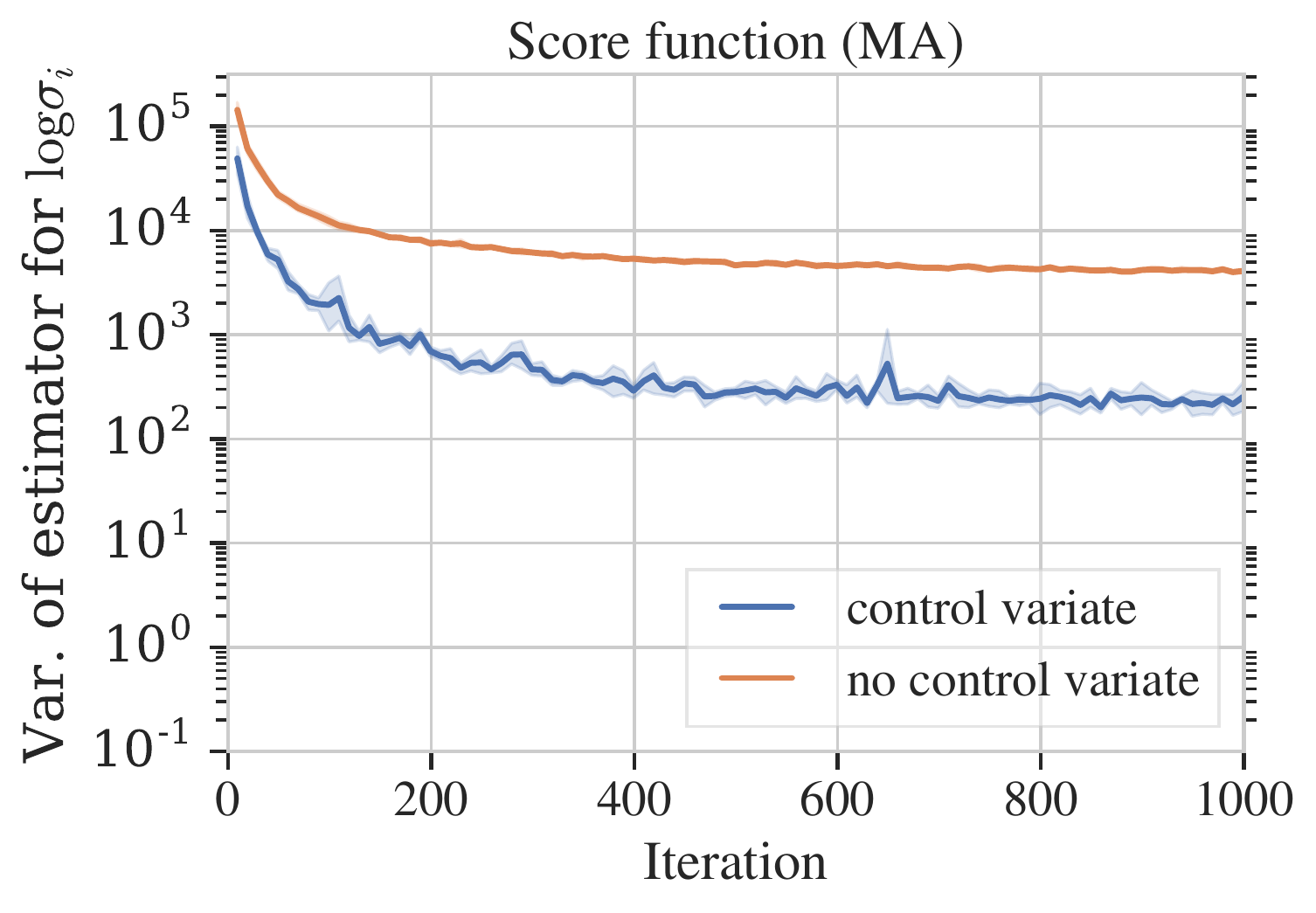}
	\includegraphics[width=0.48\textwidth]{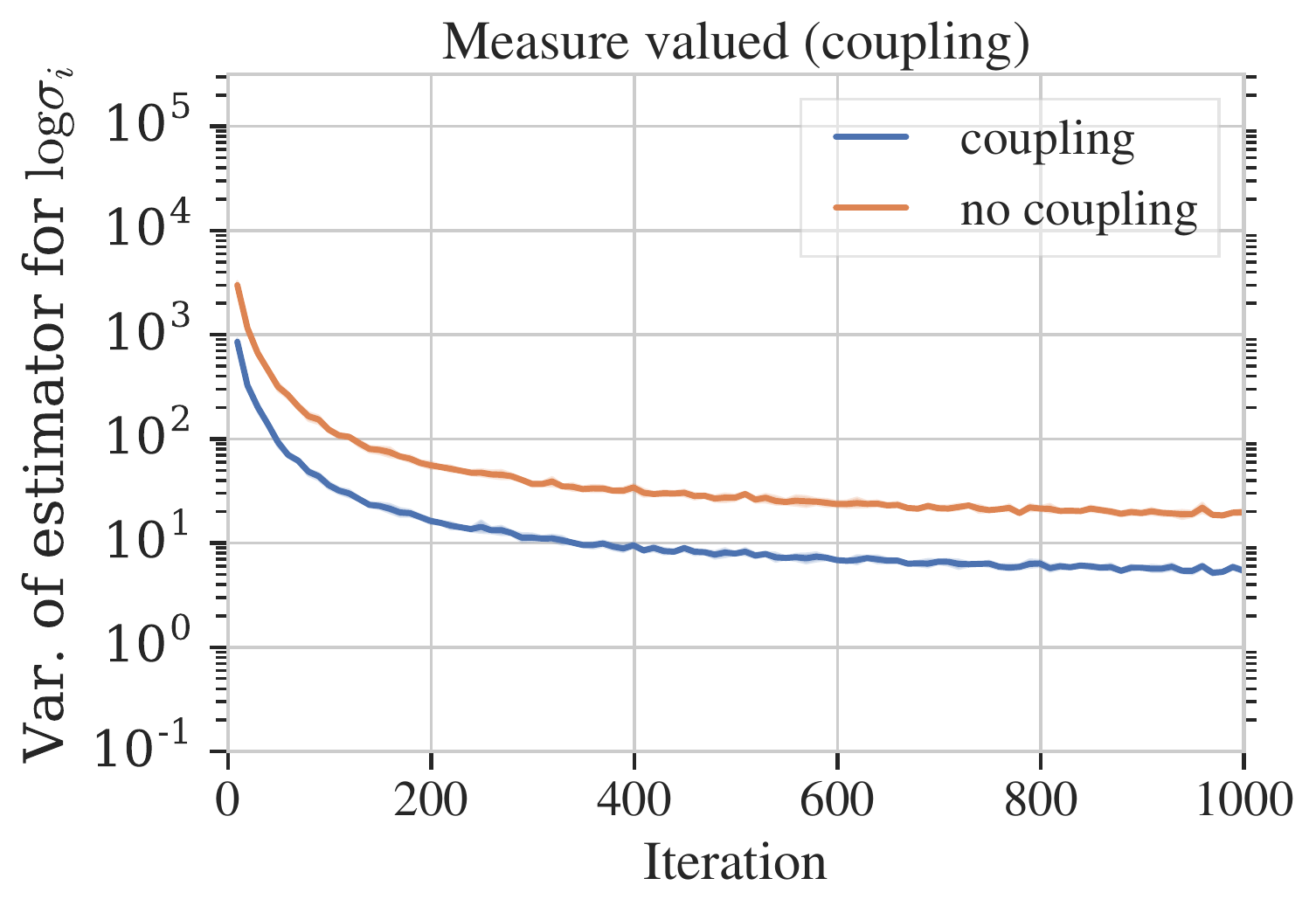}
	\caption{Quantifying the variance reduction effect of control variates. Here,
		each model is trained with a control variate, but at each iteration we assess
		the variance $\Var_{q(\vw| \vmu,\vs)}[\nabla_{\log s}f(\vw)]$ of the same
		estimator for the current parameters with batch size $B=32, N=50$.
		We observed similar variance reduction with respect to $\mu$ for all but the measure-valued estimator, where coupling had no effect on gradient variance.}
	\label{fig:cv_no_cv}
\end{figure}

\paragraph{Effect of batch size and samples.} We now vary the number of samples
from the measure that is used in computing the gradient estimator, and the batch
size, and show the comparative performance in Figures~\ref{fig:blr_num_samples}
and~\ref{fig:blr_bs}. In Figure~\ref{fig:blr_num_samples} there is a
small change in the rate of the convergence of the pathwise estimator as the
number of samples $N$ is increased. In contrast, the score-function estimator benefits much
more from increasing the number of samples used. As the batch size is increased
in Figure~\ref{fig:blr_bs}, we again see a limited gain for the pathwise
estimator and more benefit for the score-function estimator.

When control variates are used with these estimators, the score-function
estimator benefits less from an increase in the number posterior samples or an
increase in batch size compared to the vanilla score-function estimator. The
performance of the vanilla score-function estimator can match that of the score
function estimator with a delta control variate, by increasing the number of
posterior samples: using $N=100, B=32$ the score-function estimator matches the
variational lower bound obtained by the score-function estimator with a delta
control variate when $N=10, B=32$. This exposes a trade-off between
implementation complexity and sample efficiency.

\paragraph{Same number of function evaluations.}

\begin{figure}[t]
	\centering
	\includegraphics[width=0.48\textwidth]{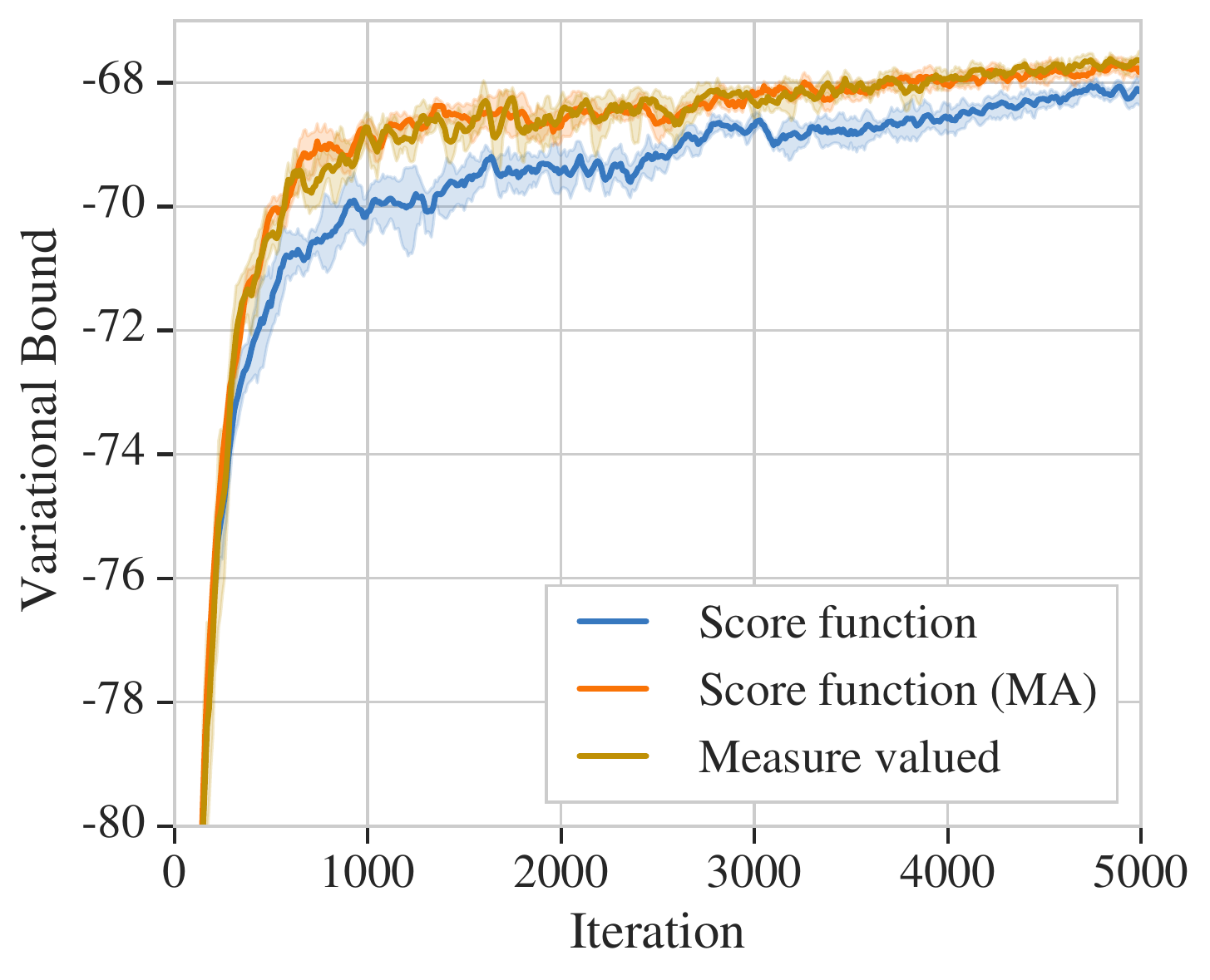}
	\includegraphics[width=0.48\textwidth]{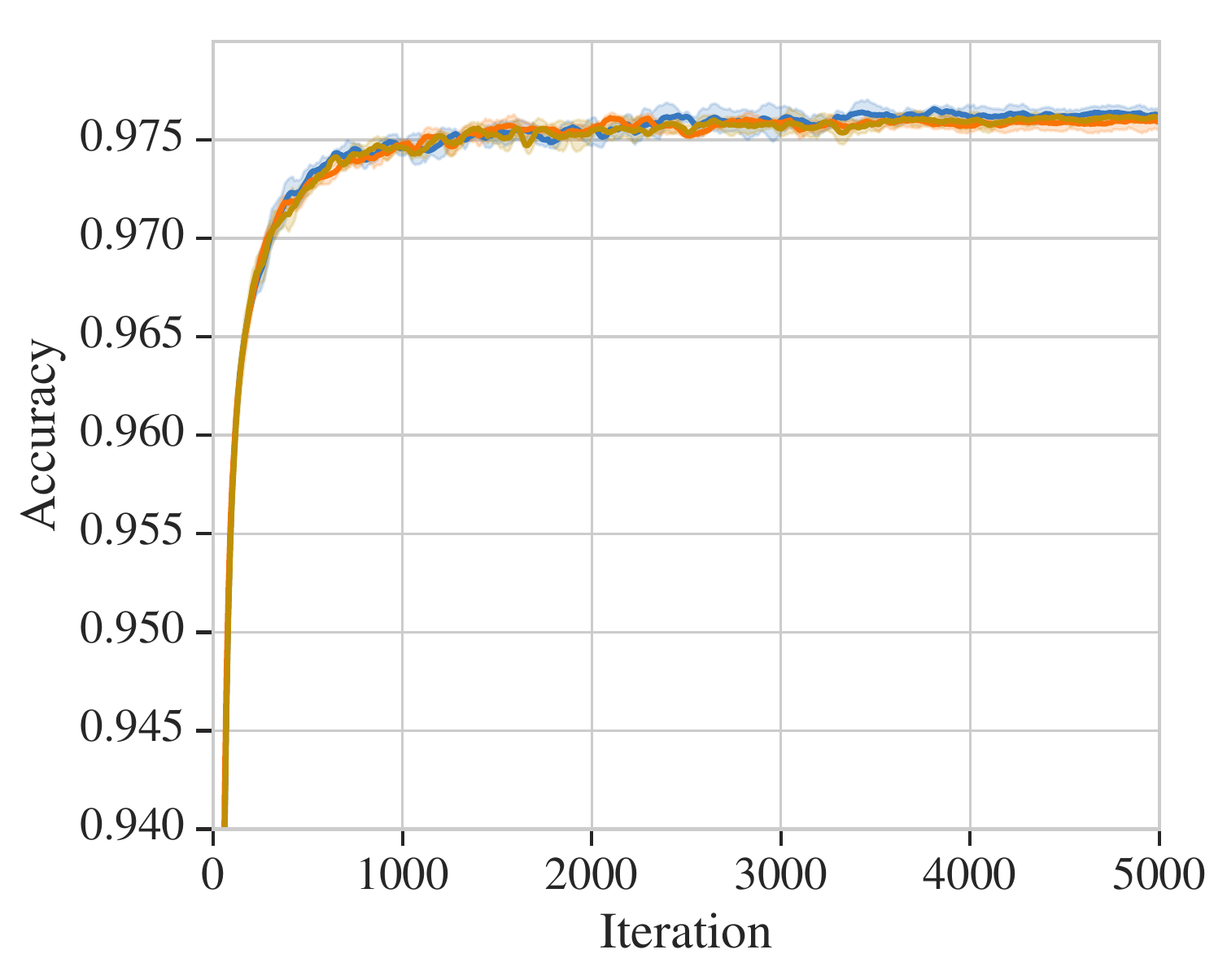}
	\includegraphics[width=0.48\textwidth]{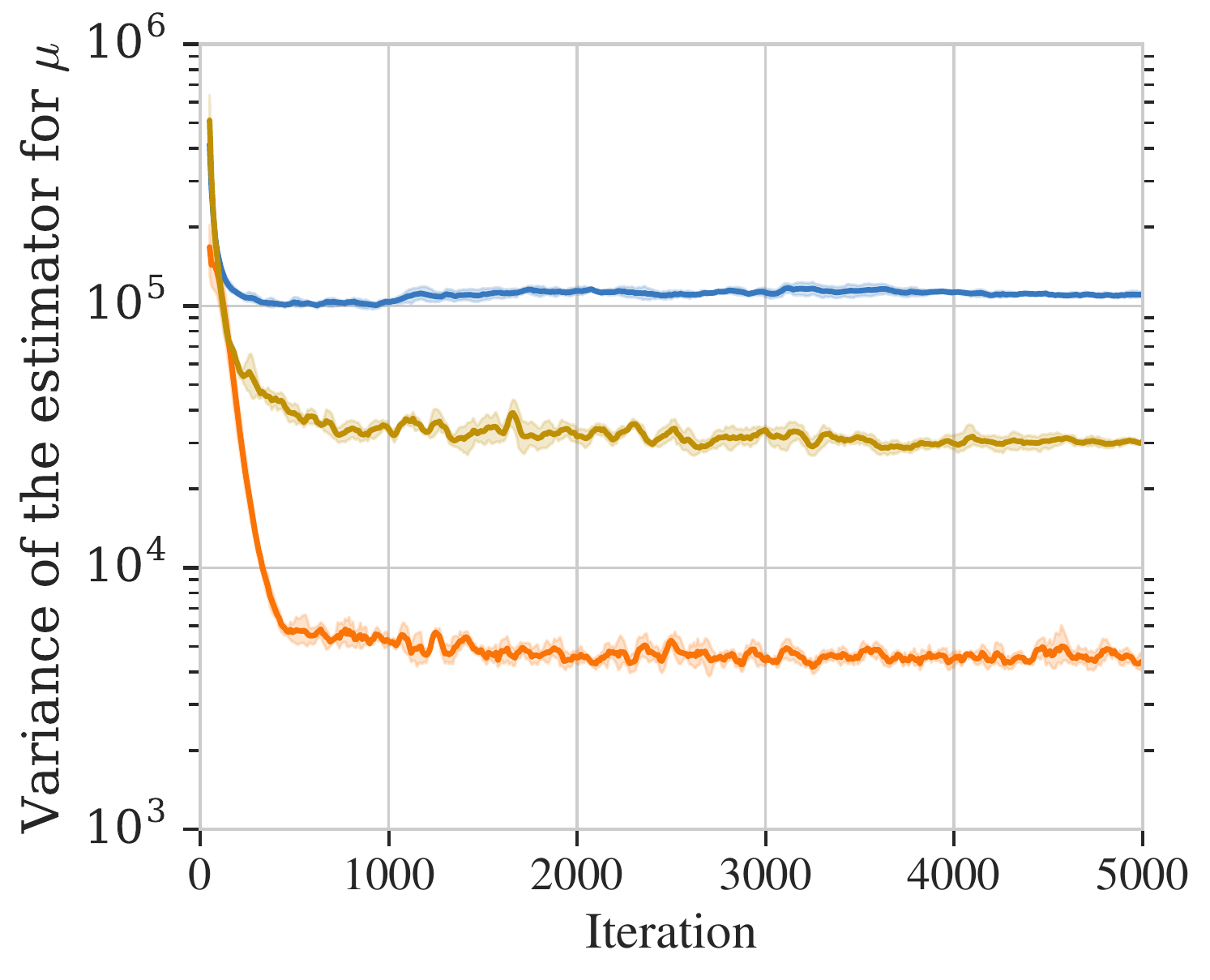}
	\includegraphics[width=0.48\textwidth]{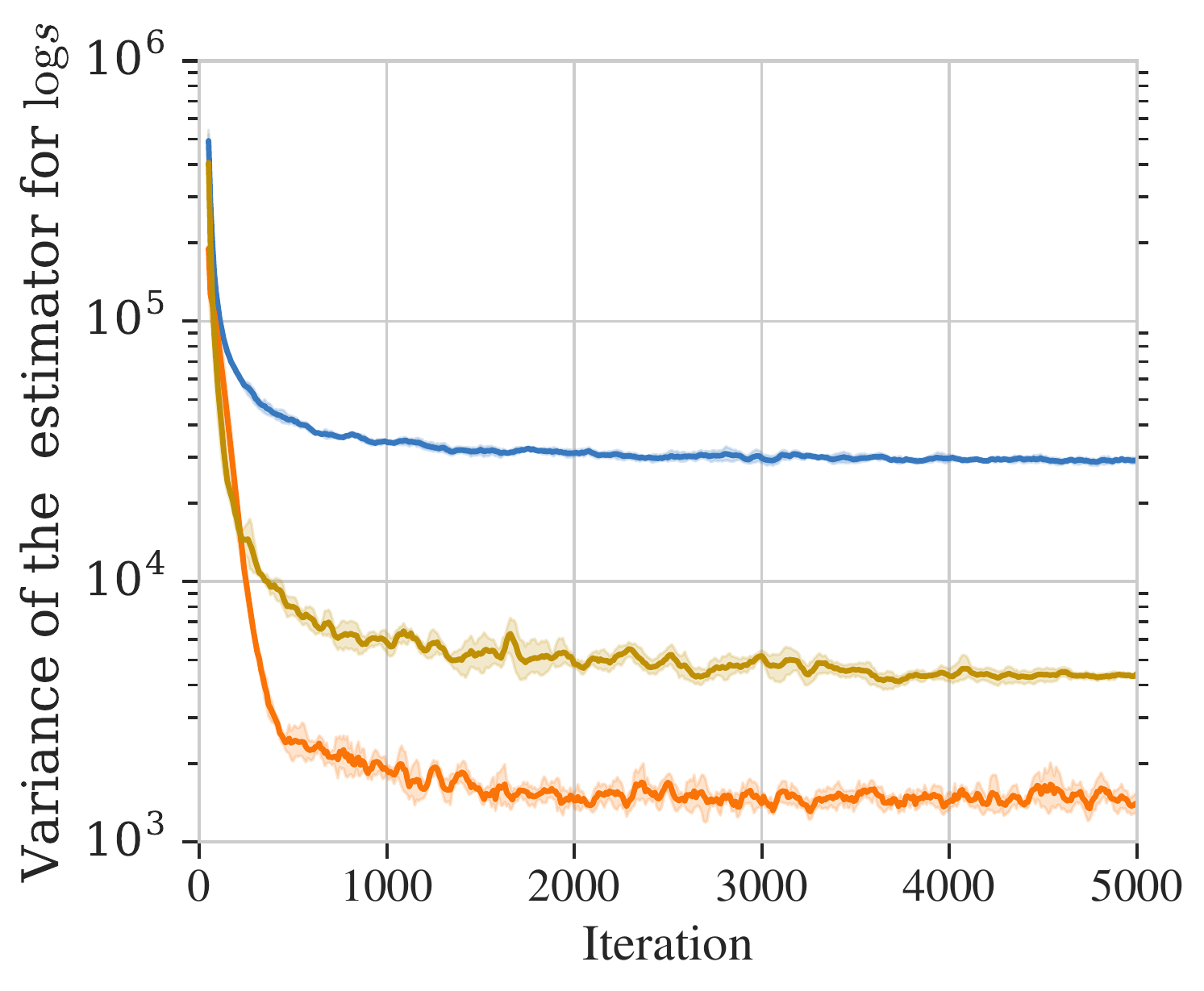}
  \caption{Comparison between the score-function and measure-valued
    estimators with the same number of function evaluations, with start learning rate ($10^{-3}$), $B=32$.
    For the measure-valued estimator $N=1$, while for score-function estimator, $N= 4D = 124$.
    For measure-valued derivatives, we always use coupling.
    In the second row, we show the
    variance of the estimator for the mean $\Var_{q(\vw| \vmu,
      \vs)}[\nabla_{\mu}f(\vw)]$ and the log-standard deviation $\Var_{q(\vw|
      \vmu,\vs)}[\nabla_{\log s}f(\vw)]$, averaged over parameter dimensions.}
	\label{fig:sc_measure_same_function_eval}
\end{figure}

Figure~\ref{fig:sc_path_1} shows that the measure-valued estimator achieves the same
variance as the pathwise estimator without making use of the gradient of the cost function.
The measure-valued estimator outperforms the score-function estimator, even when
a moving average baseline is used for variance reduction. This comes at a cost: the
measure-valued estimator uses  $2 |\vtheta| = 4 D$ more function evaluations.
To control for the cost, we also perform a comparison in which both methods use the same number of function evaluations
by using $4 D$ more samples for the score-function estimator and show results in Figure~\ref{fig:sc_measure_same_function_eval}.
In this scenario, the vanilla score function still exhibits higher variance than the measure-valued estimator, but when
augmented with a moving average baseline, which is a simple and computationally efficient variance reduction method, the score-function estimator performs best.

\paragraph{Controlling comparisons.} When working with control variates, we found
it useful to record for each iteration what the gradient variance would have
been for the current update had the control variate not been used, in order to
check that variance reduction is working. We show the results in this regime for
both pathwise and score-function estimators in Figure~\ref{fig:cv_no_cv}.

\section{Discussion and Extensions}
\label{sect:discussion}

\subsection{Testing Gradient Implementations}
Implementing Monte Carlo gradient estimators can be an error-prone process, so it is
important to verify the correctness of the implemented estimator. We consider
\textit{unbiased} gradient estimators, so a good check is to compare the Monte
Carlo estimate using a large number of samples to the exact gradient in a test
problem for which the analytic solution is known. As basic tests: if the mean
$\vmu$ of the measure is known, we can use $f(\vx) = x_i$; if the covariance
matrix of the measure is also known, we can pick $f(\vx) = (x_i - \mu_i) (x_j -
\mu_j)$. As an example, for the Gamma distribution $p(x ; \theta) = \mathcal{G}(x | \theta,
1)$, its mean is $\theta$ and the exact gradient $\nabla_{\theta} \expect{p(x |
	\theta)}{x} = 1$. Then, we approximate the left hand side using $N$ Monte Carlo
samples of a stochastic gradient estimator and compare the result to 1 with
some tolerance, choosing $N = 10^4$ samples is a good starting point.

It is also useful to check that, in expectation over a large number of
samples, different unbiased estimators produce the same
gradients for a variety of functions. With this approach, we can also use
functions for which we do not know the exact gradient of the expectation.
For methods that use control variates for variance reduction, it is useful to check
that the stochastic estimates of the expectation of the control variate and
its gradients are close to their analytically computed values.
To assess the variance
reduction effects of the control variate, it is useful to also track the
variance for the gradient estimator without the control variate, like we do in Figure~\ref{fig:cv_no_cv}.
These simple tests allow for a robust set of unit tests in code, to which more sophisticated testing methods can be added, e.g.~the tests described by \citet{geweke2004getting}.
For all the tests accompanying our implementation of the stochastic gradient estimators described in this work, see the note on the code in the introduction.

\subsection{Stochastic Computational Graphs and Probabilistic Programming.}
Computational graphs \citep{bauer1974computational} are now the standard formalism
with which we represent complex mathematical systems in practical systems,
providing a modular tool with which to reason about the sequence of computations
that transform a problem's input variables into its outputs. A computational
graph defines a directed acyclic graph that describes a sequence of
computations, from inputs and parameters to final outputs and costs.  This is
the framework in which automatic differentiation systems operate. Computational
graphs allow us to automatically track dependencies between individual steps
within our computations and then, using this dependency structure, enable the
sensitivity analysis of any parameters by  backwards traversal of the graph
\citep{griewank1989automatic, baydin2018automatic}. In standard computational
graphs, all operations are represented by nodes that are deterministic, but in
the problems that we have studied, some of these nodes will be random, giving
rise to \textit{stochastic computational graphs}.

Stochastic computational graphs provide a framework for implementing systems with both
deterministic and stochastic components, and
efficiently and automatically computing gradients using the types of estimators that we
reviewed in this paper \citep{schulman2015gradient, kucukelbir2017automatic}.
With such a framework it is easy to create complex
probabilistic objects that mix differentiable and non-differentiable cost
functions, and that use variables with complex dependencies. Simpler approaches for rich computational graphs of this type first approximate all
intractable integrals using the Laplace approximation and apply automatic
differentiation to the resulting approximation \citep{fournier2012ad}.
Stochastic variational message passing \citep{winn2005variational, paquet2013one} uses mean-field 
approximations, bounding methods, and  natural parameterisations of the distributions involved, to 
create another type of stochastic computational graph that uses more detailed information about the 
structural properties of the cost and the measure. 
Seeking greater automation, \citet{schulman2015gradient} exploit the
score-function and pathwise derivative estimators to automatically provide
gradient estimates for non-differentiable and differentiable components of the graph,
respectively.

The framework of \citet{schulman2015gradient}, \citet{kucukelbir2017automatic}, 
\citet{fournier2012ad}, \citet{parmas2018total} and 
others, shows that it is easy to combine
the score-function and pathwise estimators and automatically deploy them in
situations where they are naturally applicable. This framework has been further
expanded by incorporating methods that include variance reductions within the
computational graph, and allowing higher-order gradients to be computed more
easily and with lower variance \citep{weber2019credit, foerster2018dice,
	mao2018better, parmas18a}. There are as yet no automated computational frameworks that
include the measure-valued gradients. But as \citet{pflug1996optimization}
mentions, since the weak derivative---and also the sampling mechanism for the
positive and negative density components---is known for most of distributions in
common use (c.f. Table \ref{tab:mvd_triples}), it is possible to create a
computational framework with sets of nodes that have knowledge of their
corresponding weak derivatives to enable gradient computation. The score-function
estimator is implemented in stochastic computational graphs using surrogate
nodes and loss functions (c.f. \citet{schulman2015gradient}), and the measure-valued
gradients could be realised through similar representations.

The framework of a stochastic computational graph enables us to more easily
explore complex probabilistic systems, since we are freed from creating
customised implementations for gradient estimation and optimisation. They also
make it easier to develop probabilistic programming languages, that is programming
languages in which we specify generative and probabilistic descriptions of
systems and data, and then rely on automated methods for probabilistic
inference. Many of the approaches for automated inference in
probabilistic programs, like variational inference, rely on the automated
computation of Monte Carlo gradient estimates \citep{wingate2011nonstandard, ritchie2016deep,
	kucukelbir2017automatic}.

\subsection{Gaussian Gradients}
\label{sect:gaussian_gradients}
The case of Gaussian measures is a widely-encountered special case for gradient estimation.
If we know we will be using a Gaussian then we can derive estimators that exploit its specific 
properties. Consider the multivariate Gaussian
$\mathcal{N}(\vx |
 \vmu(\vtheta), \vSigma(\vtheta))$ with mean and variance that are functions of
 parameters $\vtheta$ whose gradients we are interested in. The most common
 approach for gradient computation is by pathwise estimation using the location-scale transform for 
 the
 Gaussian, $\vx = \vmu + \vR\vepsilon; \vepsilon \sim \mathcal{N}(\vzero,\vI)$ with
 the decomposition $\vSigma = \vR\vR^\top$:
 \begin{equation}
 \nabla_{\theta_i}  \expect{\mathcal{N}(\vx | \vmu(\vtheta), \vSigma(\vtheta))}{f(\vx)} = 
 \expect{\mathcal{N}(\vepsilon|\vzero, \vI)}{\vgamma \nabla_{\theta_i}\vmu + 
\textrm{Tr}(\vepsilon 
 \vgamma \nabla_{\theta_i}\vR)} = \expect{\mathcal{N}(\vepsilon|\vzero, \vI)}{\vgamma 
 \nabla_{\theta_i}\vmu 
 + \vgamma \nabla_{\theta_i}\vR \vepsilon}, \nonumber
 \end{equation}
 where $\vgamma = \nabla_{\vx}\cost(\vx)|_{\vx=\vmu + \vR\vepsilon}$.
 
Alternatively, since we are dealing specifically with Gaussian measures, we can use
the Bonnet's theorem \citep{bonnet1964transformations} and the Price's theorem
\citep{price1958useful, rezende2014stochastic} to directly compute the gradient
with respect to the Gaussian mean and covariance, respectively:
 \begin{align}
 & \nabla_{\mu_i} \expect{\mathcal{N}(\vx|\vmu, \vSigma)}{f(\vx)} = 
 \expect{\mathcal{N}(\vx|\vmu, \vSigma)}{\nabla_{x_i} f(\vx)},\\
 & \nabla_{\Sigma_{i,j}} \expect{\mathcal{N}(\vx|\vmu, \vSigma)}{f(\vx)} = 
 \tfrac{1}{2}\expect{\mathcal{N}(\vx|\vmu, \vSigma)}{\nabla^2_{x_i, x_j} f(\vx)}, 
\label{eq:pathwise_price} \\
 &\nabla_{\theta_i}  \expect{\mathcal{N}(\vx | \vmu(\vtheta), \vSigma(\vtheta))}{f(\vx)} =  
 \expect{\mathcal{N}(\vx | \vmu(\vtheta), \vSigma(\vtheta))}{\vgamma\nabla_{\theta_i}\vmu + 
 \tfrac{1}{2}\textrm{Tr}\left(\vH\nabla_{\theta_i} \vSigma \right)},
 \end{align}
 where $\vgamma$ and $\vH$ are the gradient and Hessian of the cost function
$f(\vx)$. Using the Price's theorem is computationally more expensive, since it requires evaluating 
the Hessian, which incurs a cubic computational cost, and will not be scalable
in high-dimensional problems. By comparison, the location-scale path has linear
cost in the number of parameters, but at the expense of higher variance.

We can also extend the Price's theorem \eqref{eq:pathwise_price} to the second order \citep{fan2015fast}:
  \begin{equation}
  \nabla^2_{\Sigma_{ij}, \Sigma_{kl}} \expect{\mathcal{N}(\vx| \vmu, \vSigma)}{f(\vx)} = 
  \tfrac{1}{4}\expect{\mathcal{N}(\vx| \vmu, \vSigma)}{\nabla_{x_i, x_j, x_k, x_l}f(\vx)},
  \end{equation}
  where $i,j,k,l$ are indices on the dimensionality of the parameters space;
$\Sigma_{ij}$ is the $(i,j)$th element of the covariance matrix $\vSigma$ and
$x_i$ is the $i$th element of the vector $\vx$. This shows that while we can
compute gradients in a form suitable for higher-order gradients, for Gaussian
measures, this comes at the cost of needing up to 4th-order differentiability
of the cost function.

Additional types of gradient estimators that work specifically for the case of Gaussian measures are 
developed by \citet{jankowiak2019pathwisemultivariate}, and explored for measure-valued derivatives 
by \citet{heidergott2008sensitivity}.

\subsection{Gradients Estimators using Optimal Transport}
\label{sect:opt_transport}
An interesting connection to the pathwise gradient estimators was identified by
\citet{jankowiak2018pathwise} using the tools of optimal transport. Because
probabilities are themselves conserved quantities, we will find that the
probabilities we use, as we vary their parameters, must satisfy the
continuity equation \citep[sect. 4.1.2]{santambrogio2015optimal}
\begin{align}
& \nabla_{\theta_i} p(\vx; \vtheta) + \textrm{div}_\vx(p(\vx; \vtheta)
\vupsilon(\theta_i)) = 0, \label{eq:disc_continuity}
\end{align}
where $\textrm{div}_\vx$ is the divergence operator
with respect to variables $\vx$, and $\vupsilon(\theta_i)$ is a velocity field for the
parameter $\theta_i$; there is a different velocity field for each parameter
$\theta_i$ for $i = 1, \ldots, D$. In univariate cases, the velocity
field $\vupsilon(\theta_i)$ can be obtained by solving the continuity equation to obtain
$\vupsilon(\theta_i) = - \tfrac{\nabla_{\theta_i} F(x; \vtheta)}{p(x;\vtheta)}$, where
$F(x; \vtheta)$  is the cumulative distribution function of $p$; also see Equation \eqref{eq:pathwise_implicit_univariate}. Using this definition, the gradient
\eqref{eq:gradient} for the $i$th parameter can be computed as
\begin{equation}
\eta_i =  \nabla_{\theta_i}\expect{p(\vx; \vtheta)}{f(\vx)} =
\expect{p(\vx; \vtheta)}{\nabla_{\vx} f(\vx) \vupsilon(\theta_i)},
\label{eq:disc_OT_grad}
\end{equation}
which can be derived by first interchanging the order of differentiation and
integration, substituting the density derivative with the expression for the
vector field given by the continuity equation and then simplifying the
expression using the product rule for divergences. One immediate connection we
see is that the solution for the velocity field is exactly the solution for the
quantity $\nabla_{\theta}\vx$ that we identified in Equation
\eqref{eq:pathwise_implicit_gradx}, when we use the cumulative distribution function as
the inverse transformation ($g^{-1}$). The form of the gradient that is
identified here was also derived in a very different way by \citet{cong2019go}.
In the multivariate case we can obtain a continuous set of solutions
of the continuity equation and then select the one with the lowest variance
\citep{jankowiak2018pathwise,jankowiak2019pathwisemultivariate}.
This set of solutions can be equivalently viewed as a set of control variates
for the pathwise gradient.
A deeper exploration of these connections, and the set
of tools that they make available from other areas of mathematics, stands to
provide us with many other approaches for gradient estimation.

\subsection{Harmonic Gradient Estimators}
\label{sect:harmonic_estimators}
We can add an additional class of gradient estimators to the set we
considered by decomposing the stochastic gradient problem \eqref{eq:gradient}
in terms of a Fourier series, resulting in the frequency domain or \textit{harmonic gradient
estimators} \citep{vazquez1994application, jacobson1999harmonic,
	jacobson1986driving}. The intuition for this estimator is that by oscillating the parameters of the
input measures sinusoidally, we can, at different frequencies, observe the
sensitivity (gradient) of the \gradCost with respect to the parameters of
interest.

The harmonic approach turns out to be
an efficient way of implementing a weighted finite-difference method with
sinusoidally varying step-sizes \citep{jacobson1994optimal,
	vazquez1994application}. This also connects us to many other weighted finite
difference schemes that might be applicable for the problem of gradient
estimation, e.g., \citep{fu2016differentiation}. In the context of Monte Carlo
gradient estimation, the difficulty in choosing the hyperparameters of the estimator (the so-called index, driving frequencies and
oscillation amplitudes, \citealt{fu1994optimization})   makes this the least generic
of the estimators we have considered; it also has high computational complexity
requiring many evaluations of the cost function e.g.~\citep{fu2016differentiation}. Despite the drawbacks of this
construction, the general use of Fourier decompositions remains an interesting
additional tool with which to exploit knowledge in our estimation problems.

\subsection{Generalised Pathwise Gradients and Stein-type Estimators}
In many statistical problems, the gradient of the entropy of a probabilistic model is an often required
informational quantity:
\begin{equation}
\nabla_{\vtheta}\mathbb{H}[p(\vx; \vtheta)] = -\nabla_{\vtheta}\expect{p(\vx; \vtheta)}{\log 
p(\vx; \vtheta)} = - \expect{p(\vx; \vtheta)}{\nabla_{\vx}\log p(\vx; \vtheta) \nabla_{\vtheta} \vx},
\label{eq:stein_entropy} 
\end{equation}
where we first expanded the definition of the entropy, and then applied the chain rule. 
Here, the log probability takes the role of the cost function in our exposition. Equation 
\eqref{eq:stein_entropy} is an expression in the more familiar form 
$\expect{\dist(\vx;
	\distParams)}{\nabla_\vx \cost(\vx) \nabla_{\distParams} \vx}$ that we encountered in the 
application of the pathwise estimator, particularly in the decoupling of sampling from 
gradient computation, whether using implicit differentiation
\eqref{eq:pathwise_implicit} or the continuity equation
\eqref{eq:disc_OT_grad}. 

To apply the pathwise estimator, we must be able to differentiate the
log-probability, which implies that it must be known.  This will not always be
the case, and one common setting where this is encountered in machine learning
is when using likelihood-free or implicit probabilistic models
\citep{mohamed2016learning}. With implicit models, we can simulate data from a
distribution, but do not know its log-probability. In such cases, the general
strategy we take is to use a substitute function that fills the role of the
unknown cost function, like the unknown log-probability, and that gives the
gradient information we need, e.g.~like the use of a classifier in generative
adversarial networks \citep{goodfellow2014generative} or a moment network in the
method of moments \citep{ravuri2018learning}. In this way, we can create a more
general pathwise estimator that can be used when the gradient of the cost
function is unknown, but that will reduce to the familiar pathwise estimator
from Section \ref{sect:pathwise_estimators} when it is. Stein's identity gives
us a general tool with which to do this.

Stein's identity \citep{gorham2015measuring, liu2016kernelized} tells us that if
a distribution $p(\vx)$ is continuously differentiable then the gradient of its
log-probability with respect to the random variable $\vx$ can be related to the gradient of a test 
function
$\vt(\vx)$ by the integral
\begin{equation}
\expect{p(\vx)}{\vt(\vx)\nabla_{\vx}\log p(\vx) + \nabla_{\vx} \vt(\vx)} = \vzero_{K\times D}.
\end{equation}
We assume that test function $\vt(\vx)$ has dimension $K$, the random
variable $\vx$ has dimension $D$, and that $\vt$ is in the Stein class of $p$ defined by 
\citet{gorham2015measuring}. By inverting this expression, we can find a general
method for evaluating the gradient of a log probability. This expression is
applicable to both normalised and unnormalised distributions, making it very general, i.e.~we can use
Stein's identity to provide the unknown component $\nabla_{\vx} f(\vx)$. 
By expanding Stein's identity as a Monte Carlo estimator and using several samples,
\citet{li2017gradient} use kernel substitution to develop a kernelised method
for gradient estimation in implicit probabilistic models. They
also provide the connections to other methods for estimating gradients of
log-probabilities, such as score-matching and denoising auto-encoders. An
alternative Stein-type gradient estimator using the spectral decomposition of
kernels was developed by \citet{shi2018spectral}, and Stein's identity was also
used in variational inference by \citet{ranganath2016operator}.

\subsection{Generalised Score Functions and Malliavin-weighted Estimators}
\label{sect:malliavinweighted}
Stein's identity gave us a tool with which to develop generalised pathwise estimators, and in a similar vein 
we can create a generalised score-function estimator. There are many problems where the 
explicit form of the density needed to compute the score is not known, such as when the density is defined 
by a continuous-time system or an implicit probabilistic model. In such cases, a generalised form of 
the score-function estimator \eqref{eq:score_estimation_final}
is $\expect{p(\vx; \vtheta)}{f(\vx)\vpsi(\vx, \vtheta)}$, i.e.~the expected
value of the cost function multiplied by a weighting function $\vpsi$.
When the weighting function is the score function, we will recover the score-function gradient, and in other 
cases, it gives us the information needed about the unknown score.

One approach for developing generalised score-function estimators that is widely
used for continuous-time systems  in computational finance is through Malliavin calculus, which is
the calculus of variations for stochastic processes
\citep{nualart2006malliavin}. Computational finance provides one example of a
setting where we do not know the density $p(\vx; \vtheta)$ explicitly, because the
random variables are instead specified as a diffusion process. The generalised
gradient method based on weighting functions, called Malliavin weights, allows
gradients to be computed in such settings, where the score function is replaced
by a Skorokhod integral  \citep{chen2007malliavin}. There are equivalent concepts
of differentiation (Malliavin derivative) and integration (Skorokhod integration)
that allow for the interchange of differentiation and integration and the
development of unbiased estimators in such settings. We defer to the papers by
\citet{fournie1999applications} and \citet{benhamou2003optimal} in computational
finance, and, closer to machine learning, the papers by
\citet{gobet2005sensitivity} and \citet{munos2006policy} in optimal control and
reinforcement learning for further exploration of these ideas.

\subsection{Discrete Distributions}
We purposefully limited the scope of this paper to measures that are continuous
distributions in their domain. There are many problems, such as those with
discrete latent variables and in combinatorial optimisation, where the
measure $p(\vx; \vtheta)$ will instead be a discrete distribution, such as the Bernoulli
or Categorical. This is a large area of research in itself, but does have some
overlap with our coverage in this paper. While the Bernoulli or Categorical are
discrete distributions over their domain, they are continuous with respect to
their parameters, which means that both the score-function estimator
and the measure-valued derivative are applicable to such estimation problems.
Coverage of this discrete distribution setting makes for an interesting review paper in itself, and existing work 
includes \citet{tucker2017rebar, maddison2016concrete, mnih2014neural, gu2015muprop, glasserman1991gradient}.

\section{Conclusion}
\label{sect:conclusion}
We have been through a journey of more than fifty years of research in one of
the fundamental problems of the computational sciences: computing the gradient
of an expectation of a function. We found that this problem lies at the core of
many applications with real-world impact, and a broad swathe of research
areas; in applied statistics, queueing theory, machine learning, computational
finance, experimental design, optimal transport, continuous-time stochastic
processes, variational inference, reinforcement learning, Fourier analysis,
causal inference, and representation learning, amongst many others subjects. The gradients in all these areas fall into one of two classes,
gradients-of-measure or gradients-of-paths. We derived the score-function
estimator and the measure-valued gradient estimator as instances of
gradients of measure, both of which exploit the measure in the stochastic
objective to derive the gradient. And we derived the pathwise estimator that
uses knowledge of the sampling path to obtain the gradient. All these methods
benefit from variance reduction techniques and we reviewed
approaches for variance reduction we might consider in practice. We further
explored the use of these estimators through a set of case studies, and explored
some of the other tools for gradient estimation that exist beyond the three
principal estimators.

\subsection{Guidance in Choosing Gradient Estimators}
With so many competing approaches, we offer our rules of thumb in choosing an estimator, which
follow the intuition we developed throughout the paper:
\vspace{-4mm}
\begin{itemize}[noitemsep]
	\item If our estimation problem involves continuous functions and measures that
are continuous in the domain, then using the pathwise estimator is a good
default. It is relatively easy to implement and its default implementation,
without additional variance reduction, will typically have variance that is low
enough so as not to interfere with the optimisation.
	\item If the cost function is not differentiable or is a black-box function then
the score-function or the measure-valued gradients are available. If the
number of parameters is low, then the measure-valued gradient will typically
have lower variance and would be preferred. But if we have a high-dimensional
parameter set, then the score-function estimator should be used.
	\item  If we have no control over the number of times we can evaluate a
black-box cost function, effectively only allowing a single evaluation of it,
then the score function is the only estimator of the three we reviewed that is
applicable.
	\item The score-function estimator should, by default, always
be implemented with at least some basic variance reduction. The simplest option is to use a baseline
control variate estimated with a running average of the cost value.
	\item When using the score-function estimator, some attention should be paid to the
dynamic range of the cost function and its variance, and ways found to
keep its value bounded within a reasonable range, e.g.~by transforming the cost
so that it is zero mean, or using a baseline.
	\item For all estimators, track the variance of the gradients if possible and
address high variance by using a larger number of samples from the measure,
decreasing the learning rate, or clipping the gradient values. It may also be
useful to restrict the range of some parameters to avoid extreme values,
e.g.~by clipping them to a desired interval.
	\item The measure-valued gradient should be used with some
coupling method for variance reduction. Coupling strategies that exploit
relationships between the positive and negative components of the density
decomposition, and which have shared sampling paths, are known for the
commonly-used distributions.
	\item If we have several unbiased gradient estimators, a convex combination of them
might have lower variance than any of the individual estimators.
	\item If the measure is discrete on its domain then the score-function or
measure-valued gradient are available. The choice will again depend on the
dimensionality of the parameter space.
	\item In all cases, we strongly recommend having a broad set of tests to verify
the unbiasedness of the gradient estimator when implemented.
\end{itemize}
\subsection{Future Work} 
With all the progress in gradient estimation that we have covered, there still
remain many more directions for future work, and there is still the need for new types
of gradient estimators. Gradient estimators for situations where we know
the measure explicitly, or only implicitly though a diffusion, are needed. New
tools for variance reduction are needed, especially in more complex
systems like Markov processes. The connections between the score function and
importance sampling remains to be more directly related, especially with
variance reduction in mind. Coupling was a tool for variance reduction that we
listed, and further exploration of its uses will prove fruitful. And there
remain opportunities to combine our existing estimators to obtain more
accurate and lower variance gradients.  The measure-valued derivatives have not
been applied in machine learning, and it will be valuable to understand the
settings in which they might prove beneficial. Gradient estimators for continuous-time
settings like the Malliavin-weighted versions, multivariate estimators derived
using optimal transport, the use of non-linear control covariates,  the use of
higher-order gradients, variants using the natural gradient and information geometry, and the 
utility of Fourier series are among many of the
other questions that remain open. %

\textit{Good gradients lead to ever-more interesting applications; ultimately, there remains much more 
to do in advancing the principles and practice of Monte Carlo gradient estimation.}

\section*{Acknowledgements}
We thank the action editor and our anonymous reviewers for their feedback. We are also grateful to many of our colleagues for their invaluable feedback
during the preparation of this paper. In particular, we would like to thank
Michalis Titsias, Cyprien de Masson d'Autume, Csaba Cspesvari, Fabio Viola,
Suman Ravuri, Arnaud Doucet, Nando de Freitas, Ulrich Paquet, Danilo Rezende,
Theophane Weber, Dan Horgan, Jonas Degrave, Marc Deisenroth, and Louise Deason. Finally, we thank many from the wider machine learning community who sent us points of improvement and requests for clarifications that strengthened this paper greatly.

\appendix
\section{Common Distributions}
\label{app:distribution}
The following table provides the definitions of the distributions that were mentioned within the 
paper.
{\renewcommand{\arraystretch}{1.5}%
\begin{table}[hbt]
	\centering
	
\begin{tabular}{|l|c | l||l|}
	\hline 
	\textbf{Name} & \textbf{Domain} & \textbf{Notation} & \textbf{Probability Density/Mass Function} \\ 
	\hline \hline 
	Gaussian & $\mathbb{R}$  &$\mathcal{N}(x | \mu, \sigma^2)$&  
	$\frac{1}{\sqrt{2\pi\sigma^2}}\exp\left(-\frac{1}{2}\left(\frac{x-\mu}{\sigma}\right)^2\right)$\\ 
	Double-sided Maxwell&  $\mathbb{R}$  & $\mathcal{M}(x |\mu, \sigma^2)$ &  
	$\frac{1}{\sigma^3\sqrt{2\pi}}(x-\mu)^2 
	\exp\left( -\frac{(x-\mu)^2}{2\sigma^2}\right) $\\ 
	Weibull &  $\mathbb{R}^+$  &$\mathcal{W}(x |\alpha, \beta, \mu) $&   $\alpha\beta 
	(x-\mu)^{\alpha-1} 
	\exp(-\beta 
	(x-\mu)^\alpha)\mathds{1}_{\{x\geq 0\}}$\\ 
	Poisson & $\mathbb{Z}$  &$\mathcal{P}(x |\theta)$ & $ 
	\exp(-\theta)\sum_{j=0}^{\infty}\frac{\theta^j}{j!}\delta_j$\\
	Erlang &  $\mathbb{R}^+$  &$\mathcal{E}r(x |\theta, \lambda)$&  $\frac{\lambda^\theta 
	x^{\theta-1}\exp(-\lambda 
	x)}{(\theta -1)!}$\\
	Gamma & $\mathbb{R}^+$  & $\mathcal{G}(x |\alpha, \beta)$& $ 
	\frac{\beta^\alpha}{\Gamma(\alpha)}x^{\alpha-1}\exp(-x\beta)\mathds{1}_{\{x\geq 0\}}$\\
	Exponential &  $\mathbb{R}^+$  &$\mathcal{E}(x |\lambda)$& $\mathcal{G}(1,\lambda) $\\
	\hline 
\end{tabular}
\caption{List of distributions and their densities.}
\label{tab:distr}
\end{table} }

\bibliography{references}
\end{document}